\documentclass{article} % For LaTeX2e
\usepackage{iclr2022_conference,times}

\usepackage{amsmath,amsfonts,bm}

\def\eqref#1{equation~\ref{#1}}
\def\1{\bm{1}}

\DeclareMathAlphabet{\mathsfit}{\encodingdefault}{\sfdefault}{m}{sl}
\SetMathAlphabet{\mathsfit}{bold}{\encodingdefault}{\sfdefault}{bx}{n}

\usepackage{hyperref}
\usepackage{url}
\usepackage{soul, xcolor}
\usepackage{graphicx}
\usepackage{array, blindtext}
\usepackage{mwe}
\usepackage{subcaption}
\usepackage{algorithm}
\usepackage[noend]{algorithmic}
\usepackage{amssymb}
\usepackage{hyperref}
\usepackage{url}
\usepackage{color,soul}
\usepackage{makecell}
\usepackage{threeparttable}
\usepackage{tikz}
\usepackage{multirow}
\usepackage{booktabs,wrapfig}
\usepackage{bbm}
\usepackage{amsthm}
\usepackage{setspace}

\newcommand{\passivelearningabbr}{\textbf{PL}}
\newcommand{\diagnosiseiteachabbr}{\textbf{D(EI)+T}}
\newcommand{\diagnosisrteachabbr}{\textbf{D(R)+T}}
\newcommand{\randomteachabbr}{\textbf{RT}}

\title{Know Thy Student: Interactive Learning with Gaussian Processes}

\author{Rose E. Wang, Mike Wu, Noah Goodman  \\
Stanford University \\
\texttt{\{rewang, wumike, ngoodman\}@stanford.edu} \\
}

\iclrfinalcopy % Uncomment for camera-ready version, but NOT for submission.
\begin{document}

\maketitle

\begin{abstract}
Learning often involves interaction between multiple agents. 
Human teacher-student settings best illustrate how interactions result in efficient knowledge passing where the teacher constructs a curriculum based on their students' abilities. 
Prior work in machine teaching studies how the teacher should construct optimal teaching datasets assuming the teacher knows everything about the student.
However, in the real world, the teacher doesn't have complete information about the student. The teacher must interact and \textit{diagnose} the student, before teaching.
Our work proposes a simple diagnosis algorithm which uses Gaussian processes for inferring student-related information, before constructing a teaching dataset.
We apply this to two settings. 
One is where the student learns from scratch and the teacher must figure out the student's learning algorithm parameters, eg. the regularization parameters in ridge regression or support vector machines. 
Two is where the student has partially explored the environment and the teacher must figure out the important areas the student has not explored; we study this in the offline reinforcement learning setting where the teacher must provide demonstrations to the student and avoid sending redundant trajectories.
Our experiments highlight the importance of diagosing before teaching and 
demonstrate how students can learn more efficiently with the help of an interactive teacher. We conclude by outlining where diagnosing combined with teaching would be more desirable than passive learning. 
\end{abstract}

\section{Introduction}

In natural systems, learning often involves interaction between multiple agents. 
Teacher-student interactions best illustrate how interactions result in efficient knowledge passing. 
One way teachers interact with students is through concept tests---short, targeted \textit{tests} administered at the start of a class to help teachers gauge which concepts students have understood
\citep{10.1145/2839509.2844559,Treagust1988DevelopmentAU,Adams2011DevelopmentAV}.
By first understanding what their students struggle on, teachers can adapt the curriculum to best address the needs of their class and avoid repeating topics their students have already mastered.
In other words, teachers construct diagnostic tests to inform them how they should later construct ``training sets'' for their students.

We take inspiration from this form of teacher-student interactions in reframing how we'd like to efficiently train and test machine learning models. 
As ML practitioners and researchers, we implicitly interact with our model learners in this way to diagnose the learner.
For example, we evaluate our models by analyzing their performance on a test set.
% we essentially try to understand what our model does well vs. poorly. 
The model's test performance can inform us about the model's training dataset, for example data-centric properties like class imbalance \citep{vanhorn2018inaturalist,rahman2013addressing}.
This feedback can then suggest remedies for how to retrain our model, for example adding more examples of minority classes to balance the dataset \citep{hernandez2013empirical}.
The collective learning that takes place between a teacher and a student inspires our work to design principled methods for constructing first diagnostic datasets, then training datasets. 

We focus on the problem setup where the teacher must first diagnose the student, then construct a teaching dataset.
The performance of the teacher is based on how well the student performs on a held-out test dataset.  
We build on prior work in machine teaching \citep{zhu2018overview,liu2016teaching,cakmak2012algorithmic} for constructing optimal teaching datasets. 
We use Gaussian Processes \citep{williams2006gaussian} as a tractable means for inferring student parameters when diagnosing.
We combine machine teaching with diagnosing, and apply this framework to linear models for ridge regression, support vector machines (SVMs) and offline reinforcement learning settings. 

We apply this to two settings. 
One is where the student learns from scratch and the teacher must figure out the student's learning algorithm parameters, eg. the regularization parameters in ridge regression or support vector machines. 
Two is where the student has partially explored the environment and the teacher must figure out the important states the student has not learn the optimal value for; we study this in the offline reinforcement learning setting where the teacher provides demonstrations to the student and should avoid sending redundant trajectories.

In summary, our work's contributions are the following: 

\begin{itemize}
    \item We show that the teacher must diagnose the student, otherwise the student can have arbitrarily bad performance.
    \item We show that if the teacher does first diagnose, the learner can learn from much fewer training examples than the teacher. In the offline RL setting, this results in $2$-$16,000 \times$ fewer  training steps.
    \item We show that in certain cases, diagnosing with machine teaching leads to recovering optimal model parameters, where passive learning or naive machine teaching fail to do so. 
\end{itemize}

\begin{figure}
    \centering
    \newcommand{\factor}{0.28}
    \subfloat[\label{fig:gridworld_env}]{\includegraphics[width=\factor\textwidth]{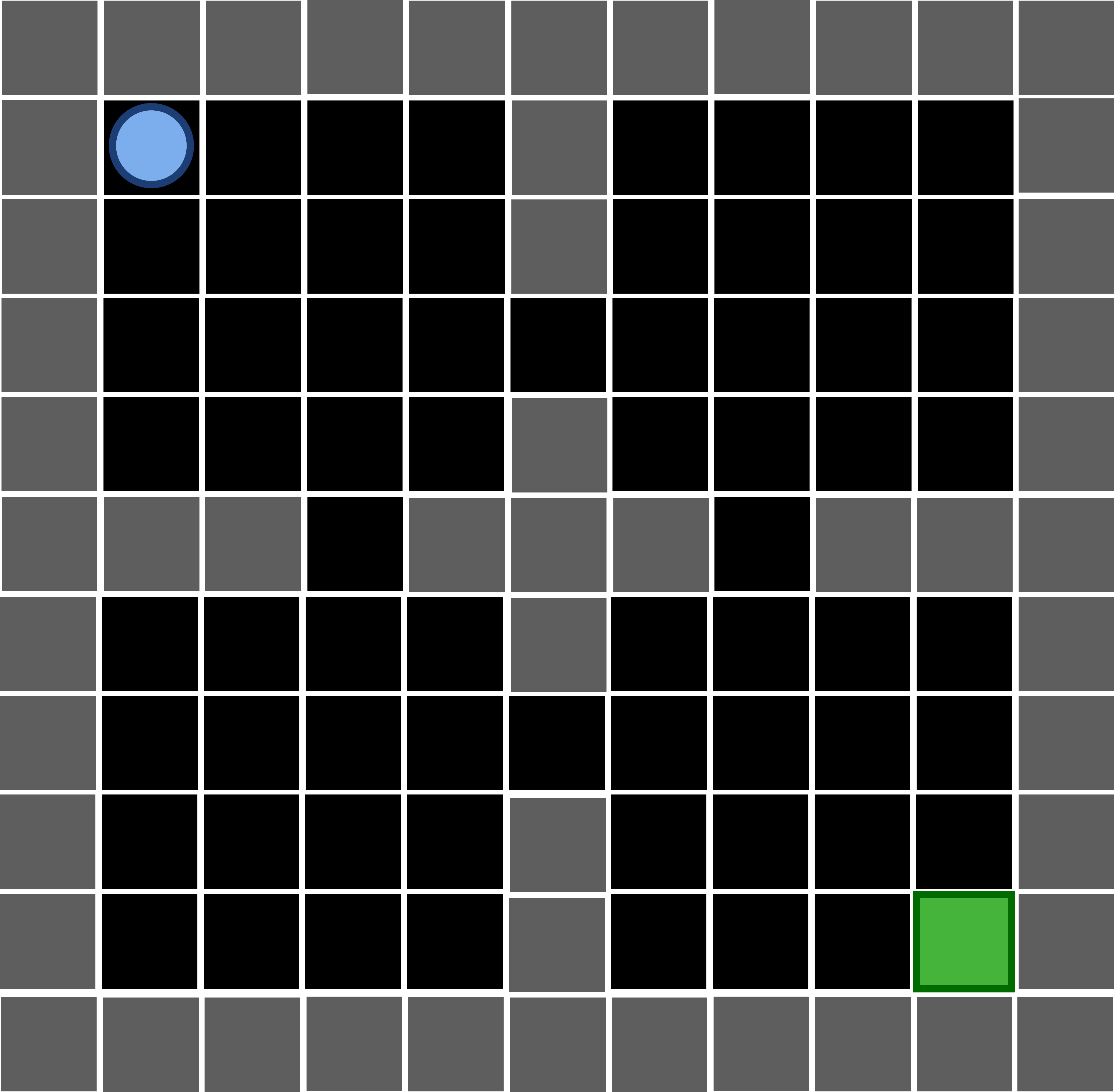}} \hfill
    \subfloat[\label{fig:gridworld_student_visited}]{\includegraphics[width=\factor\textwidth]{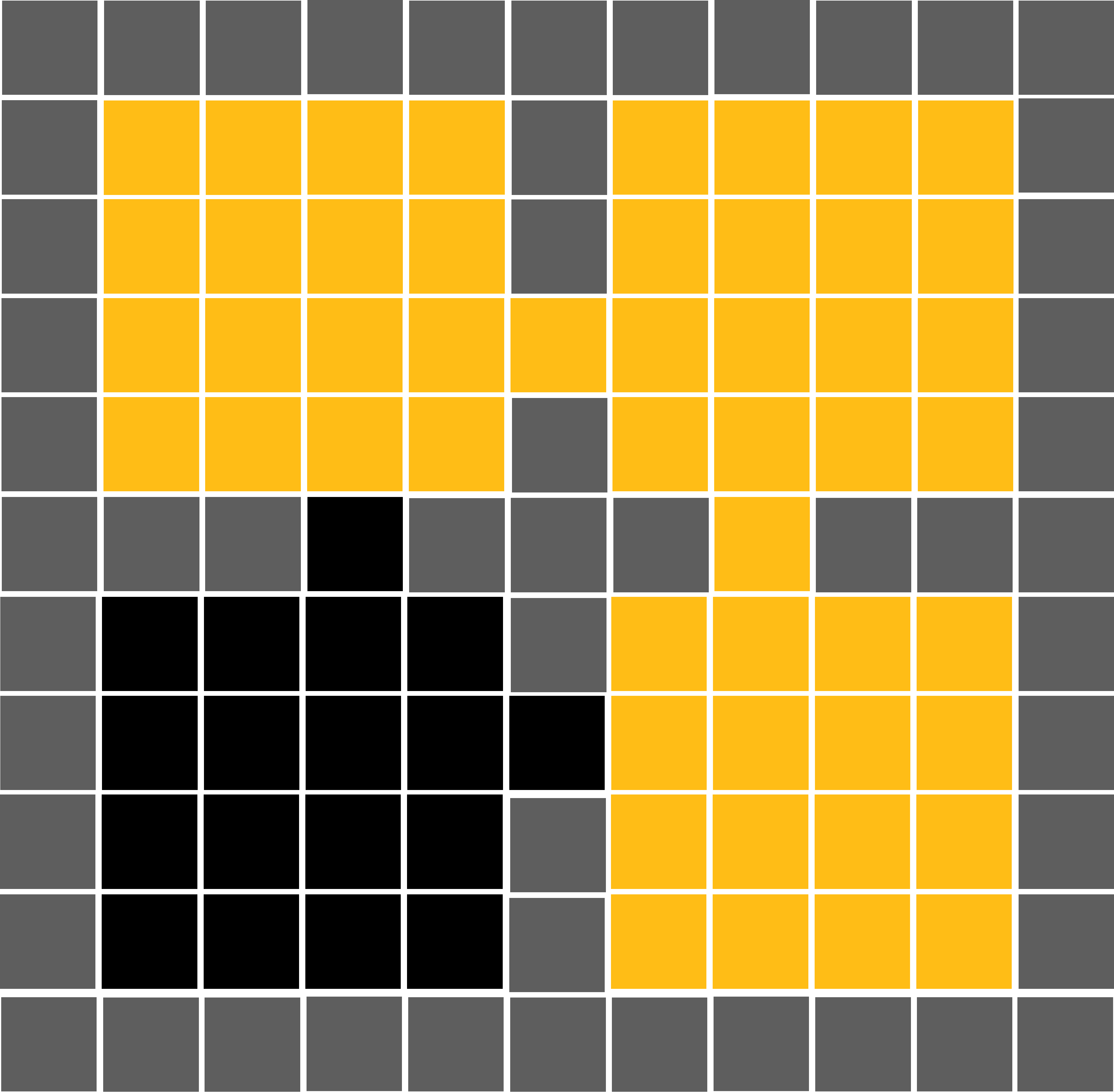}} \hfill
    \subfloat[\label{fig:gridworld_inferred}]{\includegraphics[width=\factor\textwidth]{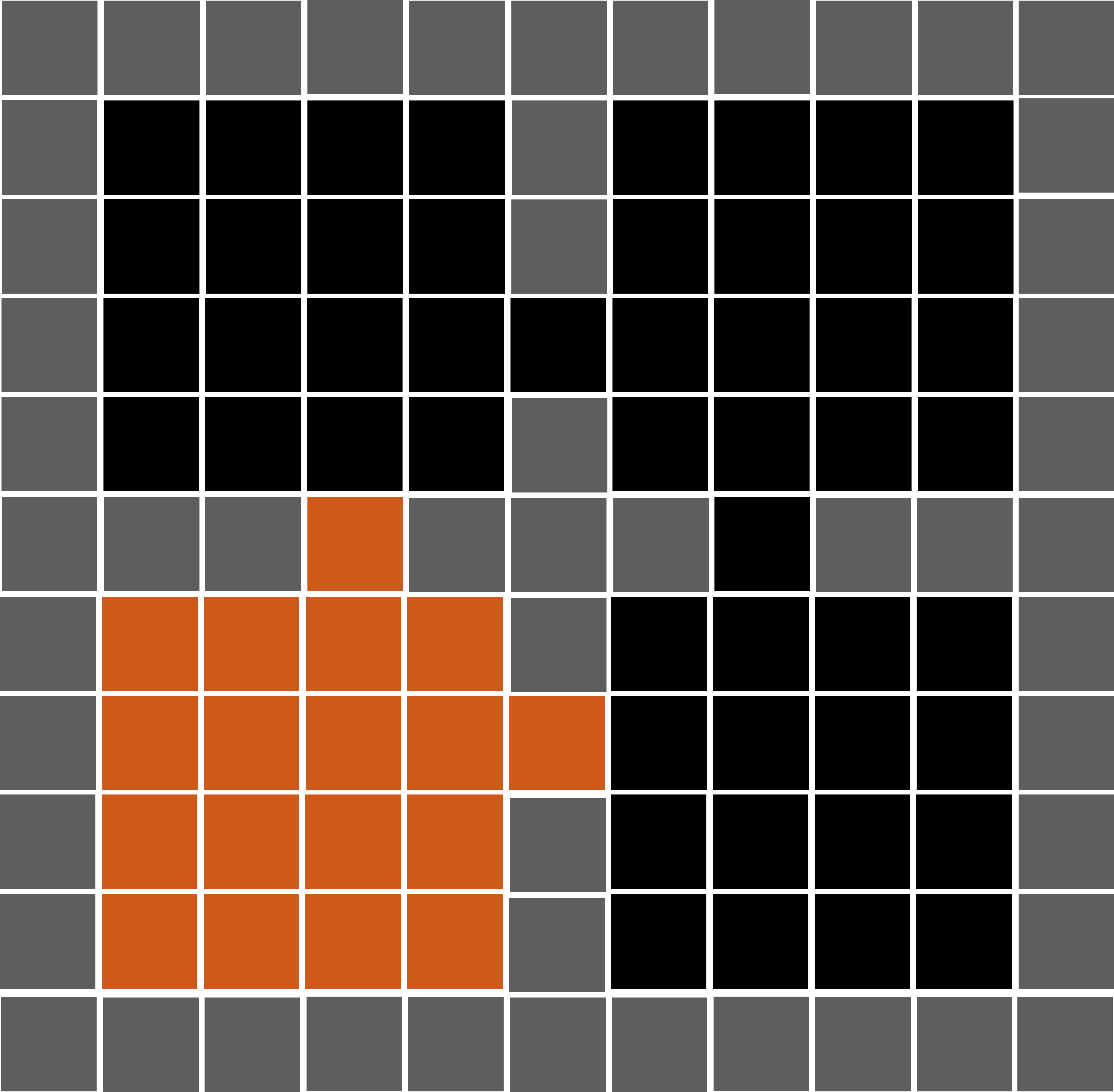}} \hfill
    \caption{Teacher needs to figure out what the student knows vs. doesn't know. 
    (a) depicts an example of a navigation environment where the student (blue, upper left) has to navigate to a goal location (green, lower right). 
    (b) illustrates that in some cases, the student has partial knowledge of the environment; from any of the yellow-highlighted cells, the student knows how to navigate to the green cell.
    (c) highlights the cells from which the student does not know how to navigate to the goal. 
    The ideal teacher is one that can identify these states and construct examples from these states.}
    \label{fig:example_teacher_inference_bob_knowledge}
\end{figure}
% \begin{figure}
%     \centering
%     \includegraphics[width=0.5\textwidth]{images/A.pdf}
%     \caption{Overview of machine teaching and learning taken from \citet{zhu2018overview}. \todo Two settings illustration: tabula rasa and Bob the student knows something}
%     \label{fig:machine_teaching_overview}
% \end{figure}

\section{Related works}

\paragraph{Machine teaching} 
The machine teaching problem is to determine an optimal training set, given knowledge about the student algorithm (eg. ridge regression) and the optimal model parameters $\theta^*$ \citep{zhu2018overview, zhu2015machine,simard_machine_2017}. 
An \textit{optimal} dataset is defined by the cardinality of the dataset, also known as the teaching dimension \citep{goldman_complexity_nodate}. 
Machine teaching has been broadly applied to settings like inverse reinforcement learning \citep{cakmak2012algorithmic,brown_machine_2019,hadfield-menell_cooperative_2016} where the goal is to construct a small demonstration dataset for the learner to most confidently learn the underlying reward function.
Our work builds on results from \citet{liu2016teaching} which derive the teaching dimension of linear learners in settings including ridge regression and SVMs.

\paragraph{Imitation learning} 
The imitation learning problem looks at how a student should learn a policy from an expert demonstration dataset, 
either by inferring the underlying reward that produced the behavior (inverse reinforcement learning, IRL) \citep{abbeel_apprenticeship_2004} or 
directly learning the policy (behavioral cloning, BC) \citep{torabi2018behavioral}. 
Our work is close to the interactive imitation learning setting \citet{ross_reduction_2011} where we rely on a teacher to provide action labels to states.
The key difference is that we rely on a teacher to first probe and determine which states it should provide labels to; 
these might be states the student has never visited before.

\paragraph{Active learning} 
The active learning problem looks at how a student should query a teacher for labels on unlabelled examples \citep{settles2009active}. 
Active learning is different from machine teaching in that the learners choose the datapoints they want to be trained on, rather than being guided by a teacher. 
However, the active learning framework might not be the right framework in settings where the learner doesn't have access to examples its never seen before.
For example, DAgger \citep{ross_reduction_2011} limits interaction between the learner and the teacher to the states the learner visits.
This doesn't fully leverage what the learner \textit{could} learn from the teacher, such as the teacher's privileged access to states that are important for a task of interest. 

\section{Interactive learning  with Gaussian Processes \label{sec:method}}

We study the problem of the teacher needing to infer properties of the student which will impact how the teacher teaches. 
In our setting, the teacher \textit{first} diagnoses, \textit{then} teaches the student. 
Extensions of this work could explore interweaving both diagnosis and teaching.

We use a Gaussian Process with a radial-basis function kernel for inference; 
alternative inference methods can be used as well.
In our settings, the teacher must either infer something about the student's learning algorithm (eg. a hyperparameter) or the student's knowledge base (eg. its original training dataset).

\paragraph{Notation} We denote $\mathcal{X}$ as the input space and $\mathcal{Y}$ as the output space.
$f_{\theta}: \mathcal{X} \rightarrow \mathcal{Y}$ is the student model parameterized by model weights $\theta \in \Theta$.
We refer to $D$ as a dataset, and $\mathcal{D}$ as the space of datasets.
A student runs an algorithm on a dataset to recover parameters that minimize the algorithm's learning objective, $A: \mathcal{D} \rightarrow \Theta$.

\begin{algorithm}[t]
    \caption{Diagnosing student model}
    \label{alg:student_diagnosis_with_gaussian_processes}
    \begin{algorithmic}[1]
        \STATE Initialize priors $p(z | H = \varnothing)$.
        \FOR{$t=1...T$ probing iterations}
            \STATE Sample a student latent $z_t \sim p(z|H_t)$
            \STATE \textit{Optional}: Construct a train dataset $D_{\text{train}} = g(z_t)$ and pass to learner, $\theta = A(D_{\text{train}})$.
            \STATE Construct a test probe $x_{\text{test}}$ and get feedback from learner, $f(\theta, x_{\text{test}})$.
            \STATE Use feedback to update beliefs $ p(z|H)$
            \vspace{0.025in}
        \ENDFOR
        \STATE \textbf{Output} $p(z|H)$
    \end{algorithmic}
\end{algorithm}

Our method for inference is simple: 

\begin{enumerate}
    \item \textbf{Sample hypothesis}: We sample a hypothesis $z$ from our posterior distribution $p(z | H)$ according to expected improvement (EI) \citep{frazier2018tutorial}. $H$ is the history set of inputs and values given to the process.
    \item \textbf{Train the learner}: We construct a train dataset based on the sampled hypothesis, $D_{\text{train}}(z) = \{(x_{\text{train}}, y_{\text{train}})\}$. 
    We pass the dataset to the learner for it to run its learning algorithm $A$ and learn initial model parameters, $\theta=A(D_{\text{train}})$. \textit{This is only applicable in the setting where the learner trains from scratch.}
    \item \textbf{Probe for feedback}: We construct a test probe $g(z) = x_{\text{test}}$ and send it to the learner. 
    We receive feedback from the learner on that probe, $\texttt{feedback}(f_{\theta}(x_{\text{test}}))$.
    \item \textbf{Update beliefs}: We use the feedback to update our beliefs, $p(z|H, x_{\text{test}}, \texttt{feedback})$.
\end{enumerate}

This is summarized in Algorithm~\ref{alg:student_diagnosis_with_gaussian_processes}.
After a few iterations, the teacher uses $p(z|H)$ to then construct a training dataset for the student. 
We apply this framework to the ridge regression, SVMs, and offline RL domains.
What the teacher needs to infer in each domain varies, however the underlying mechanism is the same as outlined above.

After probing the student, the teacher constructs the optimal teaching dataset with the maximum a posteriori (MAP) estimate of $p(z|H)$.
The optimal teaching dataset and teaching dimension lower bound have been studied in \citet{liu2016teaching} for ridge regression and SVM with linear models.
This means that if the teacher \textit{did} have complete information about its student (ie. knowing the student's regularization parameter $\lambda$), 
the teacher can construct the optimal teaching dataset $D^*_{\text{teach}}$.
In later sections, we discuss how to exactly construct the optimal dataset in both settings. 
These domains illustrate how teaching and probing co-occur in a realistic setting where agents only have partial information about each other.
Below we discuss each of the domains in more detail and what the teacher needs to infer in each domain.

\subsection{Ridge regression \label{sec:teaching_set_ridge_regression}}

A linear model student running ridge regression (without a bias) minimizes the following objective:

\begin{equation}
    l(\textbf{w}) = \sum_{i=1}^N (y_i - \mathbf{x}_i \mathbf{w})^2 + \lambda \frac{\|\mathbf{w}\|^2}{2},
    \label{eq:ridge_regression_objective}
\end{equation}

where $\bf w$ are the linear model weights, $\bf x$ is the input, $y$ is the target, and $\lambda$ is the regularizer. 
We denote the optimal model weights to be $\bf w^*$.
From \citet{liu2016teaching}, the optimal machine teaching dataset consists of a single datapoint and is a function of the learner's regularizer $\lambda$, 

\begin{align}
    D(\lambda) = \{(x_{\text{teach}}, y_{\text{teach}})\} 
    &= \{(a \mathbf{\bf{w}}^{*}, \frac{\lambda + \| \mathbf{x}_{\text{t}} \|^2}{a})\}, \label{eq:ridge_regression_teaching_dataset}
\end{align}
where $a$ is any nonzero real number.
The inhomogeneous ridge regression case (with a bias) is also discussed in \citet{liu2016teaching}. 
We omit that setting as we use ridge regression as an illustrative example for optimal teaching with probing.
However, our analysis is easily extendable to that setting.

We denote the learner's true regularization parameter as $\lambda^*$, and the teacher's MAP estimate of the regularization parameter as $\hat{\lambda}^*$. 
To simulate a teacher not having complete information about the student, we assume that the teacher does not know the learner's true $\lambda^*$; the teacher does not have complete information about \textit{how} a student learns from data.

Concretely, the inference steps outlined in Section~\ref{sec:method} are:
\begin{enumerate}
    \item We sample a hypothesis for $\lambda \sim p(\lambda | H)$. 
    \item We construct a training dataset $D(\lambda) = (x_{\text{train}}, y_{\text{train}})$ based on the optimal teaching dataset outlined in Equation~\ref{eq:ridge_regression_teaching_dataset}. 
    \item We construct a test probe $g(\lambda) = x_{\text{test}}$ for the learner and the real learner sends back its prediction, $f_{A(D(\lambda^*))}(x_{\text{test}})$. \footnote{Empirically, we found that both randomly sampled probes or a training point from Step 2 allow for fast inference.} 
    We simulate how a learner with $\lambda$ would have responded, $f_{A(D(\lambda))}(x_{\text{test}})$.  
    We calculate the response difference, $(f_{A(D(\lambda))}(x_{\text{test}}) - f_{A(D(\lambda^*))}(x_{\text{test}}))^2$.
    \item We use the difference as the target function for the Gaussian Process. It is the feedback used to update our beliefs over $\lambda$.
\end{enumerate}

Our setting uses ordinary least squares linear regression for determining $\mathbf{w}$. 

\subsection{Support Vector Machines (SVM) \label{sec:teaching_set_svm}}

A linear SVM student minimizes the following objective:

\begin{equation}
    l(\mathbf{w}) = \sum_{i=1}^N \max(1 - y_i \mathbf{x}_i \mathbf{w}, 0) + \lambda \frac{\|\mathbf{w}\|^2}{2}
    \label{eq:svm_objective}
\end{equation}

From \citet{liu2016teaching}, the optimal teaching dataset consists of $N$ identical training items where $N = \lceil {\lambda \| \mathbf{w}^* \|^2} \rceil$. 
The teaching dataset is 
\begin{equation}
    D(\lambda) = \{(x_i, y_i)\}_{i=1}^N = \{(\frac{\lambda \mathbf{w}^*}{\lceil \lambda \| \mathbf{w}^* \|^2 \rceil}, 1)\}_{i=1}^N
    \label{eq:svm_teaching_dataset}
\end{equation}

Similar to ridge regression, the teacher does not know the learner's regularization parameter $\lambda^*$ and must infer this through interaction.
Concretely, the inference steps are:
\begin{enumerate}
    \item We sample $\lambda \sim p(\lambda | H)$. 
    \item We construct a training dataset $D(\lambda) = (x_{\text{train}}, y_{\text{train}})$ based on the optimal teaching dataset outlined in Equation~\ref{eq:svm_teaching_dataset} for SVMs.  
    \item We construct a random test probe and pass it to the learner. The SVM learner sends back the distance of that point from its separating hyperplane.
    \footnote{We also tried receiving class predictions from the learner but found this to be an extremely sparse feedback signal. 
    There might be smart ways to go about this, such as considering approaching for optimal probing. 
    Related works in this direction include optimal experiment design \citep{pronzato2008optimal, foster2019variational}.}
    We simulate what a learner with $\lambda = \hat{\lambda}$ would have responded and calculate the differences in their responses, 
    ie. $(f_{A(D(\lambda))}(x_{\text{test}}) - f_{A(D(\lambda^*))}(x_{\text{test}}))^2$.
    \item We use the difference as feedback to update our beliefs over $\lambda$.
\end{enumerate}

Our setting uses linear support vector classification for determining $\mathbf{w}$. 

\subsection{Offline reinforcement learning \label{sec:teaching_set_behavioral_cloning}}
An offline RL student updates its Q-values given a dataset of trajectories $D=\{\tau_i\}_{i=1}^{N}$ where $\tau = [s_0, a_0, r_0, ..., s_{T-1}, a_{T-1}, r_{T-1}, s_T]$ is a trajectory of states and actions: 
$Q(s_t, a_t) = r_t + \gamma * V(s_{t+1}), \quad \forall t = 0 ... T-1$ 
\footnote{If function approximators are used, then this turns into fitting a parametric model $\pi_{\theta}$ where we represent the policy $\pi$ in terms of a soft Q function $Q_{\theta}$ similar to \citet{reddy2019sqil}.}

In this domain, we assume the student can be tabula rasa or partially trained. 
An example is given in Figure~\ref{fig:example_teacher_inference_bob_knowledge}.
To the best of our knowledge, there isn't prior work in studying the teaching dimension problem in the offline RL setting. 
Nonetheless, intuitively we would like the teacher to propose trajectories that maximize coverage over the environment and minimizes overlaps with states the learner has already seen. 
The teacher is tasked to infer which states the learner has seen in order to provide demonstrations that are novel to the learner. 

The inference steps follow as,
\begin{enumerate}
    \item We sample a state $s  \sim p(s | H)$. 
    \item We do \textit{not} construct a training dataset in the behavioral cloning setting.
    \item 
    The sampled state is the test probe. We pass the sampled state $s$ to the learner and the learner sends back the action they would take, $a_{\text{learner}} = \arg\max_a Q_{\text{learner}}(s, a)$.
    The feedback is a binary function of whether the learner takes the same action as the teacher would, $\mathbbm{1}[a_{\text{learner}} = a_{\text{teacher}}]$.
    \item We use the difference as feedback to update our beliefs over $\mathcal{S}$.
\end{enumerate}

Our method uses $\epsilon$-greedy Q-learning.
Note that in the RL setting, states are used for diagnosing the student but the teacher ultimately teaches in the form of demonstrated trajectories.
An open question that we hope to later pursue is the form of probe and teaching, and how this impacts the choice of target function (Step 3).

\section{Experiments \label{experiments}}

\begin{figure}
    \centering
    \newcommand{\factor}{0.20}
    \newcommand{\extraspace}{0.5em}
    \subfloat[EmptyRoom 7x7]{\includegraphics[width=\factor\textwidth]{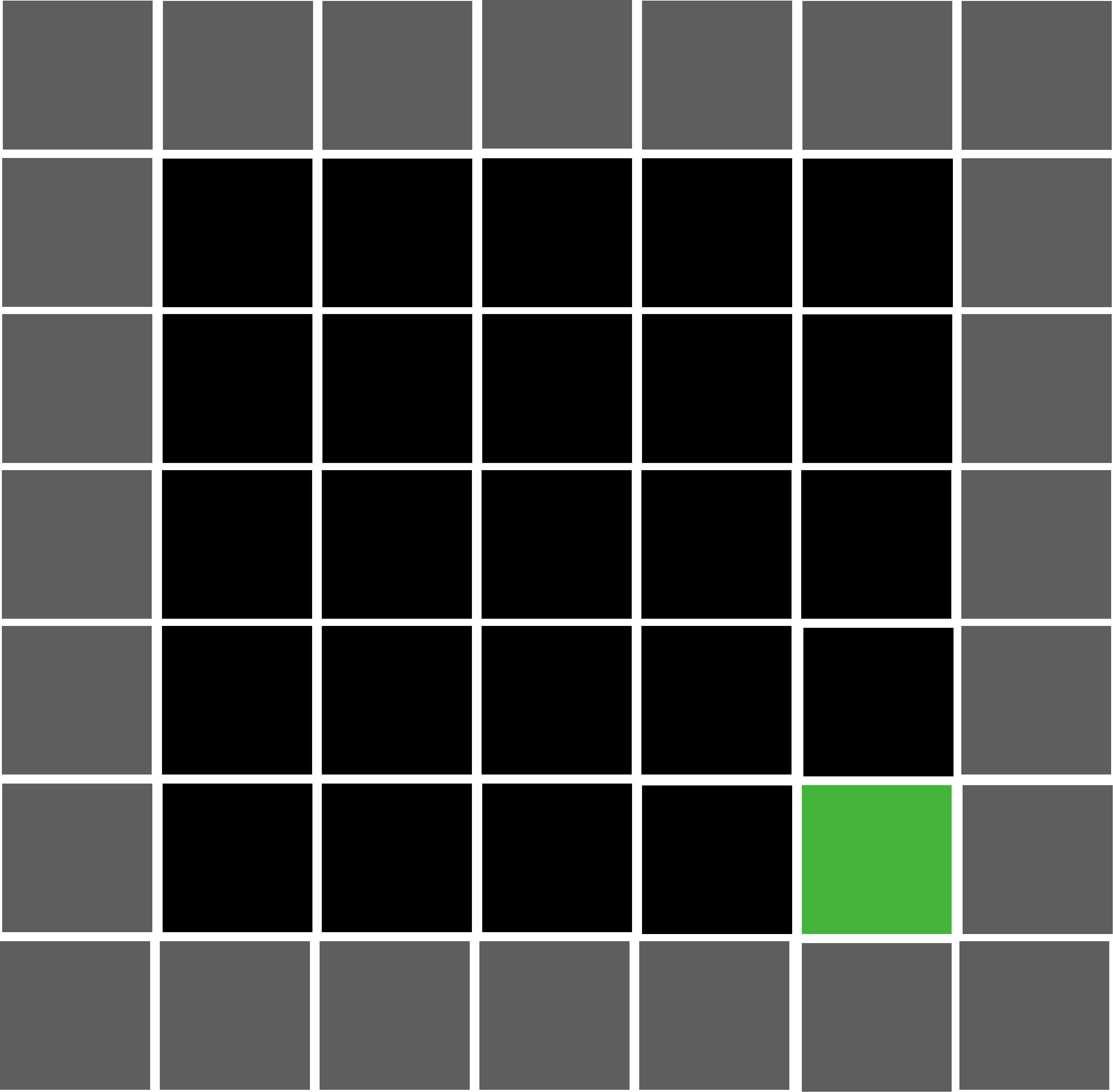}}
    \hspace{\extraspace}
    \subfloat[EmptyRoom 9x9]{\includegraphics[width=\factor\textwidth]{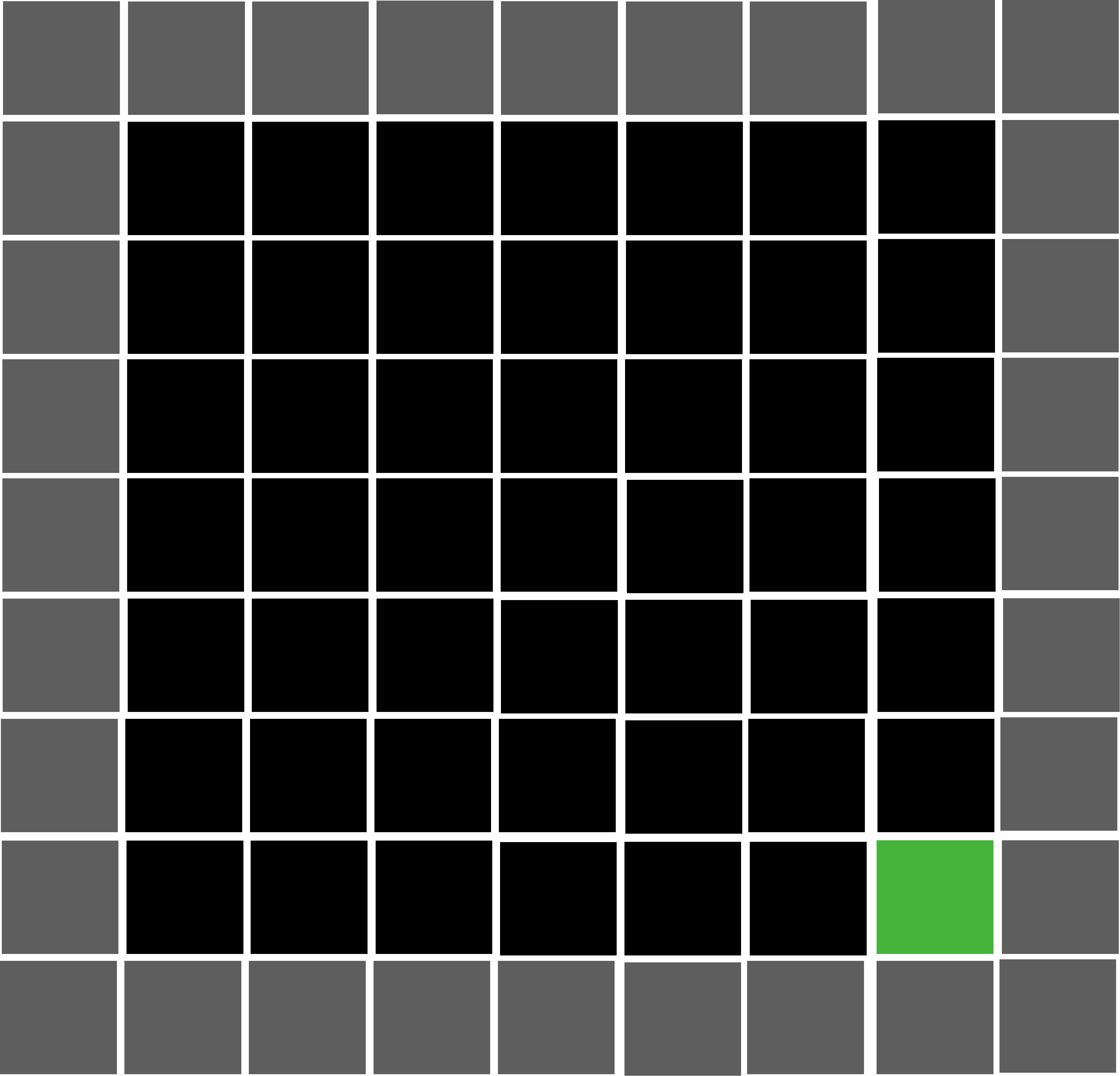}}
    \hspace{\extraspace}
    \subfloat[4 Rooms 9x9]{\includegraphics[width=\factor\textwidth]{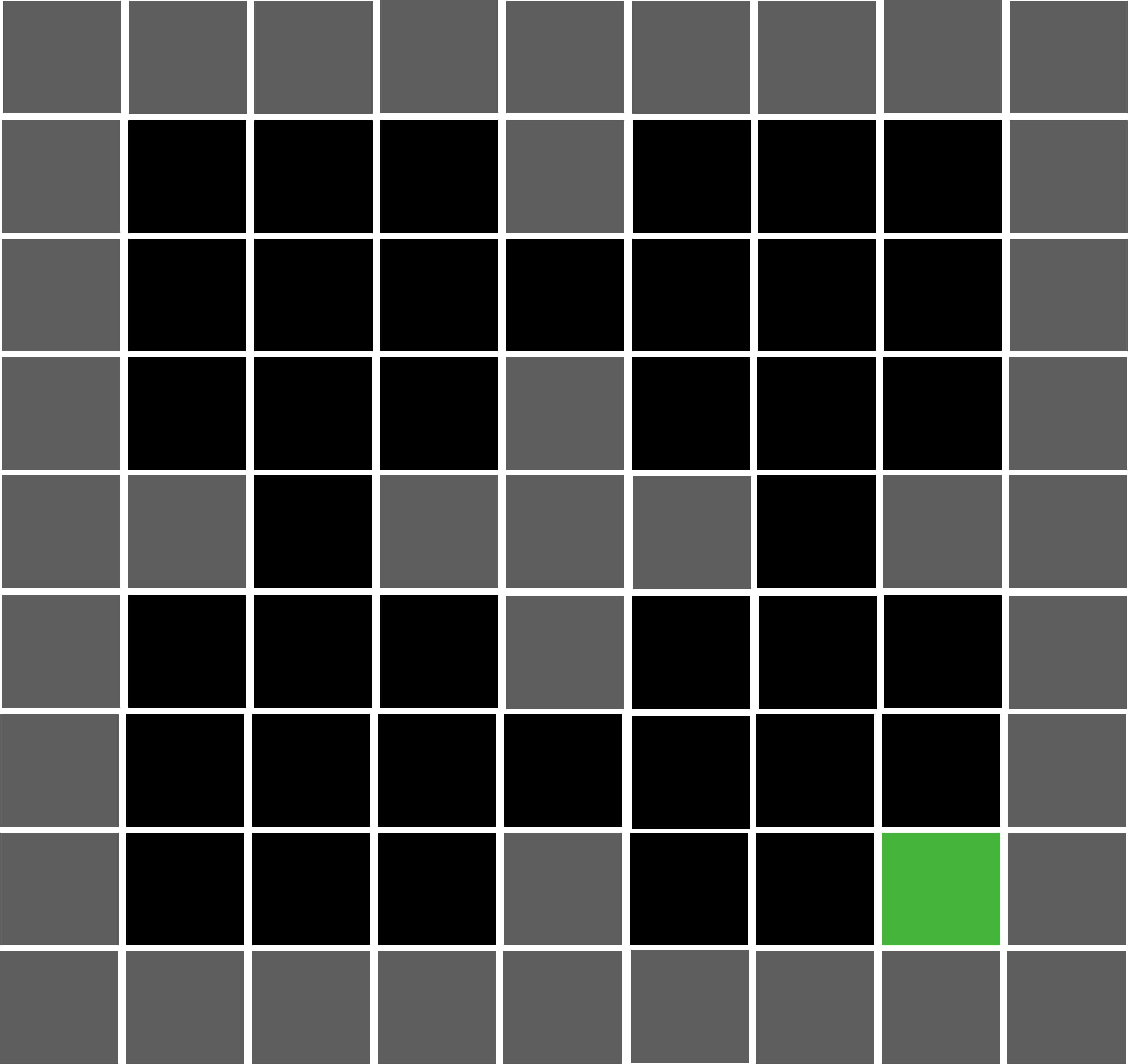}}
    \hspace{\extraspace}
    \subfloat[4 Rooms 11x11]{\includegraphics[width=\factor\textwidth]{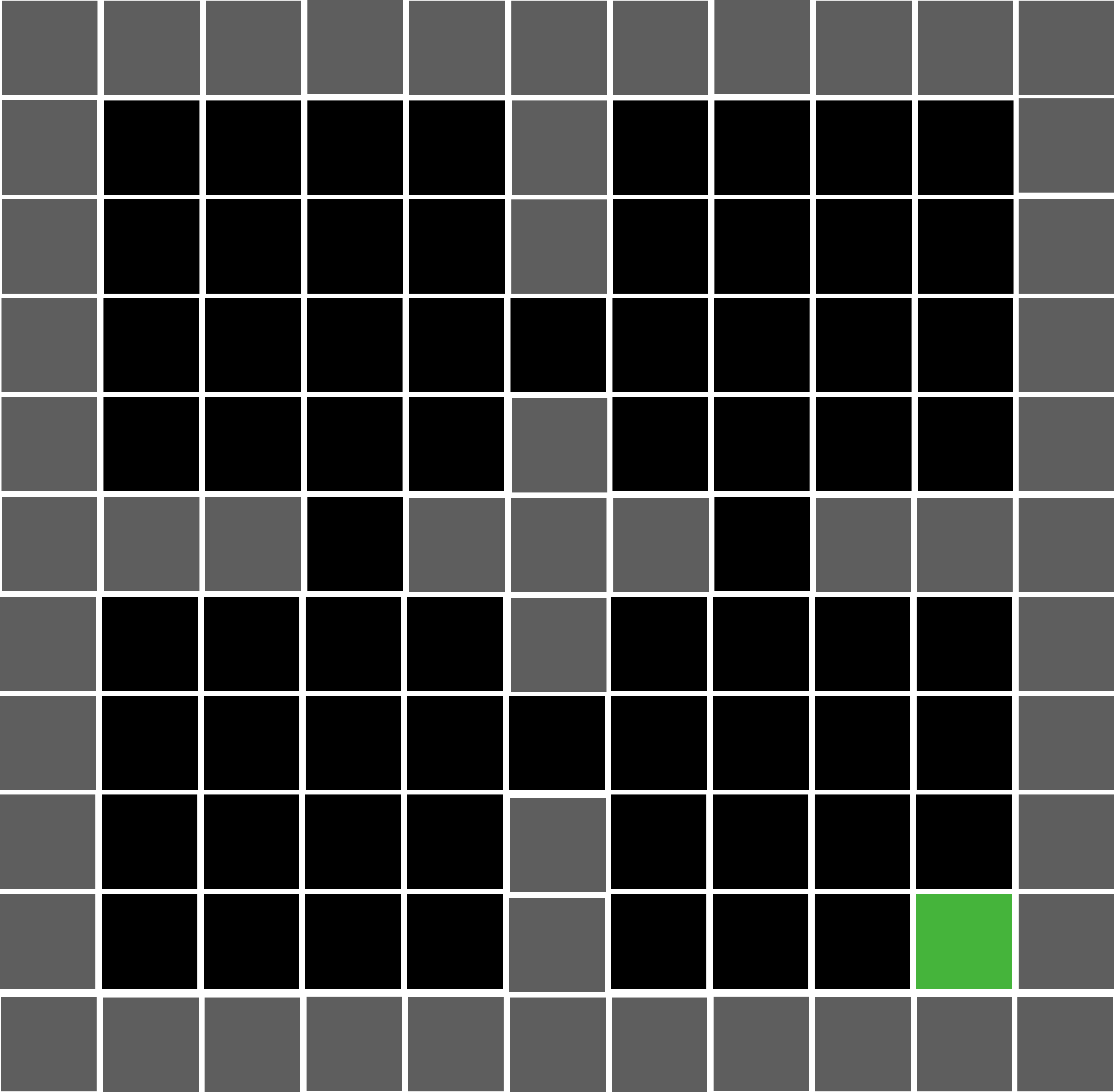}}
    \caption{Environments used for the RL setting. The learner can be initialized anywhere in the environment. Its goal is to navigate to the green grid cell.}
    \label{fig:gridworld_environments}
\end{figure}

We now evaluate the utility of understanding the learner before teaching and compare different teaching methods against passive learning (learning without a teacher and just from the original training dataset). 
In particular, we try to answer the following  research questions (\textbf{RQ}s):

\begin{description}
    \item \textbf{RQ1}: How important is it to understand the learner when teaching? 
    \item \textbf{RQ2}: When would you prefer teaching with diagnosing over passive learning? 
\end{description}

We examine this in three settings: Ridge regression, classification with support vector machines and offline RL in sequential decision making settings. 
The ridge regression and classification settings are meant to be simple, illustrative examples of how probing would be combined with optimal teaching, 
when the optimal teaching dataset can be determined with full knowledge of the student \citet{liu2016teaching}.
The regression and two-way separable classification data are generated from \citet{scikit-learn} without added noise.
We use the MiniGrid environments \citep{gym_minigrid} for the RL setting; examples of the environments are shown in Figure~\ref{fig:gridworld_environments}.

\paragraph{Methods}
Our main method first diagnoses the student with expected improvement (EI) as the acquisition function, then teaches the student. 
We denote our method as \textbf{D(EI)+T}. 
We compare three alternatives to our method. 
One is to diagnose with random sampling as the acquisition function, then teach the student; this is denoted as \textbf{D(R)+T}. 
Two is to teach randomly by sampling from the prior $p(z)$ and immediately teach the student; we denote this as \textbf{RT}. 
For instance, this would mean that in ridge regression, the teacher samples a random $\lambda$ and constructs a teaching dataset according to Equation~\ref{eq:ridge_regression_teaching_dataset} without interacting with the student. 
Three is to passive learn without a teacher, \textbf{PL}. 
In ridge regression and SVMs, this means to passively learn over a fixed training dataset, and in the offline RL setting, this means the student collects and learns from its own data. 
For both methods \textbf{D(EI)+T} and \textbf{D(R)+T} where diagnosis is run, we allow the teacher to probe the student 5 times, ie. $T=5$ in Algorithm~\ref{alg:student_diagnosis_with_gaussian_processes}.

\subsection{Ridge regression}
\begin{figure}[t]
    \centering
    \newcommand{\factor}{0.33}
    \subfloat[MSE (1D) \label{fig:metrics_diagnosis_feature_dim_1}]{\includegraphics[width=\factor\textwidth]{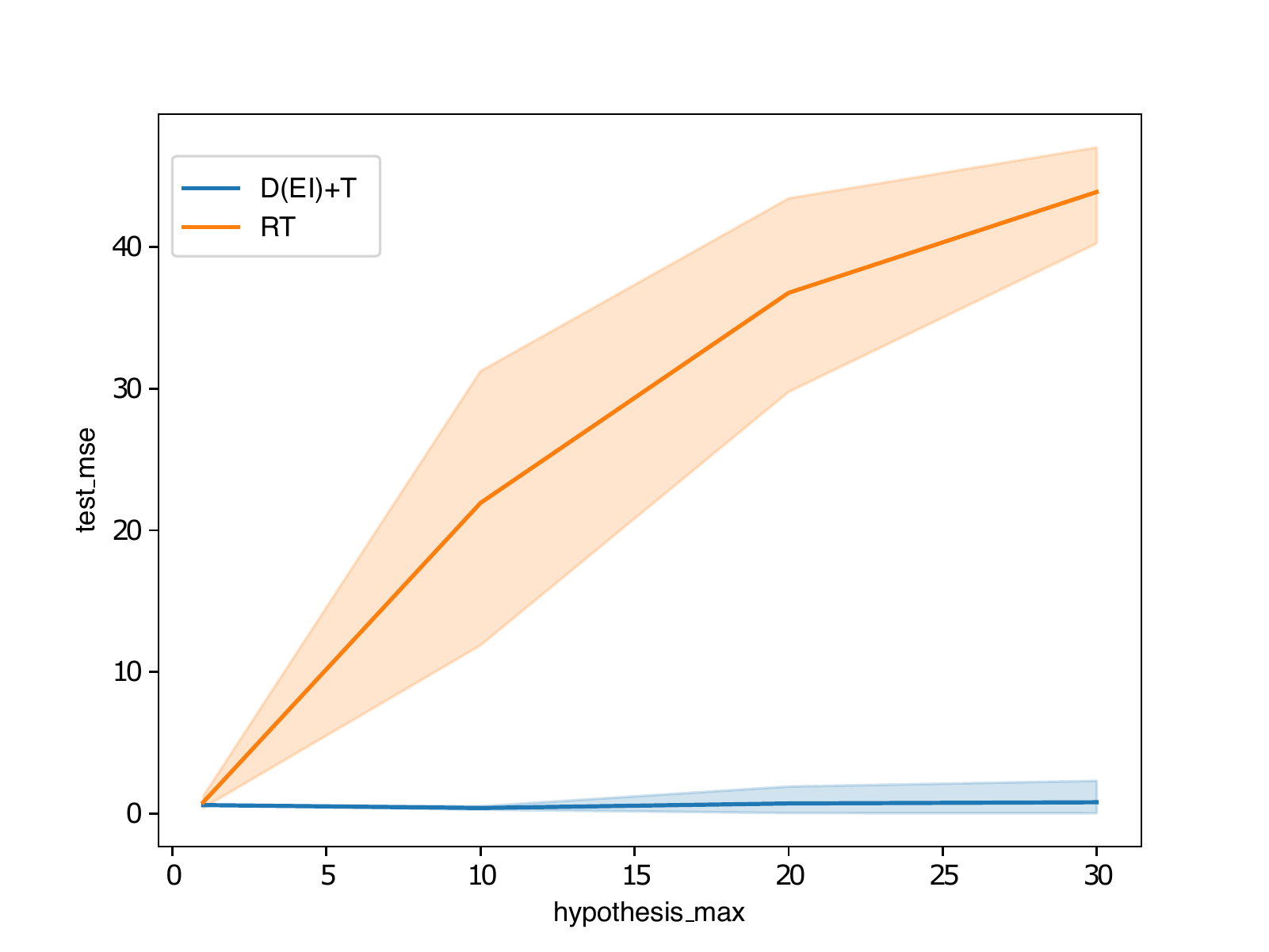}}
    \subfloat[MSE (2D) \label{fig:metrics_diagnosis_feature_dim_2}]{\includegraphics[width=\factor\textwidth]{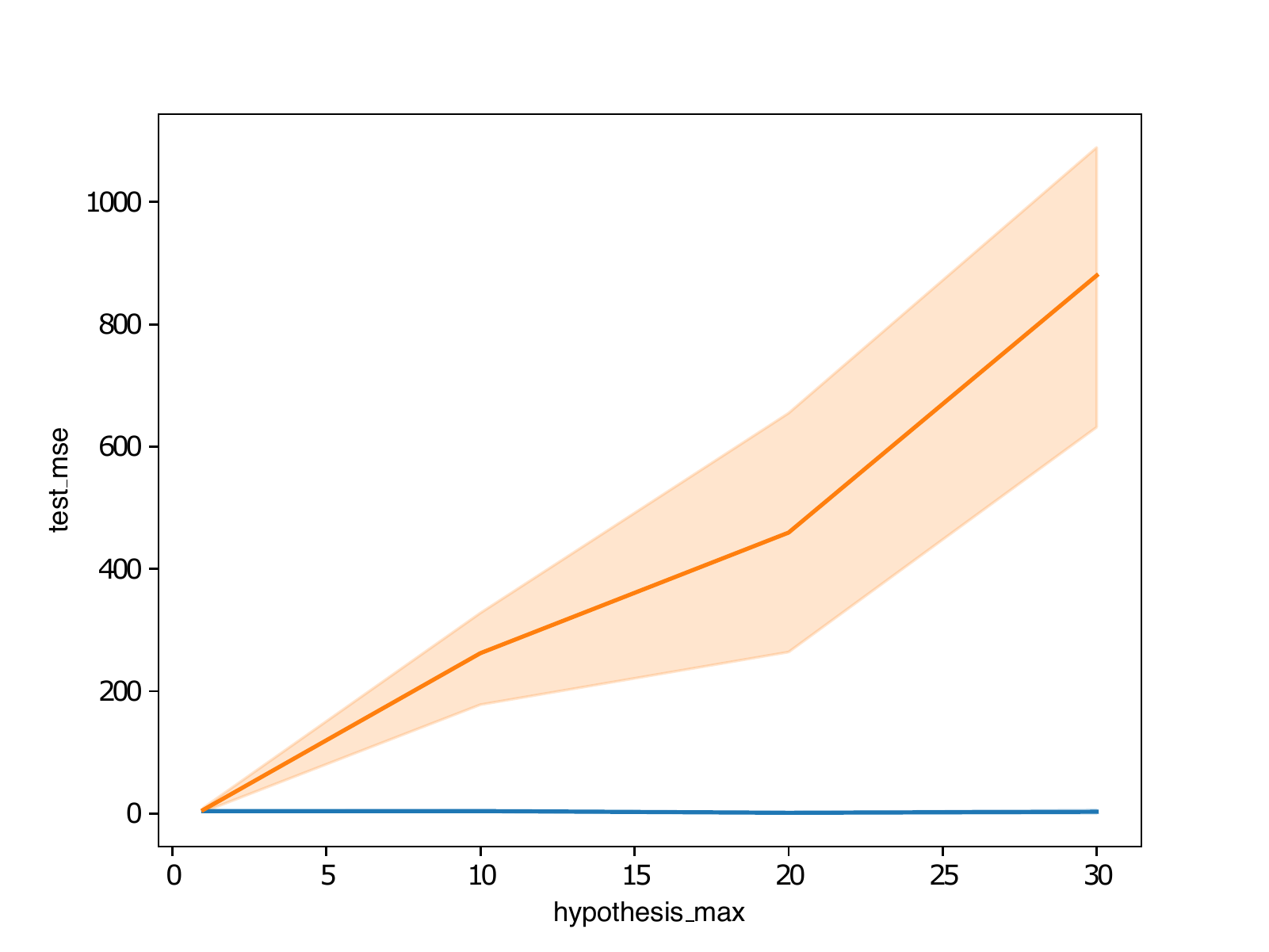}}
    \subfloat[MSE (3D) \label{fig:metrics_diagnosis_feature_dim_3}]{\includegraphics[width=\factor\textwidth]{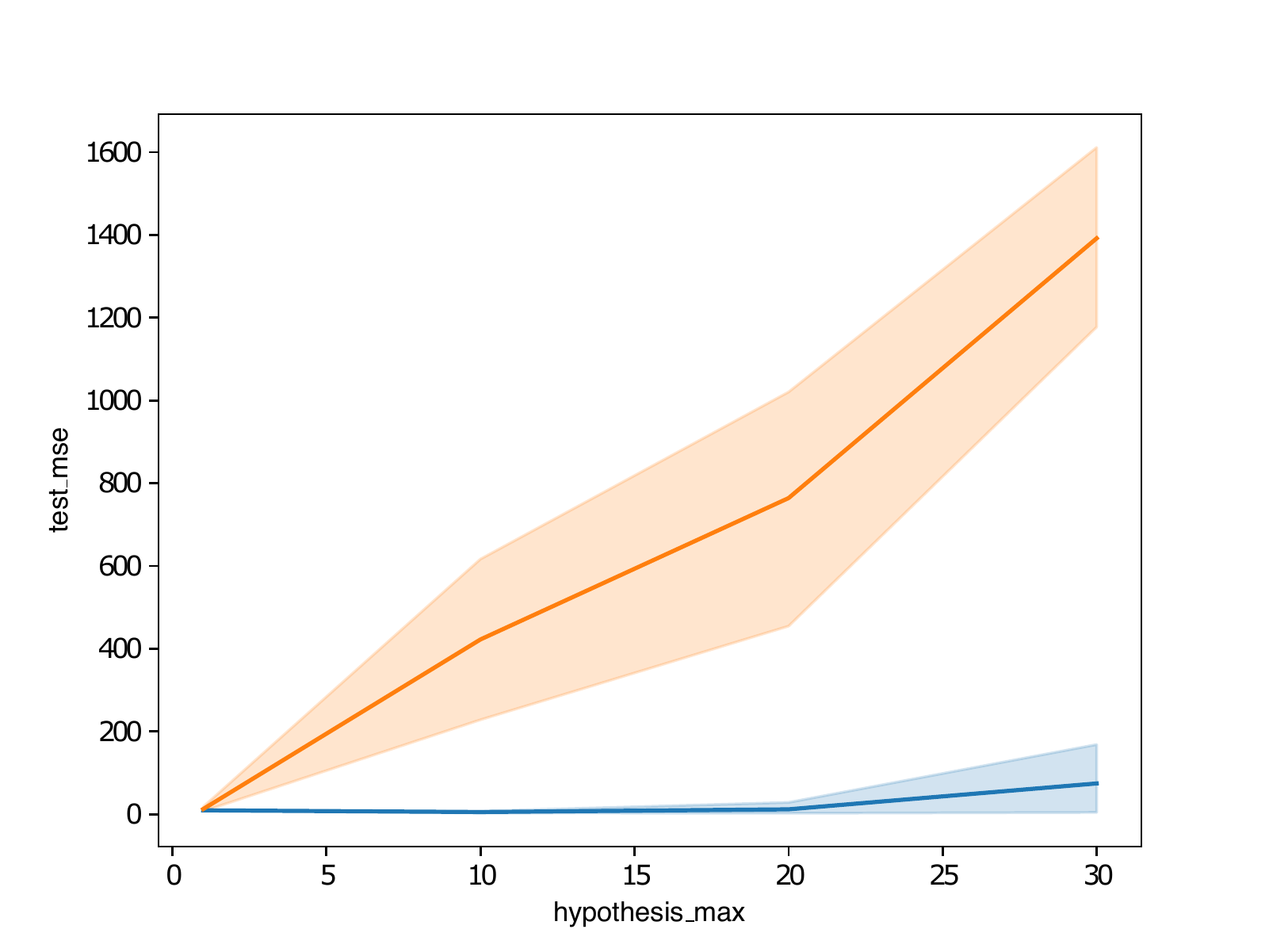}}
    \caption{Mean squared error on held-out regression dataset over 10 runs. Proper diagnosis of the student matters as the hypothesis space (x-axis) over possible student models grows and when the number of feature dimensions grows (a-c). }
    \label{fig:metrics_diagnosis_of_student_parameters_feature_dimensions}
\end{figure}

We compare the accuracy and sample efficiency (number of training examples) across the methods. 
Figure~\ref{fig:metrics_diagnosis_of_student_parameters_feature_dimensions} compares \randomteachabbr~ and \diagnosiseiteachabbr~ to answer \textbf{RQ1}: 
How important is it for the teacher to learn about the student before teaching?
We compare a teacher that randomly samples across an increasing hypothesis space for $\lambda$ compared to a teacher that probes the student. 
We see that randomly guessing information about the student can lead to arbitrarily bad performance for the learner; 
these observations hold also across increasing feature dimension.
With only 5 probes, \diagnosiseiteachabbr~ allows the student to predict with minimal error. 
Thus, to answer \textbf{RQ1}: 
It's important for the teacher to learn about the student, otherwise the student can perform poorly under larger belief spaces and feature spaces.

\begin{figure*}
    \centering
    \newcommand{\factor}{0.45}
    \begin{subfigure}[b]{\factor\textwidth}
        \centering
        \includegraphics[width=\textwidth]{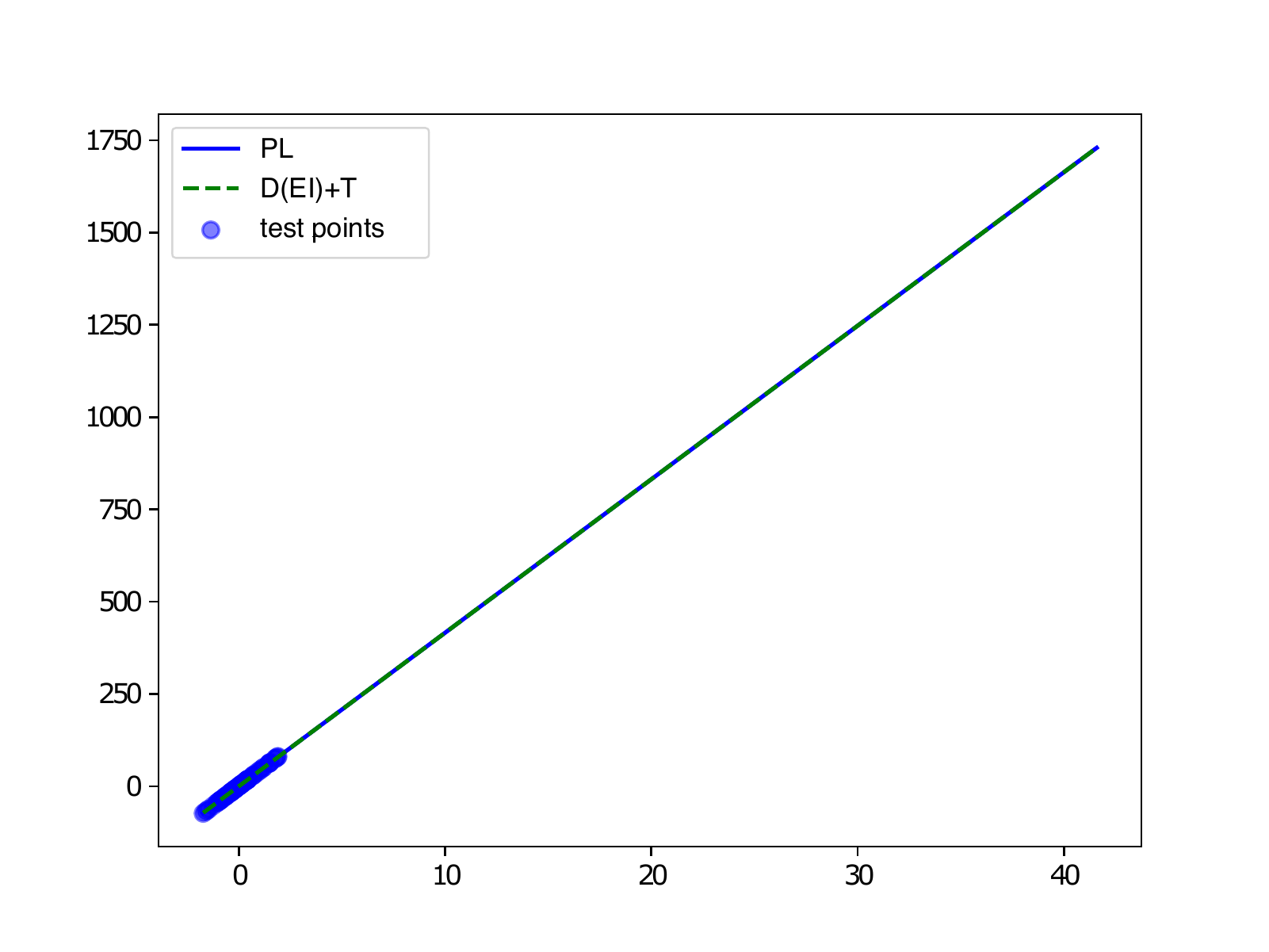}
        \caption[$\lambda=1$]%
        {{\small $\lambda=1$}} 
    \end{subfigure}
    \hfill
    \begin{subfigure}[b]{\factor\textwidth}  
        \centering 
        \includegraphics[width=\textwidth]{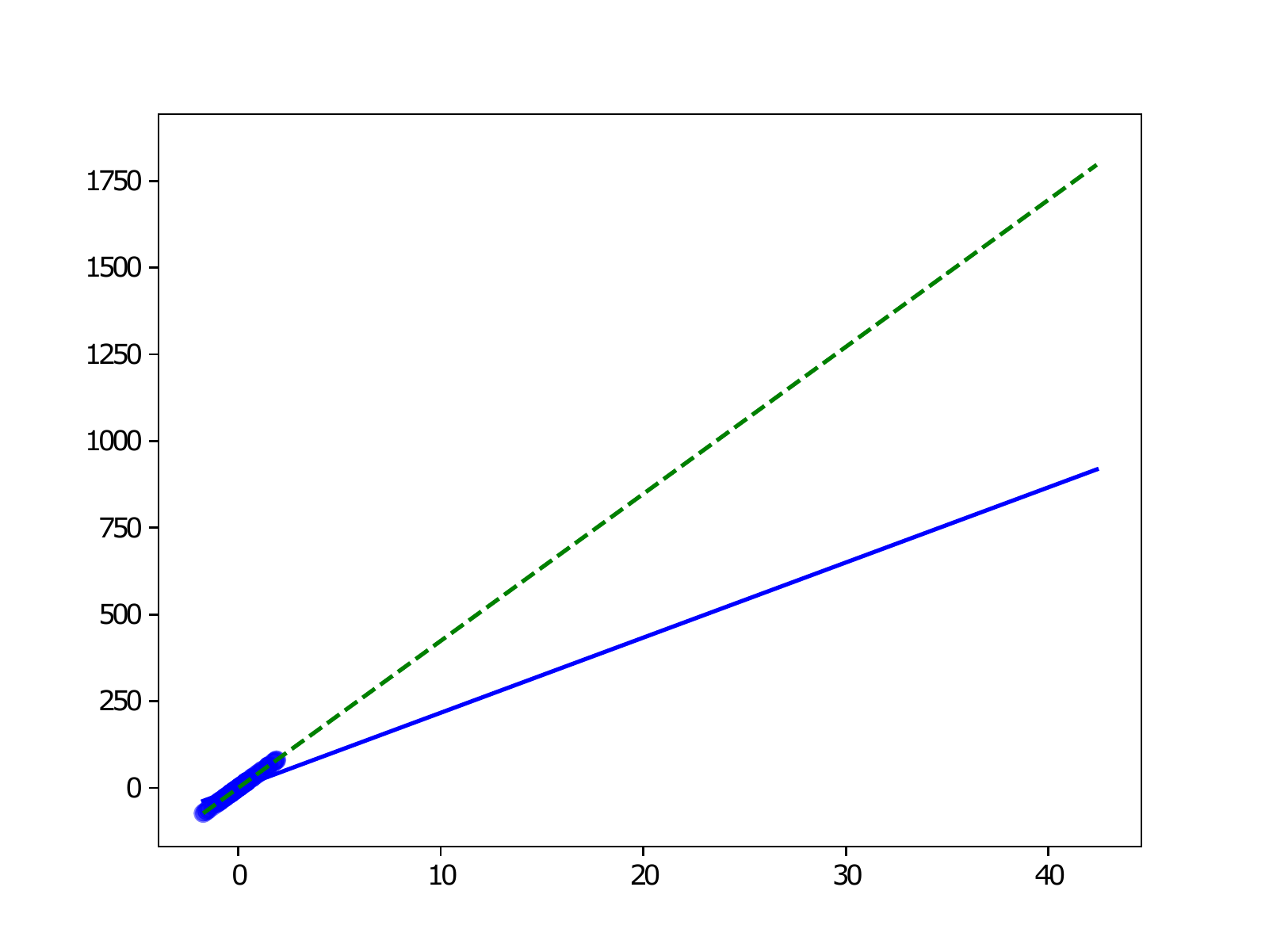}
        \caption[$\lambda=50$]%
        {{\small $\lambda=50$}}
    \end{subfigure}
    \vskip\baselineskip
    \begin{subfigure}[b]{\factor\textwidth}   
        \centering 
        \includegraphics[width=\textwidth]{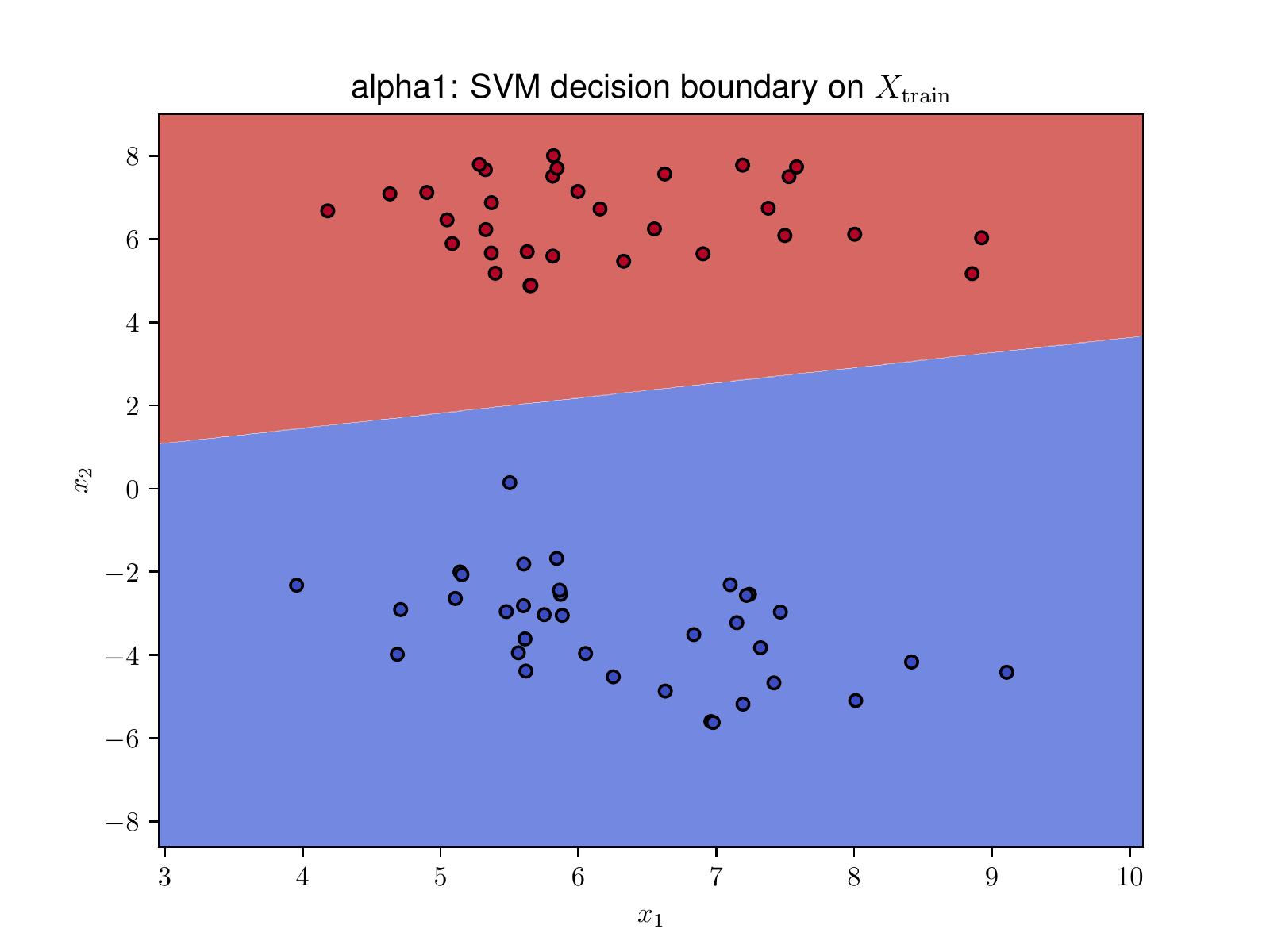}
        \caption[$\lambda=1$]%
        {{\small $\lambda=1$}}
        \label{fig:svm_decision_boundary_lambda_1.0}
    \end{subfigure}
    \hfill
    \begin{subfigure}[b]{\factor\textwidth}   
        \centering 
        \includegraphics[width=\textwidth]{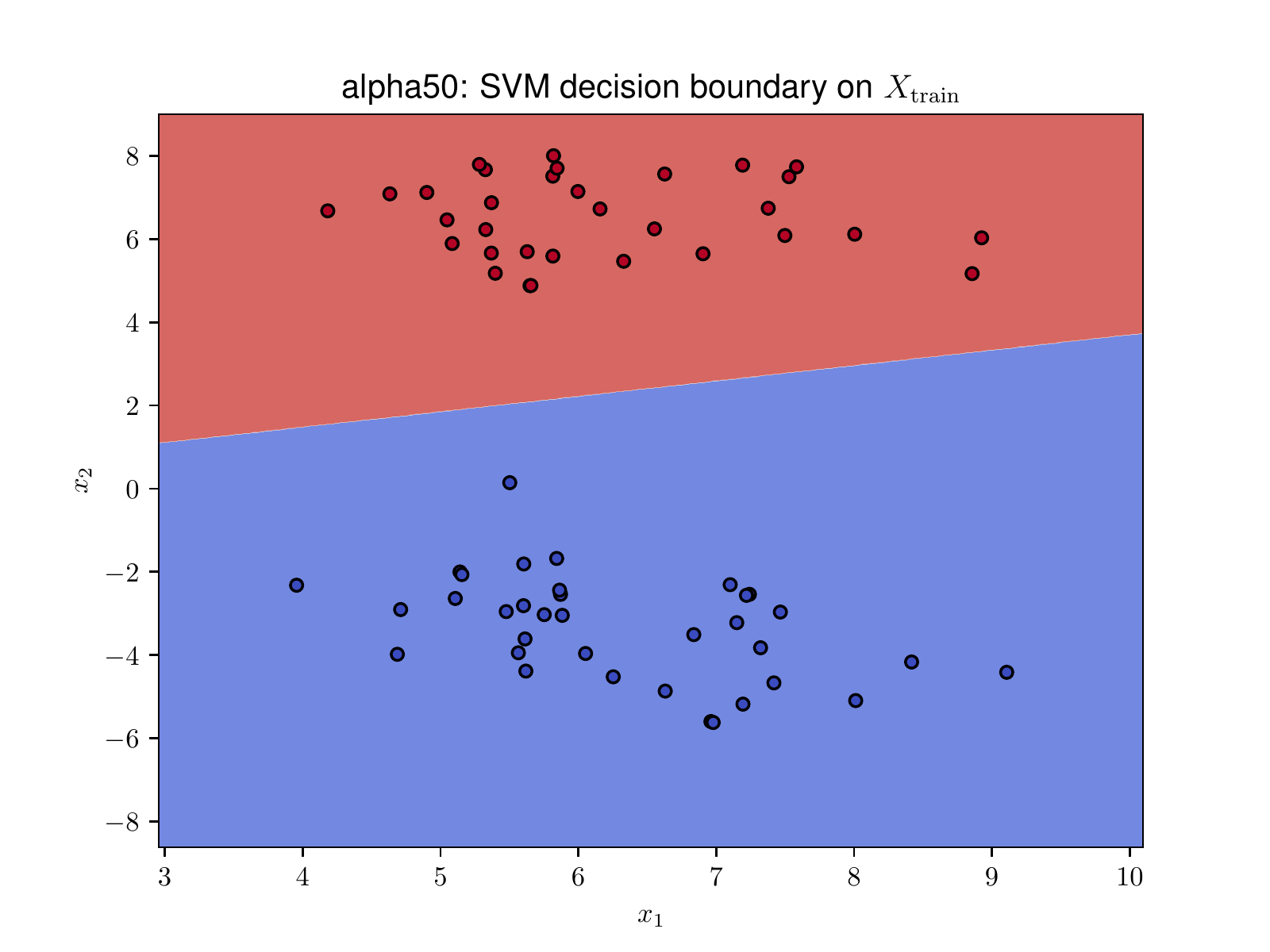}
        \caption[$\lambda=50$]%
        {{\small $\lambda=50$}}
        \label{fig:svm_inferring_lambda_not_as_important}
    \end{subfigure}
    \caption[ ]
    {\small (a-b) illustrate a 2D regression task where the x- and y-axes correspond to the two coordinates. The first image is where the learners have a low regularization parameter ($\lambda=1$), and the second image is where the learners have a high regularization parameter ($\lambda=50$). The weights from two learners are plotted as lines. One is a learner trained on a teaching dataset (dashed, green line) by \diagnosiseiteachabbr. Another is the \passivelearningabbr~ learner (full, blue line) trained on a \textit{fixed} training regression dataset. 
    The plotted points in blue are points from the test dataset. 
    The \diagnosiseiteachabbr~ learner always finds the best fit model (ie. recovers $\theta^*$) whereas the \passivelearningabbr~ learner does not in the high regularization region. (c-d) however show that this is not the case for the SVM learners: both learners learn to separate the classes accurately.} 
    \label{fig:nonIID_vs_IID_recovery_theta}
\end{figure*}

Figure~\ref{fig:nonIID_vs_IID_recovery_theta} compares \passivelearningabbr~ and \diagnosiseiteachabbr~ to answer \textbf{RQ2}:
When would you prefer teaching with diagnosing over passive learning? 
With a fixed training dataset size $N=100$, we vary the learner's regularization parameter from $\lambda=1.0$ to $\lambda=50.0$. 
We see that with particular learners (ie. low regularization), both methods do comparably in Figure~\ref{fig:svm_decision_boundary_lambda_1.0}. 
However, with other learners (ie. high regularization), \passivelearningabbr~ is not able to recover the best model. 
A teacher who diagnoses and then teaches, on the other hand, does enable the student to recover the best model. 
To answer \textbf{RQ2}: 
These results suggest teaching is preferable in low-data settings where sampling from the training distribution is expensive; 
the setting doesn't easily allow for more data to remedy the effects of high regularization or peculiarities in the student's algorithm.

\subsection{Support Vector Machine}

We conducted the same experiments as in ridge regression for the two-way separable classification with SVMs. 
However, we found separable classification settings with SVMs unable to yield differences among \passivelearningabbr, \randomteachabbr, \diagnosiseiteachabbr: 
all three methods learn to separate the data within a reasonable range over $\lambda$. 
Figure~\ref{fig:svm_inferring_lambda_not_as_important} provides intuition into why the three methods perform similarly on the classification domain. 
The separating hyperplane is not sensitive  to increasing $\lambda$ values. 
Therefore, when the classes are easy to separate, learners from each method will classify the points the same way.
We extended these experiments to non-separable classes, however results were inconclusive as they required hand-designing the degree of overlap between classes.

\subsection{Offline reinforcement learning}

\begin{figure}
    \centering
    \newcommand{\factor}{0.50}
    \subfloat[\label{fig:rl_teacher_sample_efficiency}]{\includegraphics[width=\factor\textwidth]{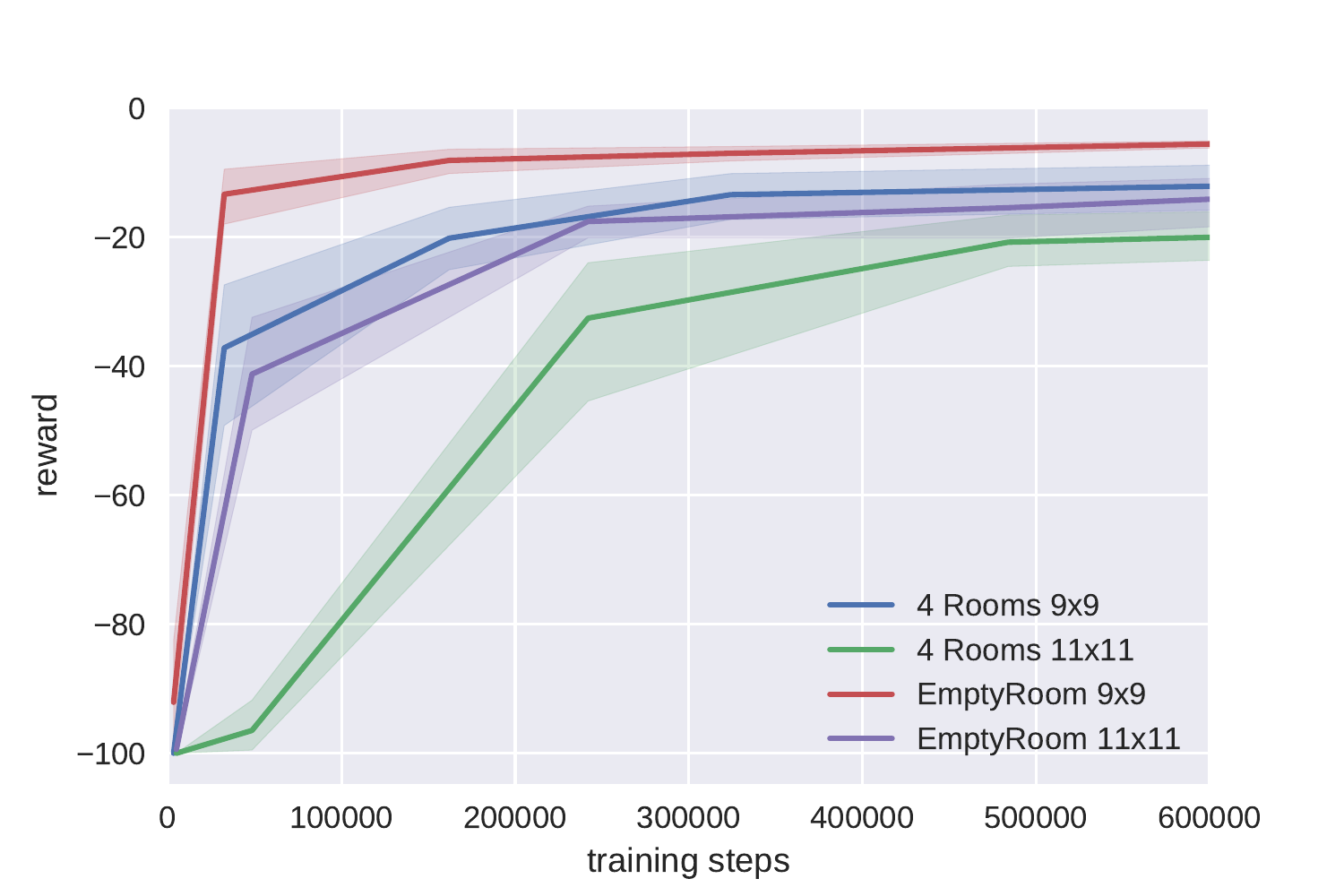}} 
    \subfloat[]{\includegraphics[width=\factor\textwidth]{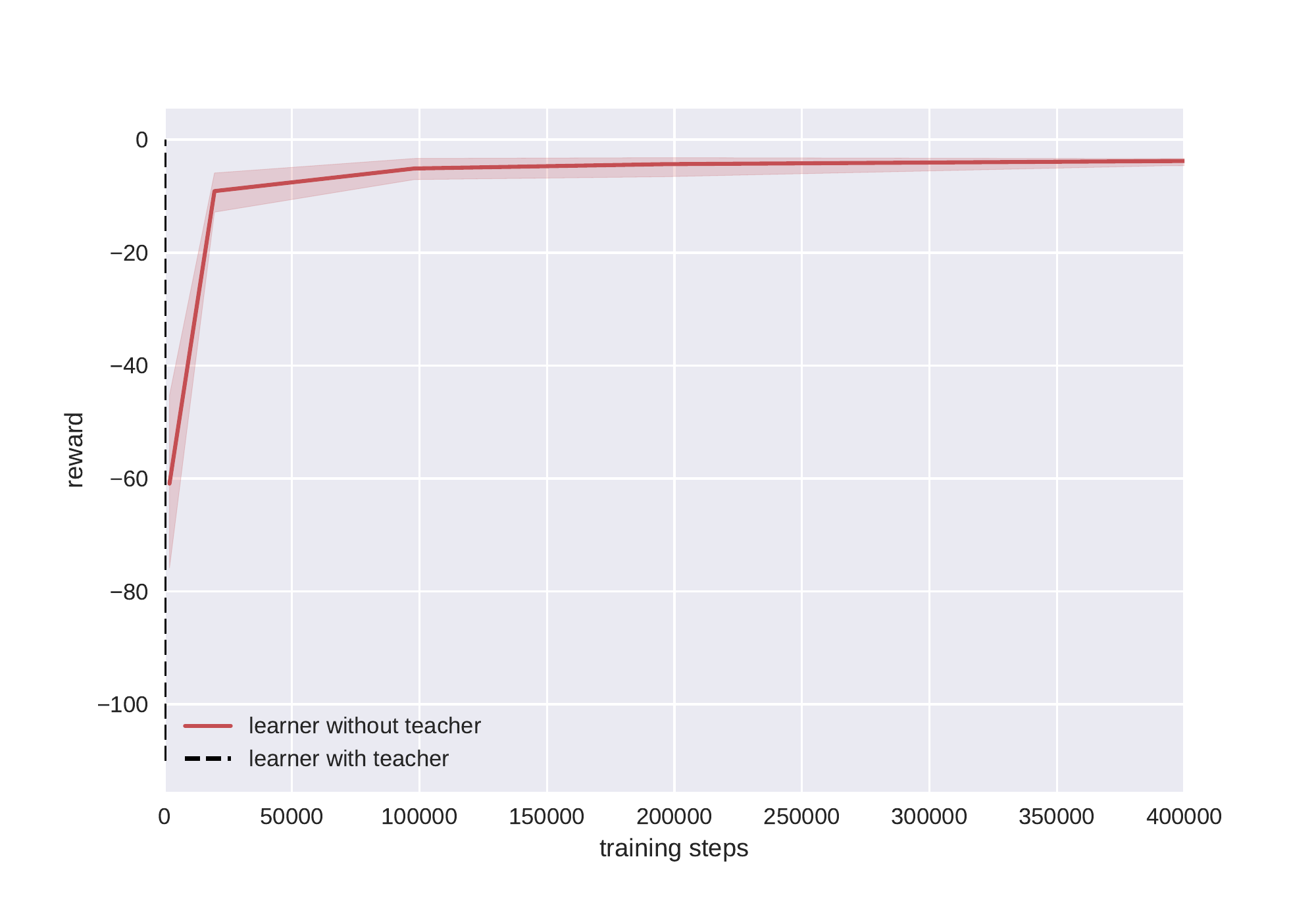}}
    \caption{Comparing the number of training samples needed for a learner with vs. without a teacher. 
    (a) shows the number of samples needed a learner without a teacher across environments.
    (b) compares the number of samples needed by a learner without a teacher (red curve) and with a (black horizontal line $x \approx 50$) on the EmptyRoom 7x7 environment: with the help of a teacher sending demonstrations, the learner can learn much more quickly.
    All graphs were generated on 10 runs.
    }
    \label{fig:single_agent_sample_efficiency}
\end{figure}

There are four environments we use for the RL setting; 
examples of the environments are shown in Figure~\ref{fig:gridworld_environments}. 
The learner can be initialized anywhere in the environment.  
The learner's goal is to navigate to the green cell.
The reward for this setting is sparse: 
The learner receives $-1$ with every step except for when it reaches the goal where it receives $0$ as its reward.
The learner is evaluated based on how well it's able to navigate to the goal from any initial state. 
The teacher is trained until convergence by running $\epsilon$-greedy Q-learning with $\epsilon=0.3$. 
for 5000 episodes, each episode with 100 steps. The teacher learns the optimal policy at convergence. The learning curves of the teacher are shown in Figure~\ref{fig:rl_teacher_sample_efficiency}.

We assume different levels of student knowledge: Either the student has no experience (it has never interacted with the environment), it knows one optimal path, it knows a few optimal paths, or it knows all optimal paths to the goal. 
An example is illustrated in Table~\ref{table:gridworld7x7} on the left-hand column. 
The goal of the teacher is to then infer which states the student can perform well in, and ideally construct a teaching dataset of trajectories where the student would perform poorly.

We compare four different methods. 
One is \passivelearningabbr~ where the passive learner runs standard $\epsilon$-greedy Q-learning without the teacher.
Two is \randomteachabbr~ where the teacher randomly samples a state from the environment and executes its policy until it reaches the goal; this is the trajectory that is sent over per sample. 
Three is \diagnosisrteachabbr~ where the teacher randomly samples a state for probing when approximating the target function for the Gaussian process. 
It then picks states that maximizes the target function as the initial states for running its policy and collecting trajectories.
Four is \diagnosiseiteachabbr~ where the teacher samples a state with the highest expected improvement. 
Similarly to \diagnosisrteachabbr, \diagnosiseiteachabbr~ picks the states that maximize the target function as the initial states for collecting trajectories.

To answer \textbf{RQ1}, we compare against \randomteachabbr~ in Table~\ref{table:gridworld7x7} for EmptyRoom 7x7; 
results for the other environments can be found in the appendix.
The first takeaway is that probing methods like \diagnosisrteachabbr~ and \diagnosiseiteachabbr~ both teach better than \randomteachabbr~ (middle column). 
Additionally, teachers that probe students avoid sending repeated states which the learner has already seen before (right-hand column). 
Interestingly, \diagnosisrteachabbr~ tends to perform better or comparably to \diagnosiseiteachabbr.  
This suggests that that the target function we use (action matching) might not be ideal for determining states to probe a student with. 

To answer \textbf{RQ2}, we compare \passivelearningabbr~ with \diagnosiseiteachabbr~ in Figure~\ref{fig:single_agent_sample_efficiency}.
Noticeably, the student can benefit a lot from having a teacher provide demonstrations as it's able to learn from much fewer samples and still achieve high performance.
Thus, a benefit to having a teacher is avoiding inefficient exploration.

\section{Conclusion and future work}
% This kind of collective learning setting contrasts with most machine learning settings which considers the learner in isolation of others or treats the teacher as a fixed oracle \citep{settles_active_nodate} (active learning). 
Our work showcases three simple settings which highlight the utility of having a teacher first learn about its student before teaching. 
We believe this form of collective learning---algorithmically constructing diagnostics for the learner in order to construct model-tailored training---can pave principled ways for diagnosing models and proposing remedies. 

There are several interesting questions which we hope to later explore. These include, 

\begin{itemize}
    \item What is an optimal target function (form of feedback) to use for the Gaussian Process? 
    The offline RL experiments suggest that we should revisit this choice. 
    They also suggest that if a reasonably generic target function can be found in the RL setting, training RL policies can be made in a more sample-efficient and targeted way.
    A sensible, alternative target function could be the difference in value function between the student and teacher.
    However, one key limitation is that the teacher can access the (near-)optimal value function which is challenging to obtain in the first place.
    \item Under what conditions would you prefer machine teaching over passive learning?
    We'd like to formalize this as a function of the training set and problem difficulty.
    Under which domains (or types of domains) would machine teaching be useful for?  
    \item How should we sample from a desirable data distribution given what we know about the student? 
\end{itemize}

\subsubsection*{Acknowledgments}
REW is supported by the National Science Foundation Graduate Research Fellowship. 
The authors would give special thanks to Dorsa Sadigh, Willie Neiswanger, Emma Brunskill, Tatsunori Hashimoto, Gregory Valiant, Andy Shih, Gabriel Poesia, and Rohith Kuditipudi for their helpful discussions. 
The authors would also like to thank Sidd Karamcheti, Jenn Grannen, and the anonymous reviewers for their feedback on the paper.

\bibliography{iclr2022_conference}
\bibliographystyle{iclr2022_conference}

\appendix
\section{Appendix}

\begin{center} 
    \centering
    \newcommand{\factor}{0.20}
    \newcommand{\graphfactor}{0.33}
    \newcommand{\graphf}{0.40}
    \begin{table} 
    \begin{tabular}{c | c | c} 
    \toprule
    \bf Student experience & \bf Performance  & \bf \% overlap \\
    \midrule\midrule
    \includegraphics[width=0.25\textwidth]{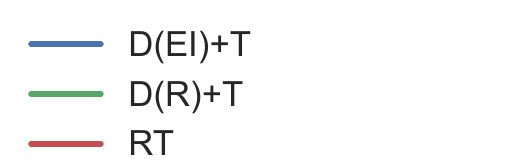}  \\ \includegraphics[width=\factor\textwidth]{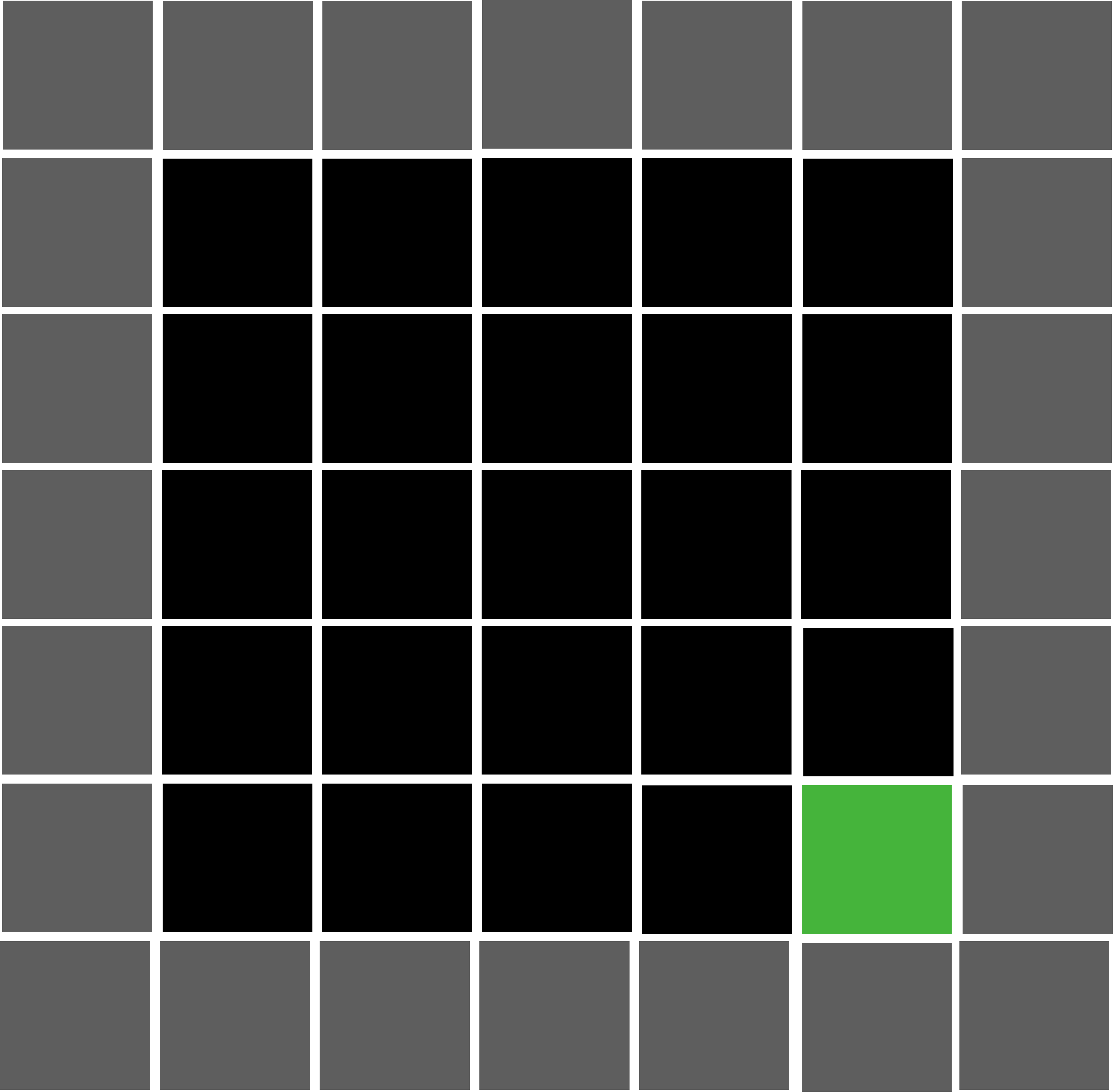} & \includegraphics[width=\graphfactor\textwidth]{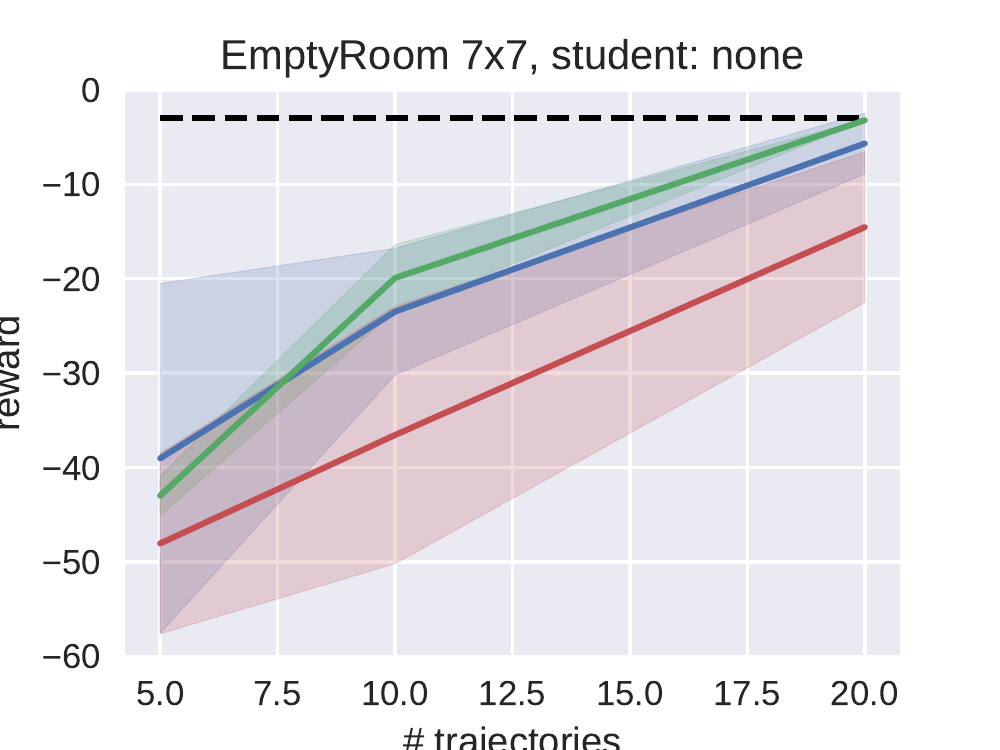} & \includegraphics[width=\graphf\textwidth]{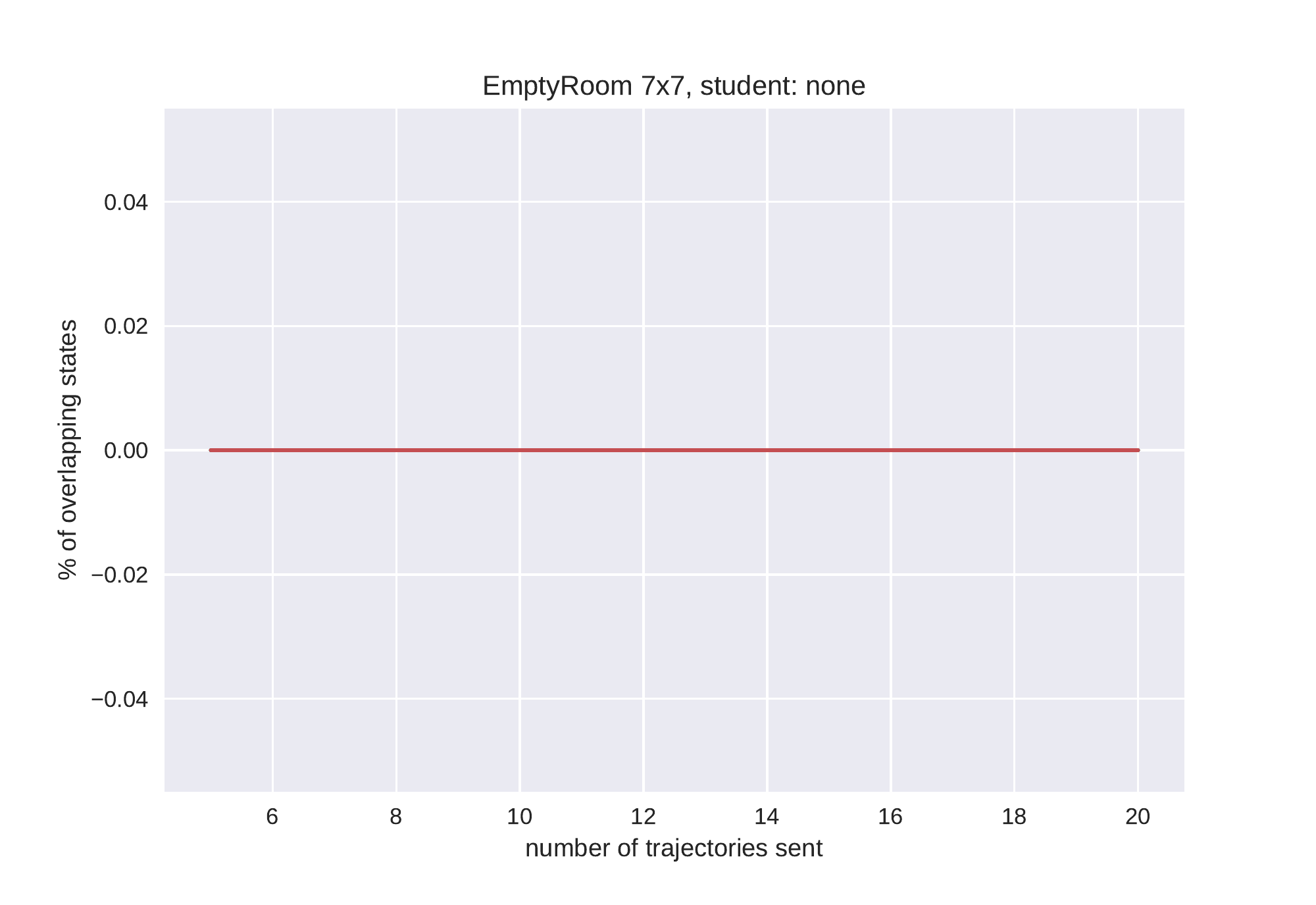}
     \\
    \midrule
    \includegraphics[width=\factor\textwidth]{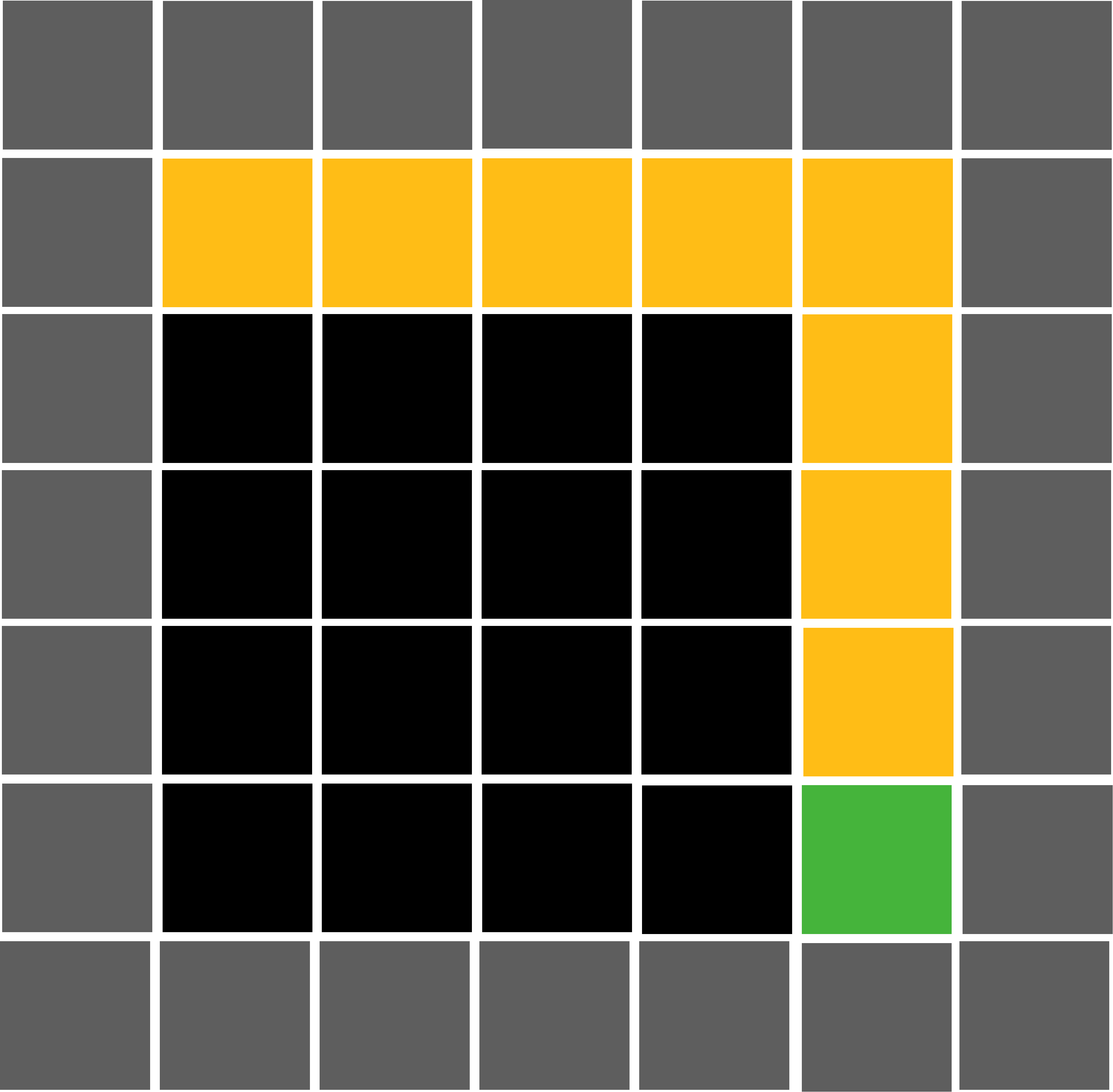} & \includegraphics[width=\graphfactor\textwidth]{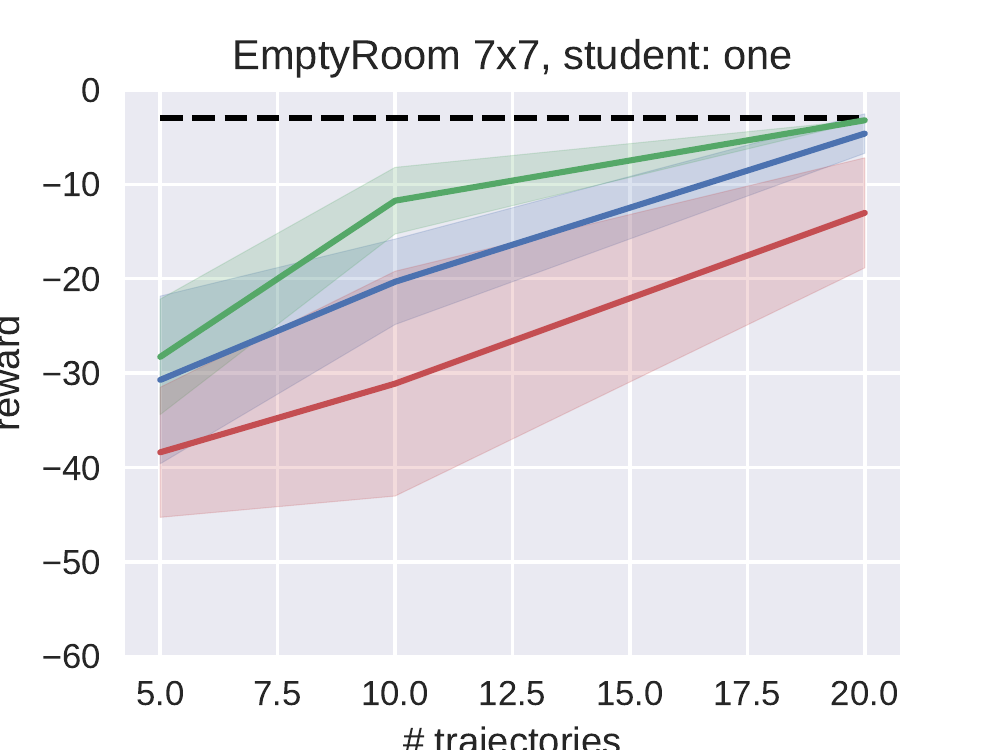} & \includegraphics[width=\graphf\textwidth]{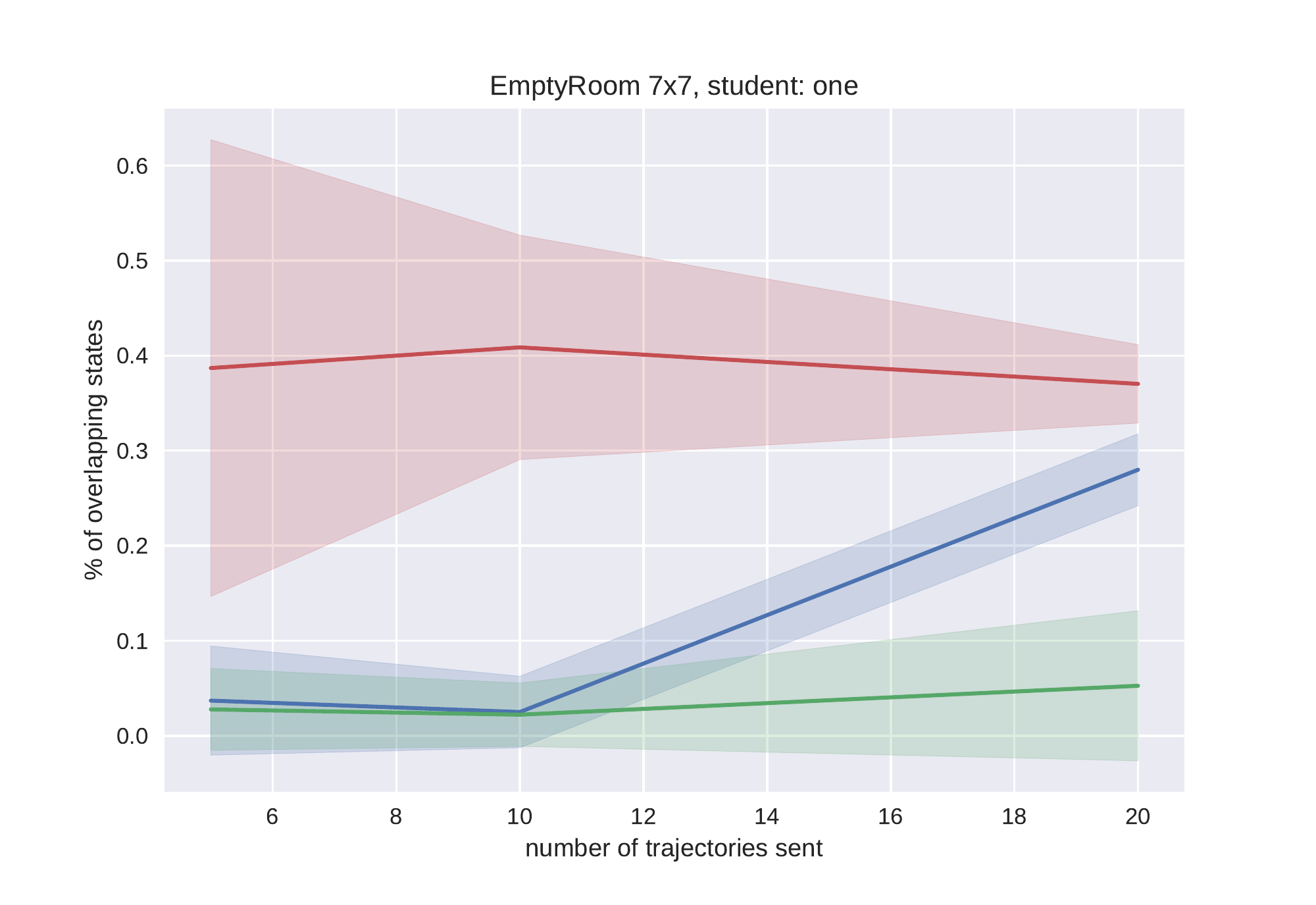} \\ 
    \midrule
    \includegraphics[width=\factor\textwidth]{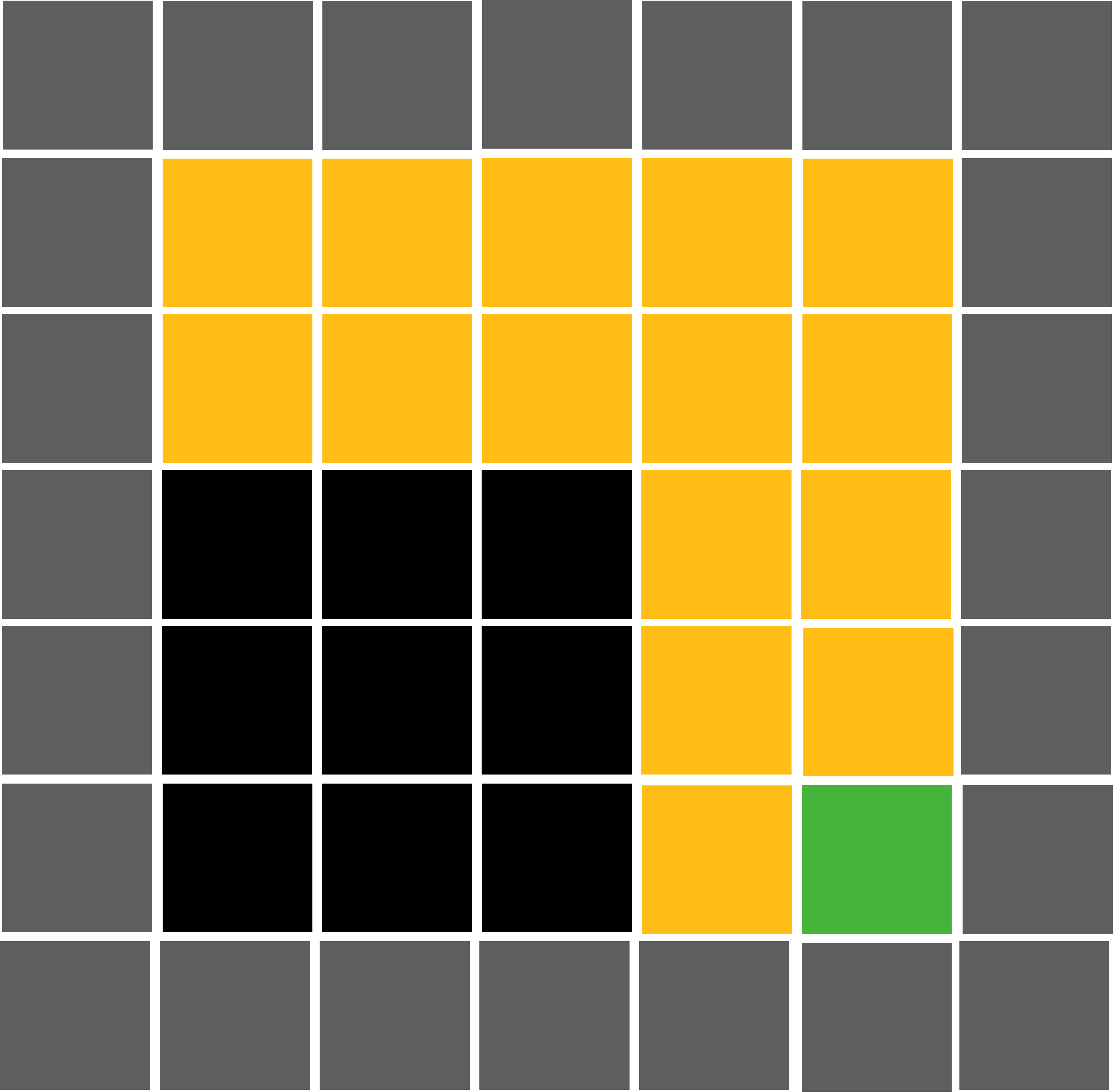} & \includegraphics[width=\graphfactor\textwidth]{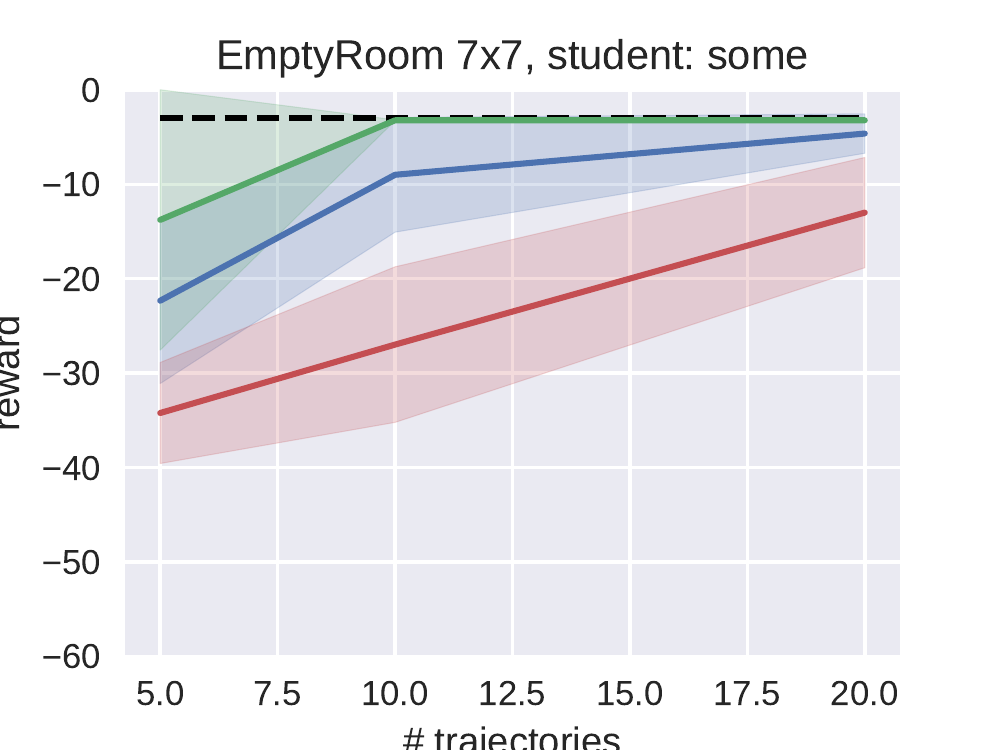} & \includegraphics[width=\graphf\textwidth]{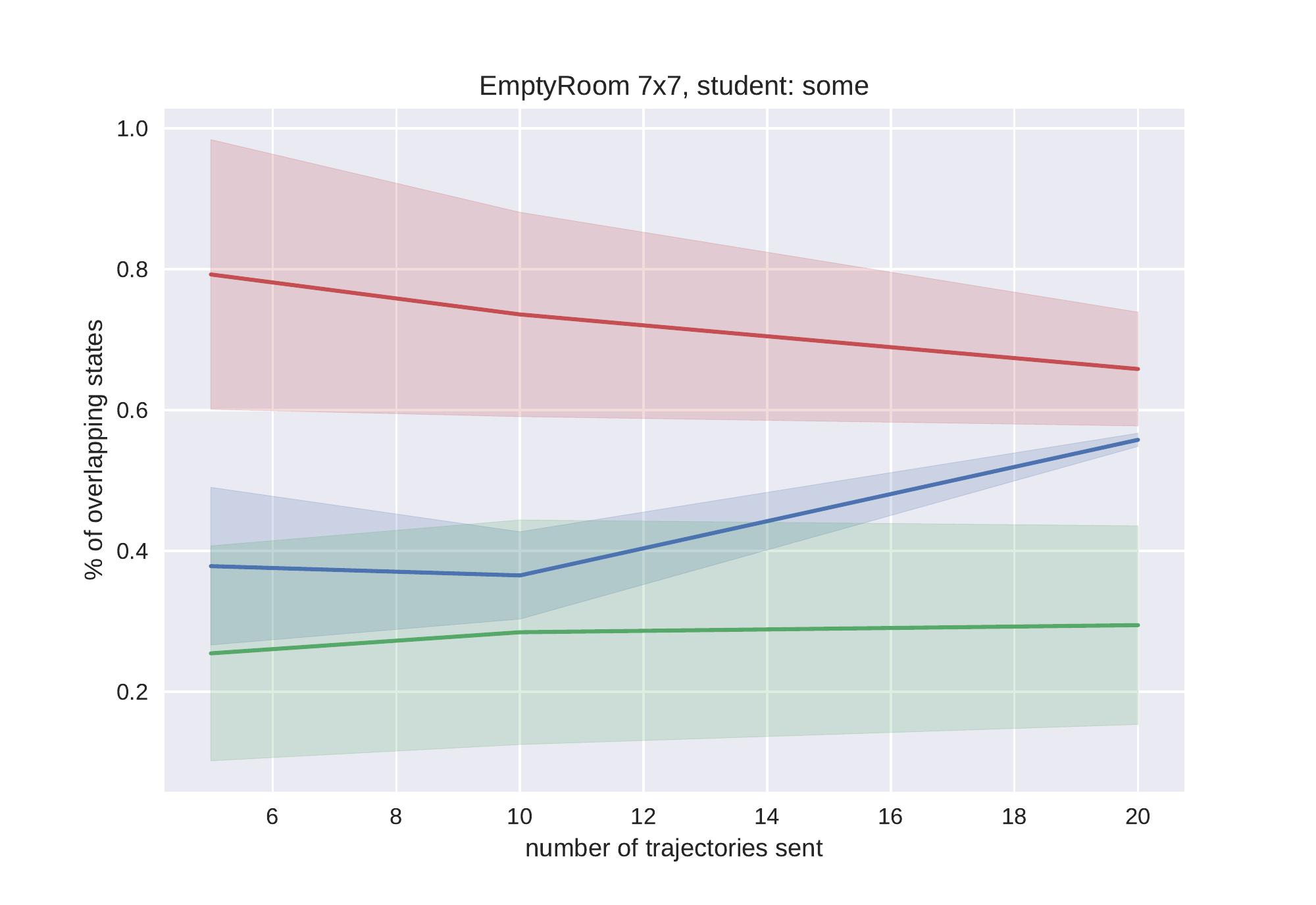} \\ 
    \midrule
    \includegraphics[width=\factor\textwidth]{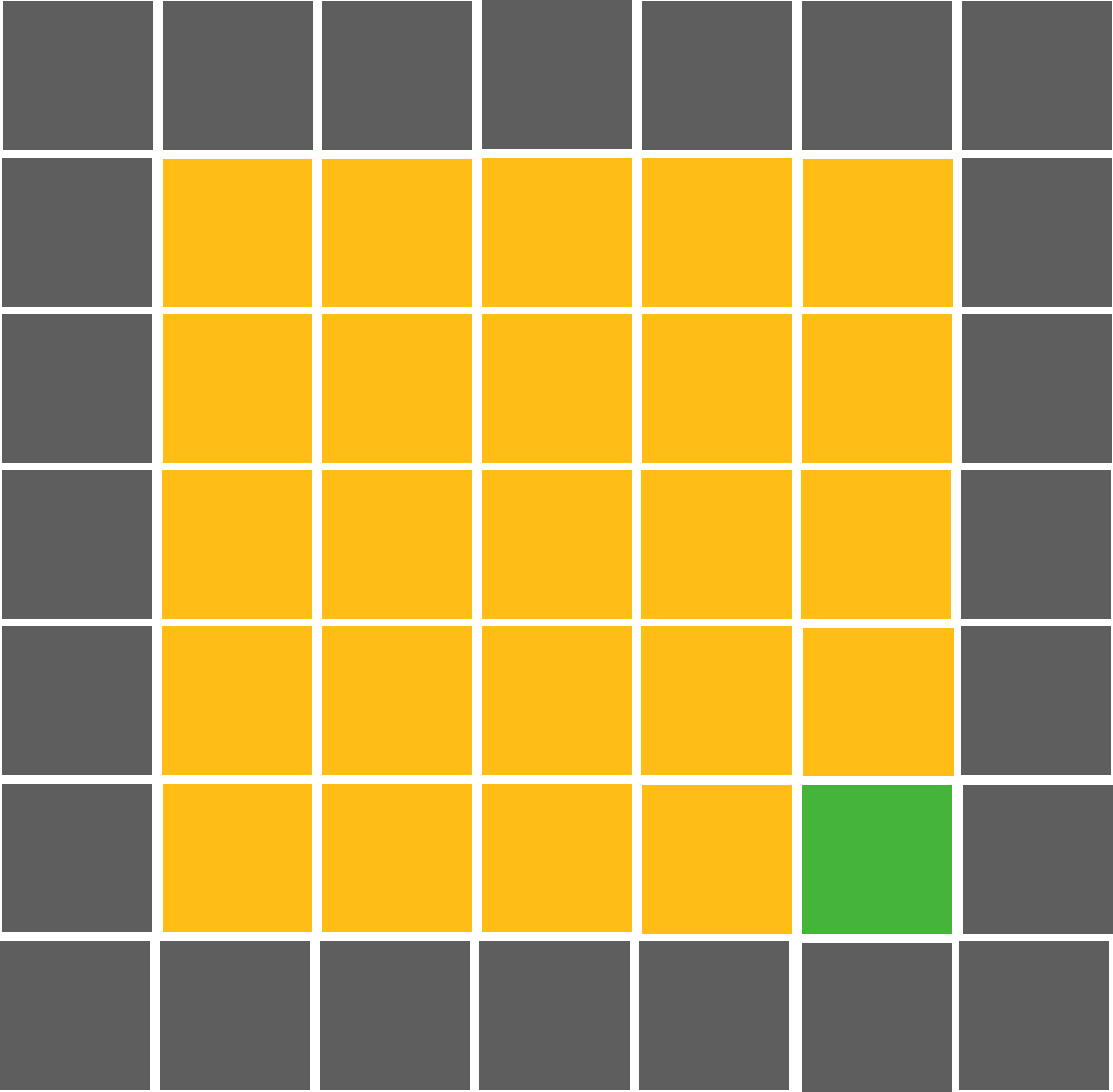} & \includegraphics[width=\graphfactor\textwidth]{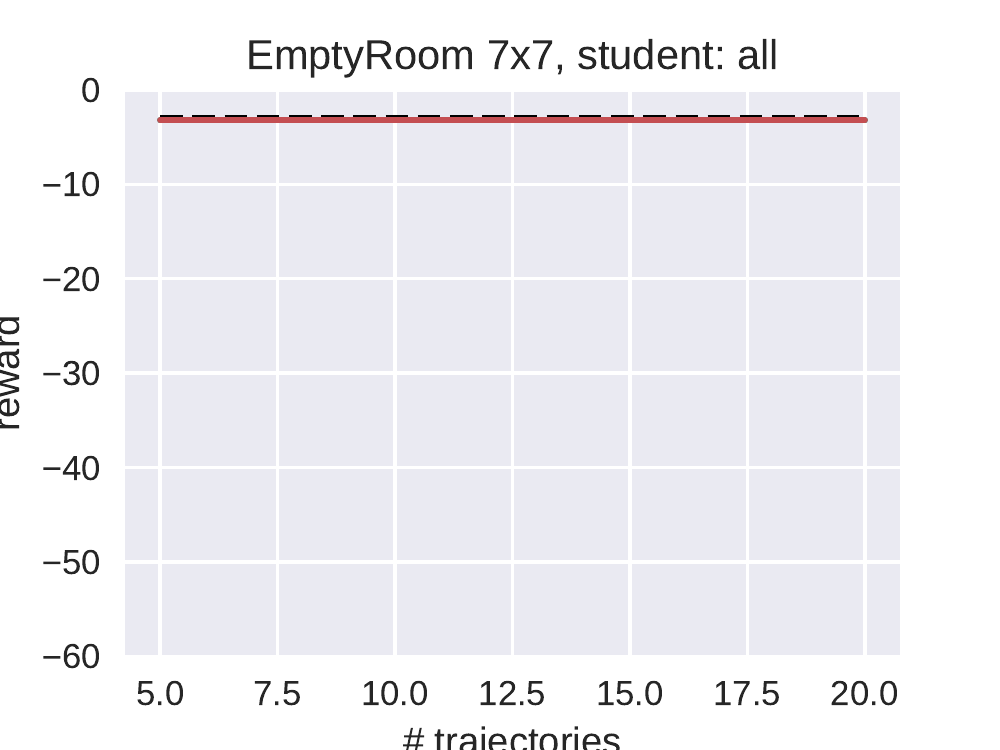} & \includegraphics[width=\graphf\textwidth]{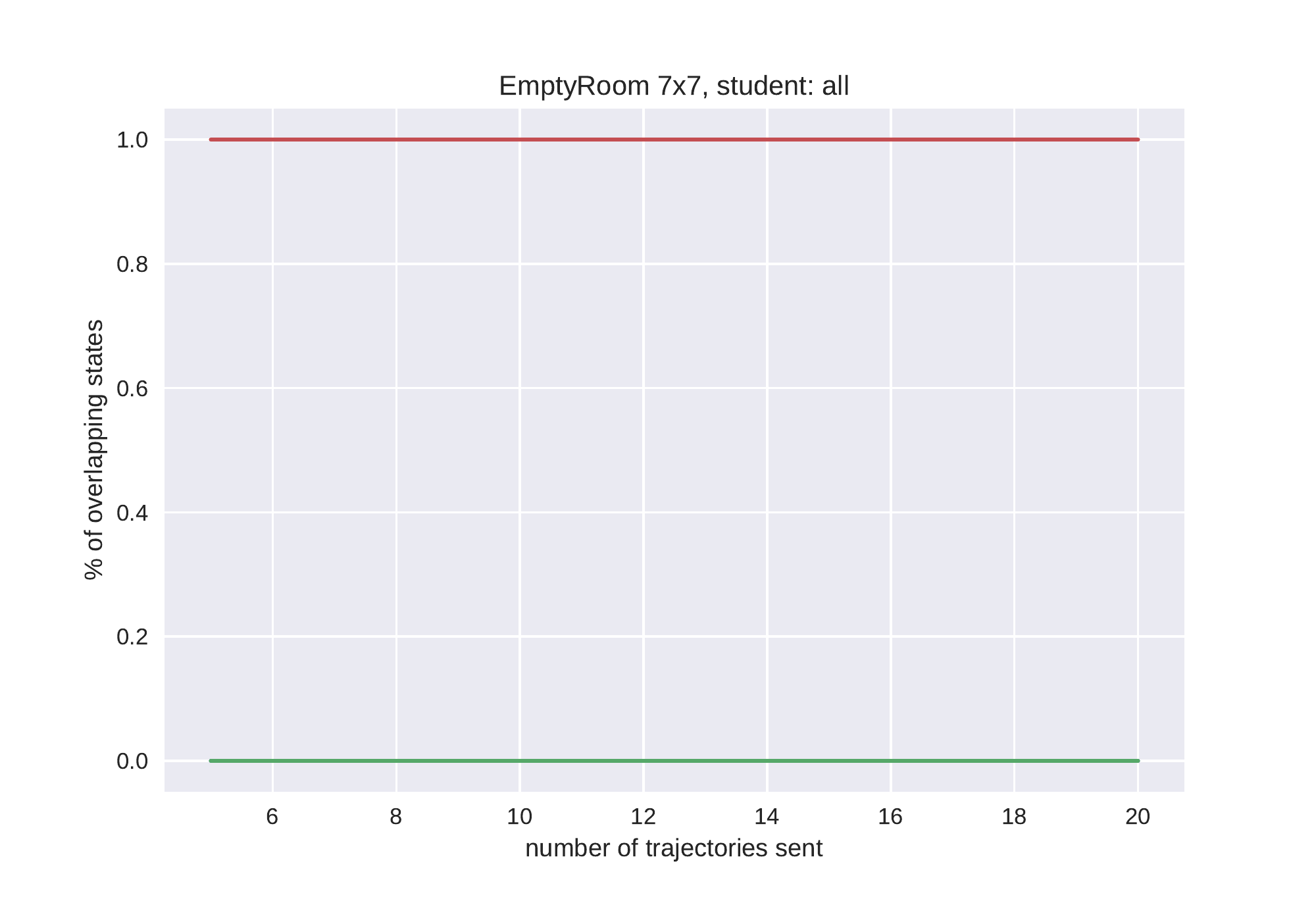} \\ 
    \bottomrule
    \end{tabular} 
    \caption{Student experience with how different teachers teach the student, EmptyRoom 7x7. The dashed lines in the middle column denote the optimal policy's performance.} 
    \label{table:gridworld7x7} 
    \end{table} 
\end{center}

\begin{center} 
    \centering
    \newcommand{\factor}{0.20}
    \newcommand{\graphfactor}{0.33}
    \begin{table} 
    \begin{tabular}{c | c | c} 
    \toprule
    \bf Student experience & \bf Performance  & \bf \% overlap \\
    \midrule\midrule
    \includegraphics[width=0.25\textwidth]{images/gridworld/legend_reward_by_num_examples.pdf}  \\ \includegraphics[width=\factor\textwidth]{images/gridworld/7x7/7x7_none.pdf} & \includegraphics[width=\graphfactor\textwidth]{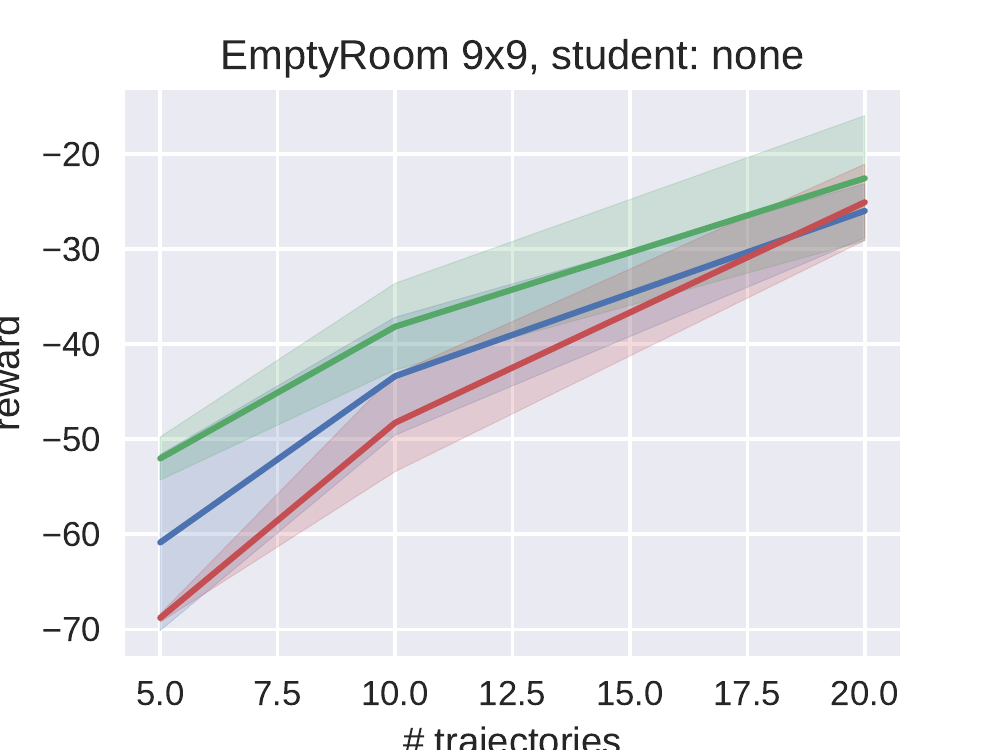} & \includegraphics[width=\graphfactor\textwidth]{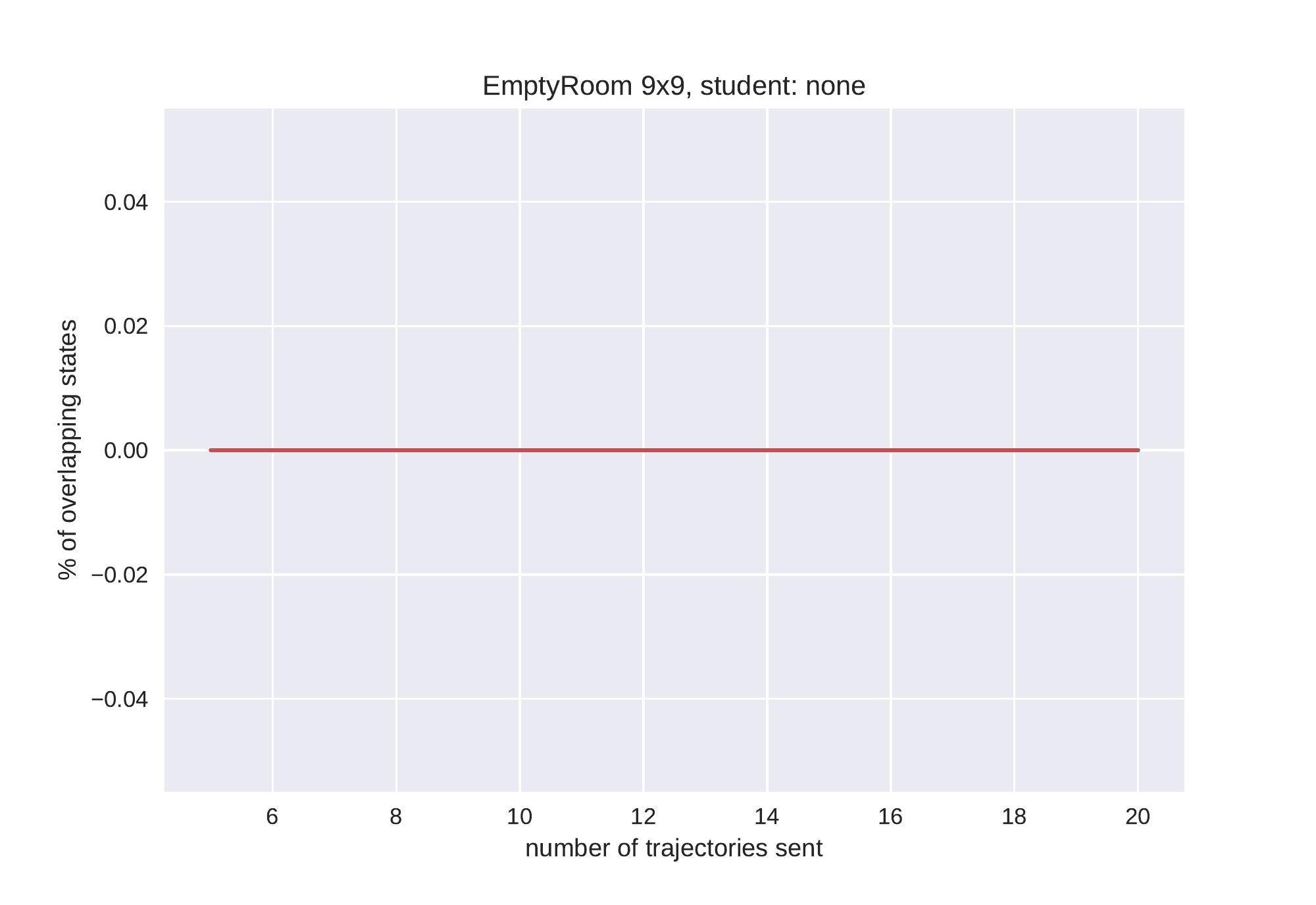}
     \\
    \midrule
    \includegraphics[width=\factor\textwidth]{images/gridworld/7x7/7x7_one.pdf} & \includegraphics[width=\graphfactor\textwidth]{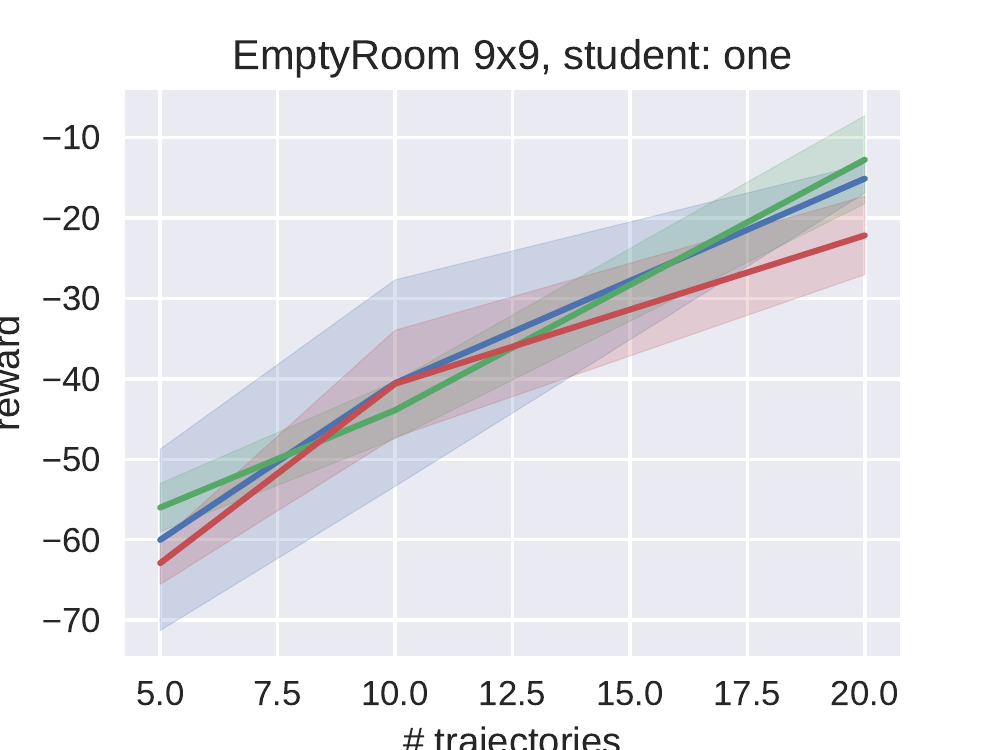} & \includegraphics[width=\graphfactor\textwidth]{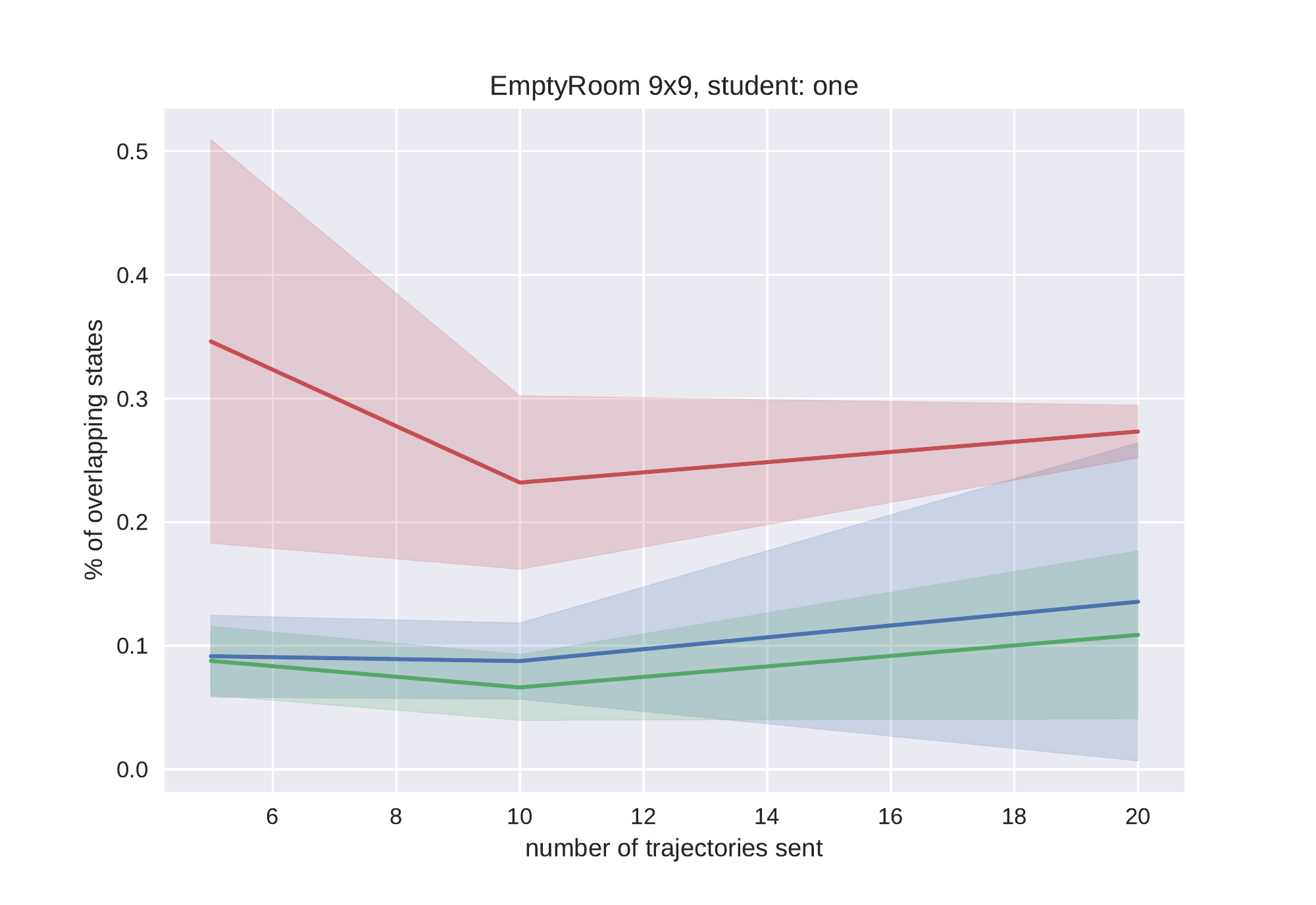} \\ 
    \midrule
    \includegraphics[width=\factor\textwidth]{images/gridworld/7x7/7x7_some.pdf} & \includegraphics[width=\graphfactor\textwidth]{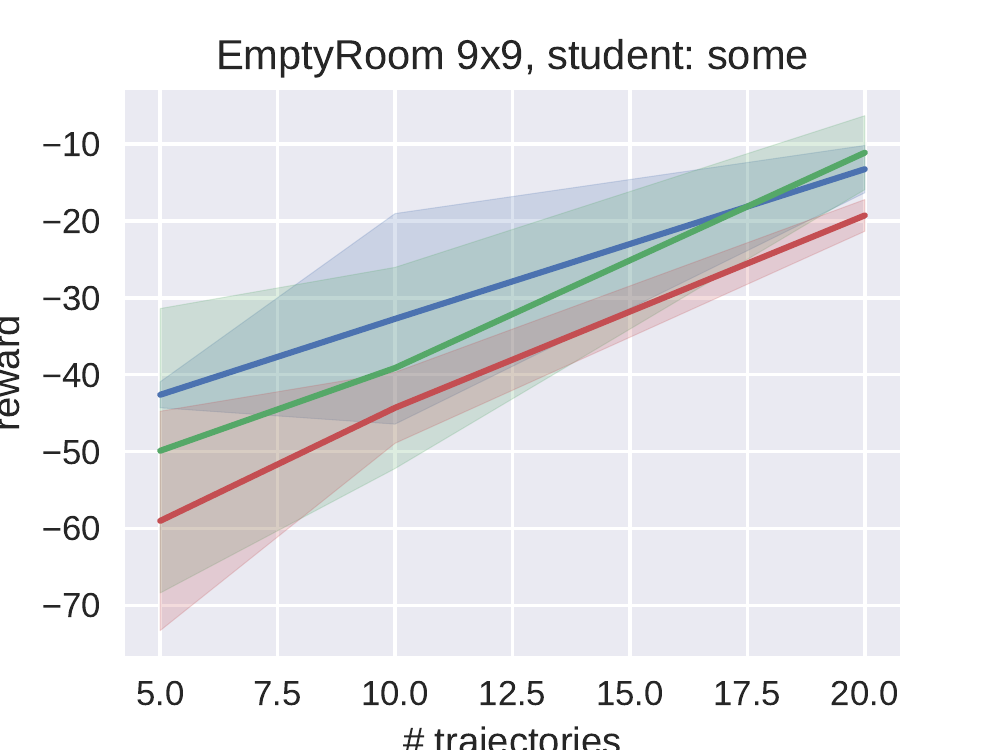} & \includegraphics[width=\graphfactor\textwidth]{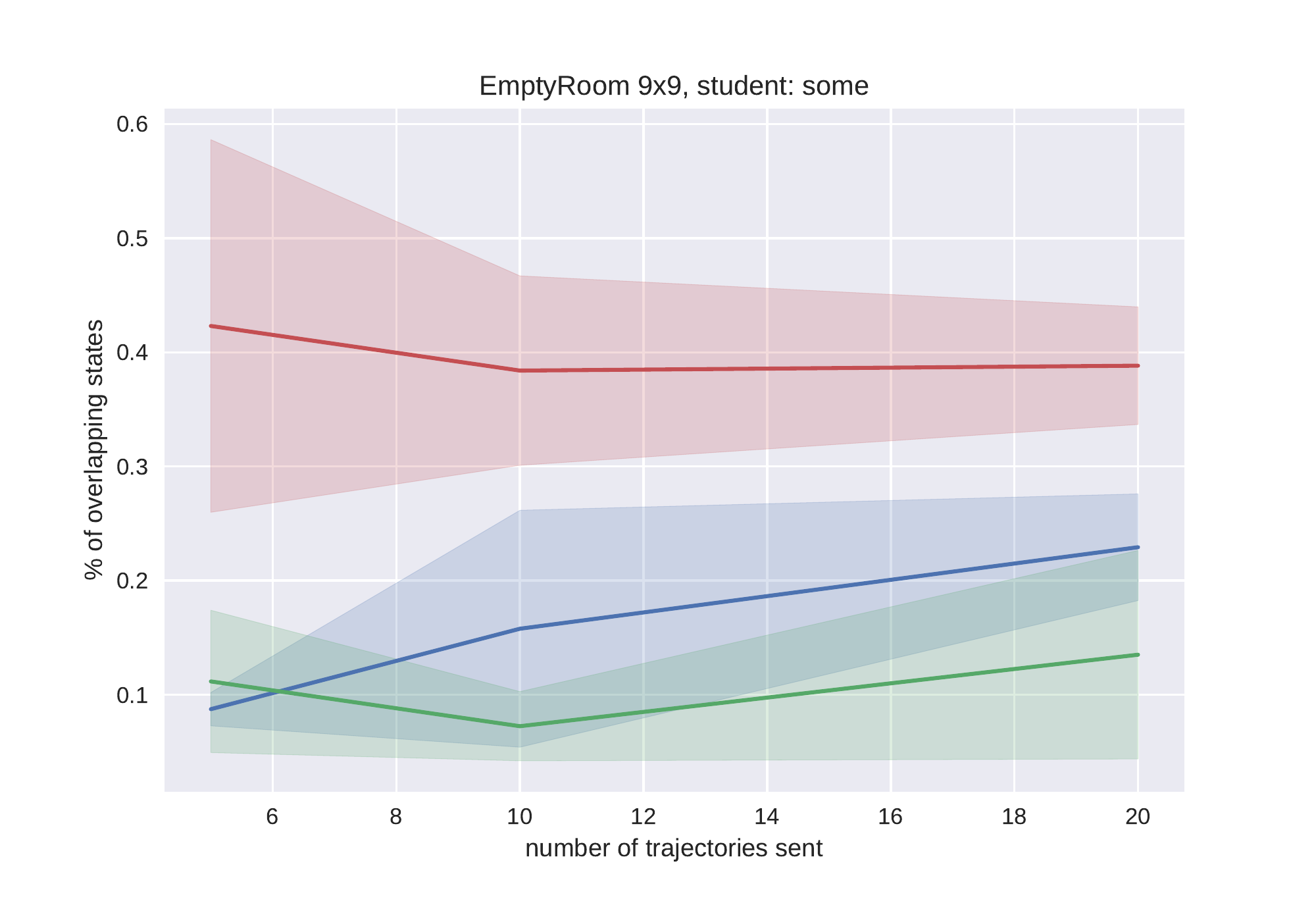} \\ 
    \midrule
    \includegraphics[width=\factor\textwidth]{images/gridworld/7x7/7x7_all.pdf} & \includegraphics[width=\graphfactor\textwidth]{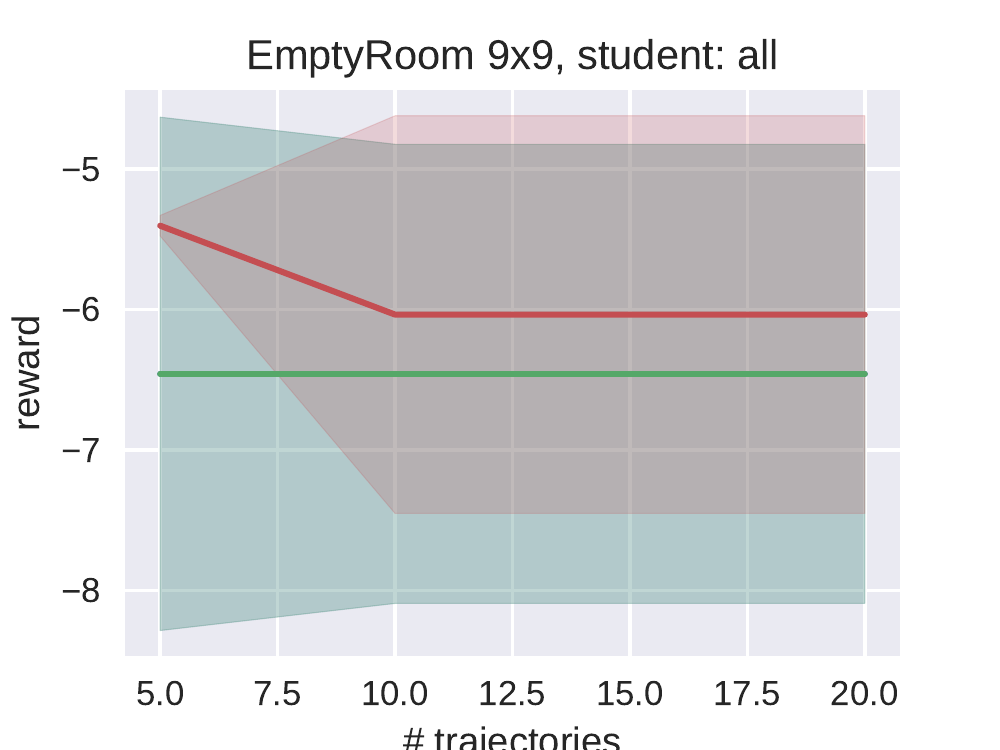} & \includegraphics[width=\graphfactor\textwidth]{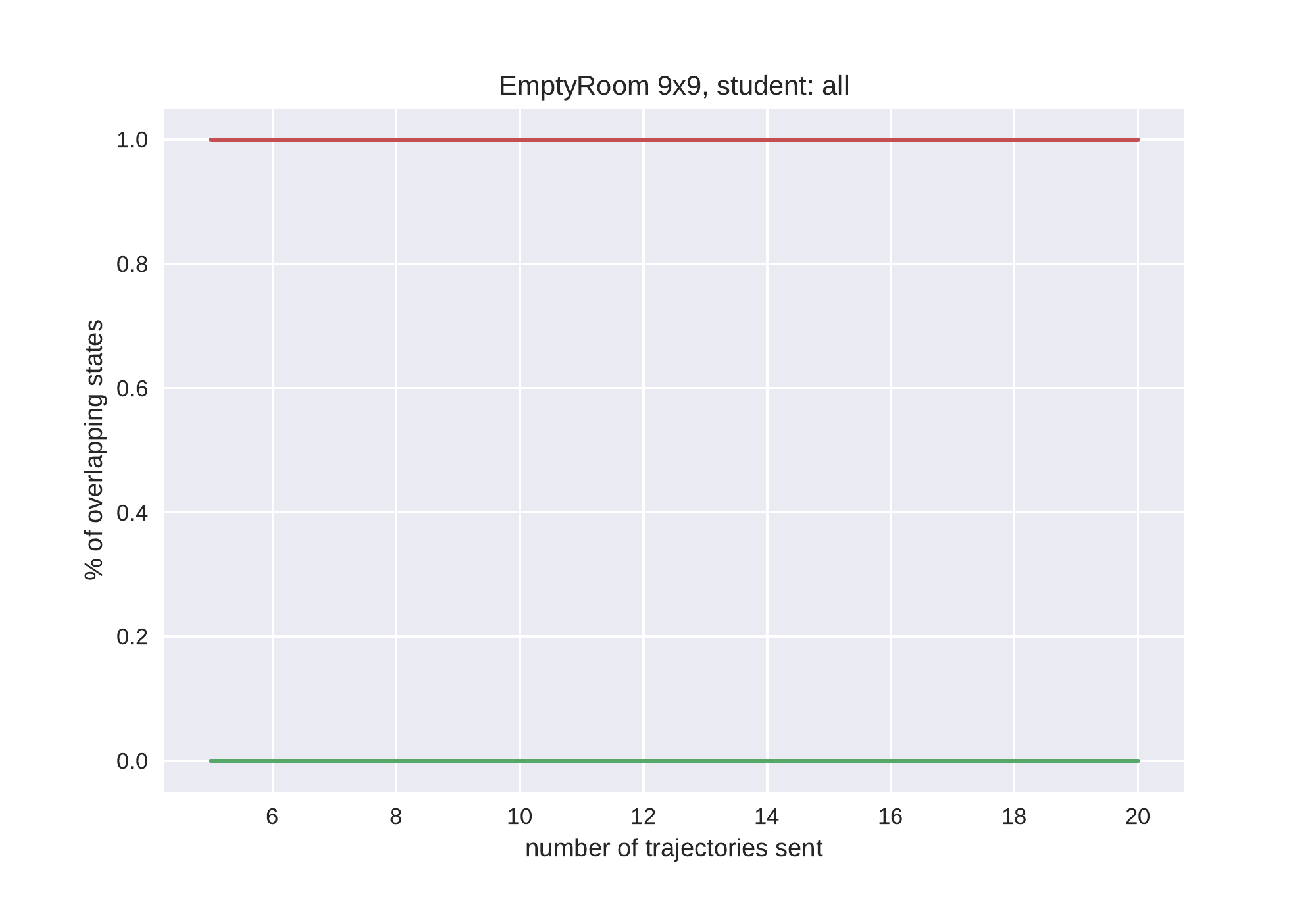} \\ 
    \bottomrule
    \end{tabular} 
    \caption{Student experience with how different teachers teach the student, EmptyRoom 9x9. The dashed lines in the middle column denote the optimal policy's performance.} 
    \label{table:gridworld9x9} 
    \end{table} 
\end{center} 

\begin{center} 
    \centering
    \newcommand{\factor}{0.20}
    \newcommand{\graphfactor}{0.33}
    \begin{table} 
    \begin{tabular}{c | c | c} 
    \toprule
    \bf Student experience & \bf Performance  & \bf \% overlap \\
    \midrule\midrule
    \includegraphics[width=0.25\textwidth]{images/gridworld/legend_reward_by_num_examples.pdf}  \\ \includegraphics[width=\factor\textwidth]{images/gridworld/7x7/7x7_none.pdf} & \includegraphics[width=\graphfactor\textwidth]{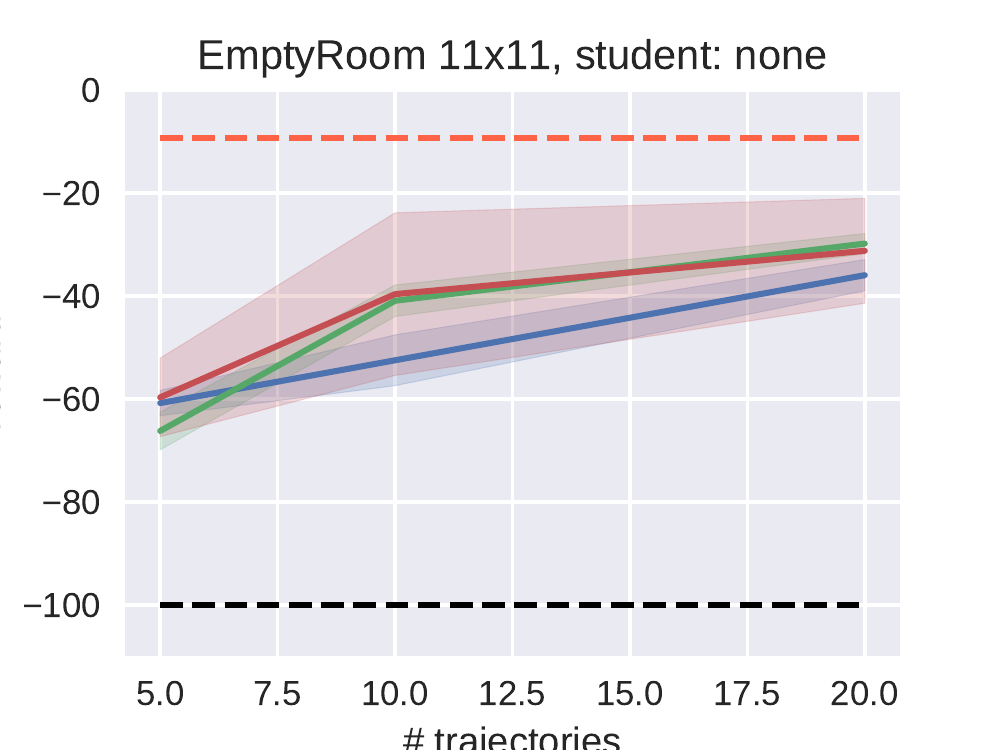} & \includegraphics[width=\graphfactor\textwidth]{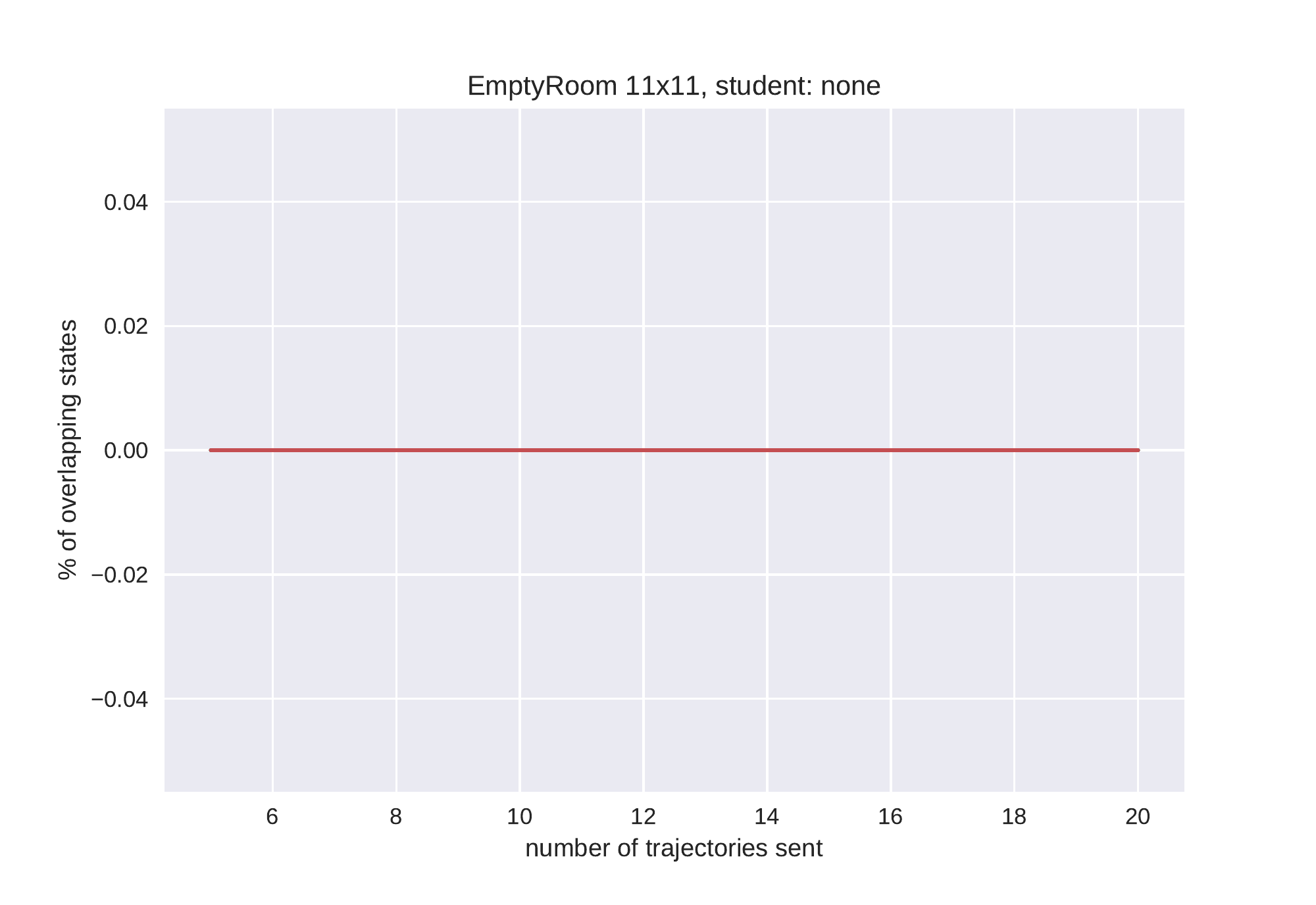}
     \\
    \midrule
    \includegraphics[width=\factor\textwidth]{images/gridworld/7x7/7x7_one.pdf} & \includegraphics[width=\graphfactor\textwidth]{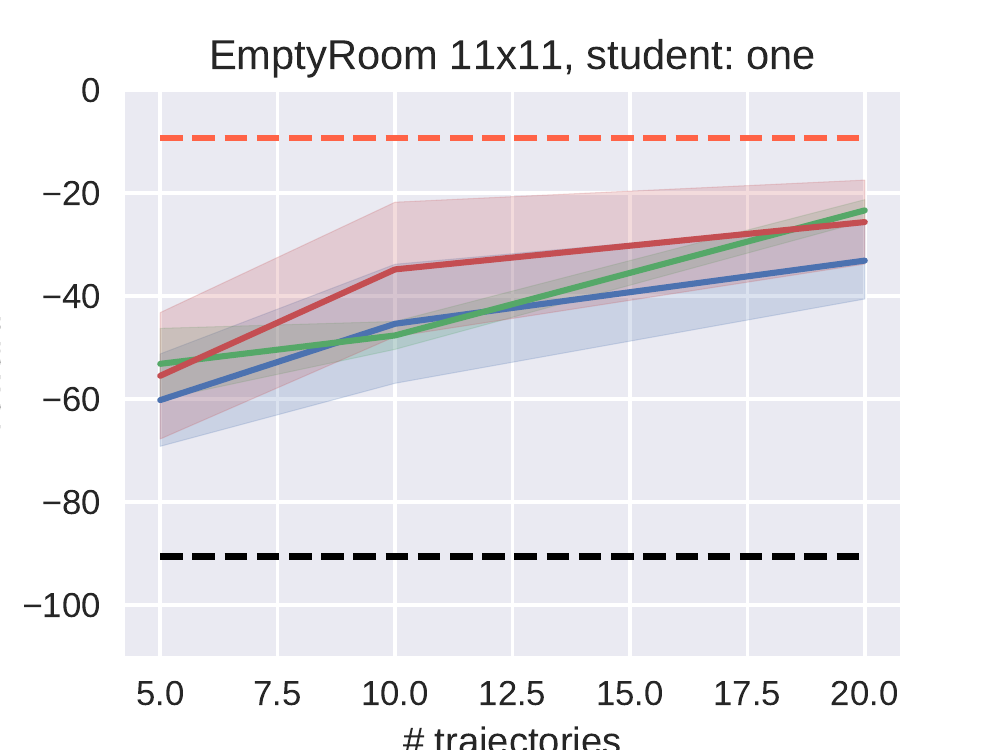} & \includegraphics[width=\graphfactor\textwidth]{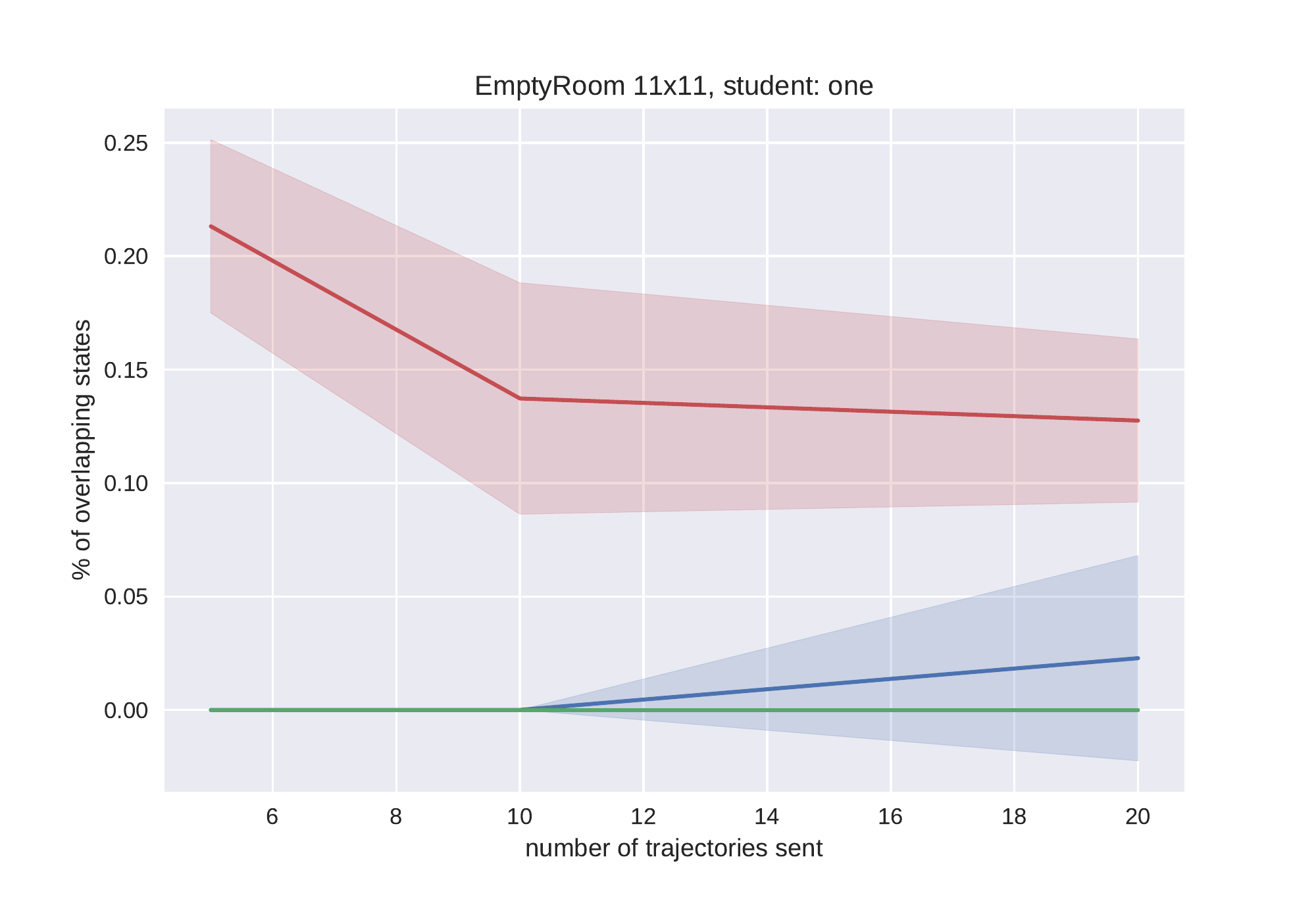} \\ 
    \midrule
    \includegraphics[width=\factor\textwidth]{images/gridworld/7x7/7x7_some.pdf} & \includegraphics[width=\graphfactor\textwidth]{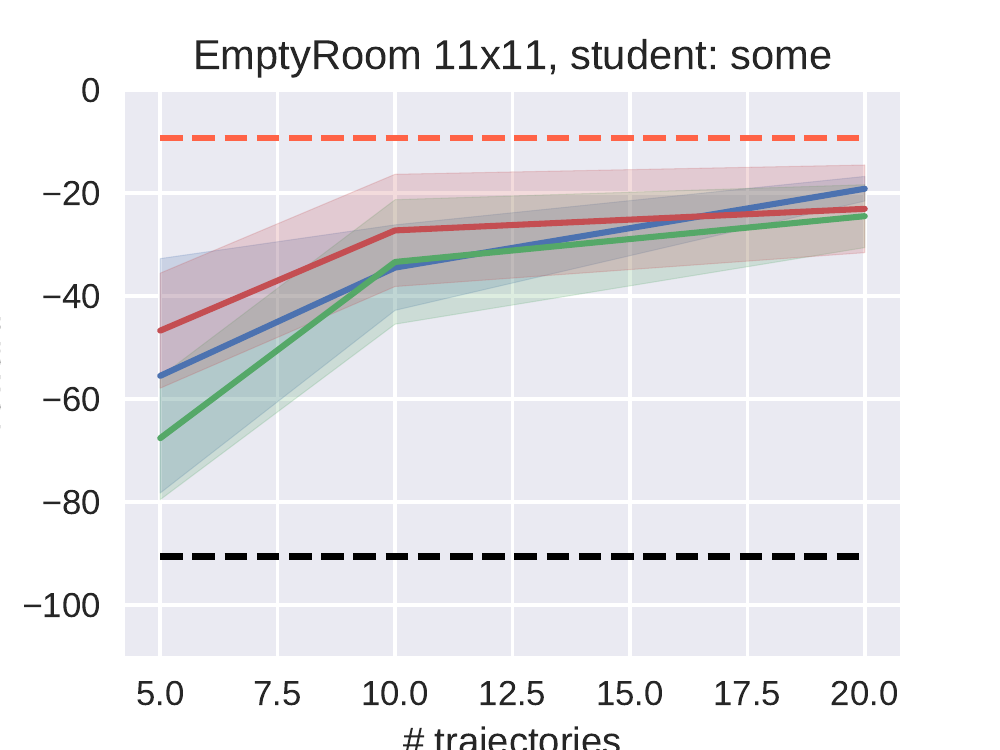} & \includegraphics[width=\graphfactor\textwidth]{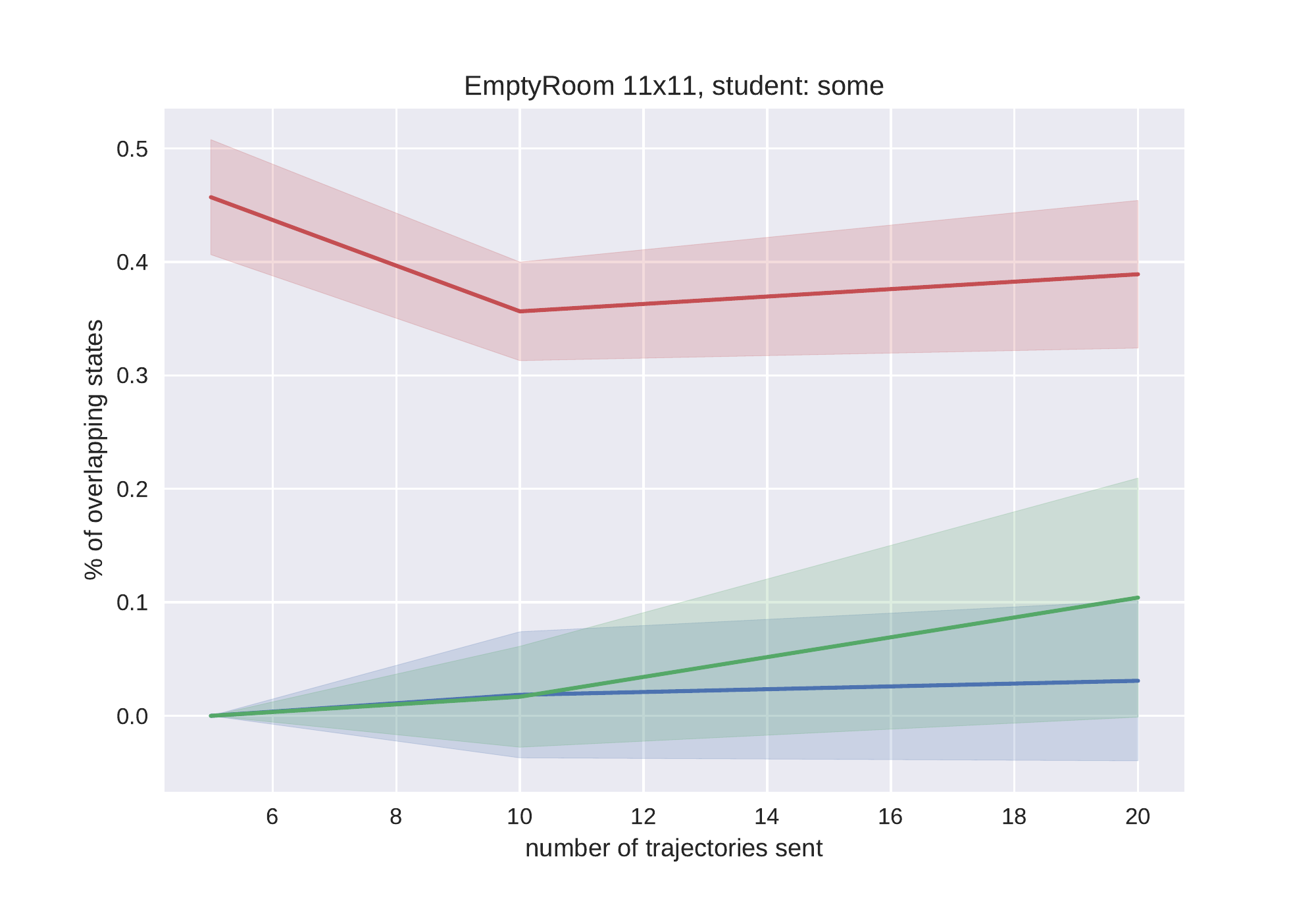} \\ 
    \midrule
    \includegraphics[width=\factor\textwidth]{images/gridworld/7x7/7x7_all.pdf} & \includegraphics[width=\graphfactor\textwidth]{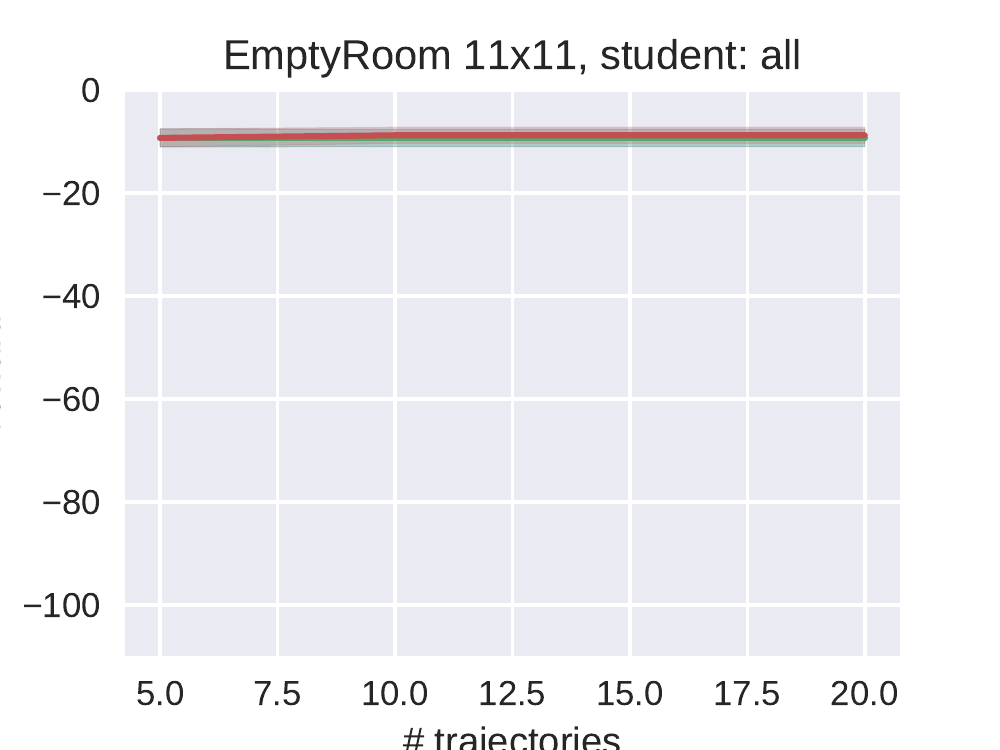} & \includegraphics[width=\graphfactor\textwidth]{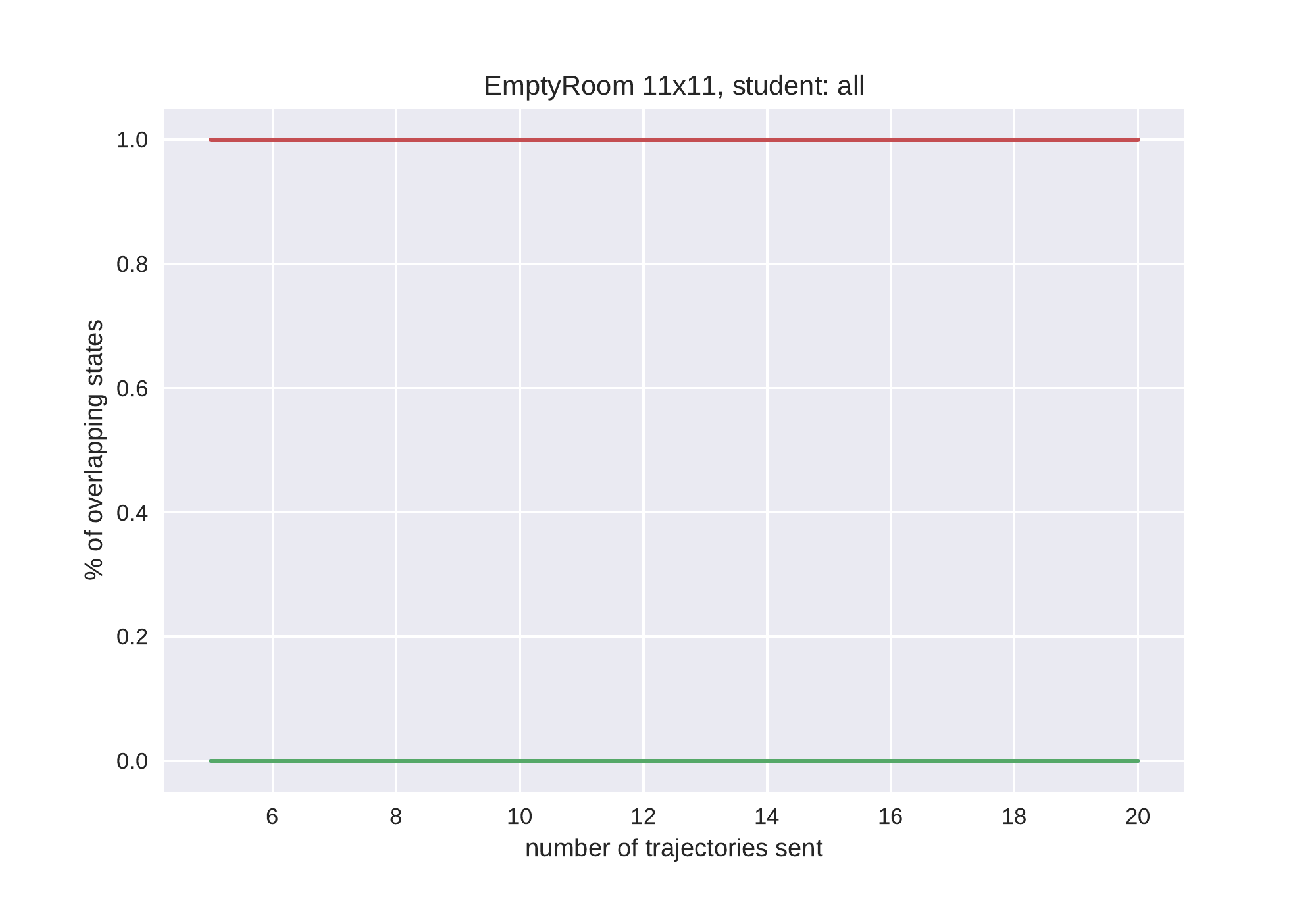} \\ 
    \bottomrule
    \end{tabular} 
    \caption{Student experience with how different teachers teach the student, EmptyRoom 11x11. The dashed lines in the middle column denote the optimal policy's performance.} 
    \label{table:gridworld11x11} 
    \end{table} 
\end{center}

\begin{center} 
    \centering
    \newcommand{\factor}{0.20}
    \newcommand{\graphfactor}{0.33}
    \begin{table} 
    \begin{tabular}{c | c | c} 
    \toprule
    \bf Student experience & \bf Performance  & \bf \% overlap \\
    \midrule\midrule
    \includegraphics[width=0.25\textwidth]{images/gridworld/legend_reward_by_num_examples.pdf}  \\ \includegraphics[width=\factor\textwidth]{images/gridworld/7x7/7x7_none.pdf} & \includegraphics[width=\graphfactor\textwidth]{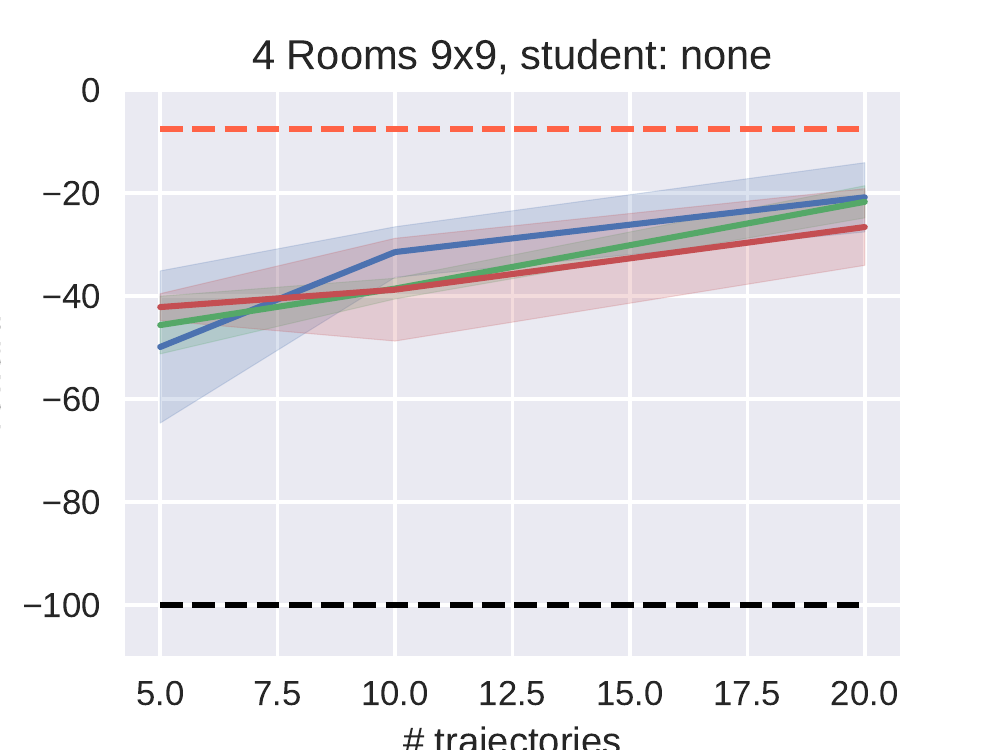} & \includegraphics[width=\graphfactor\textwidth]{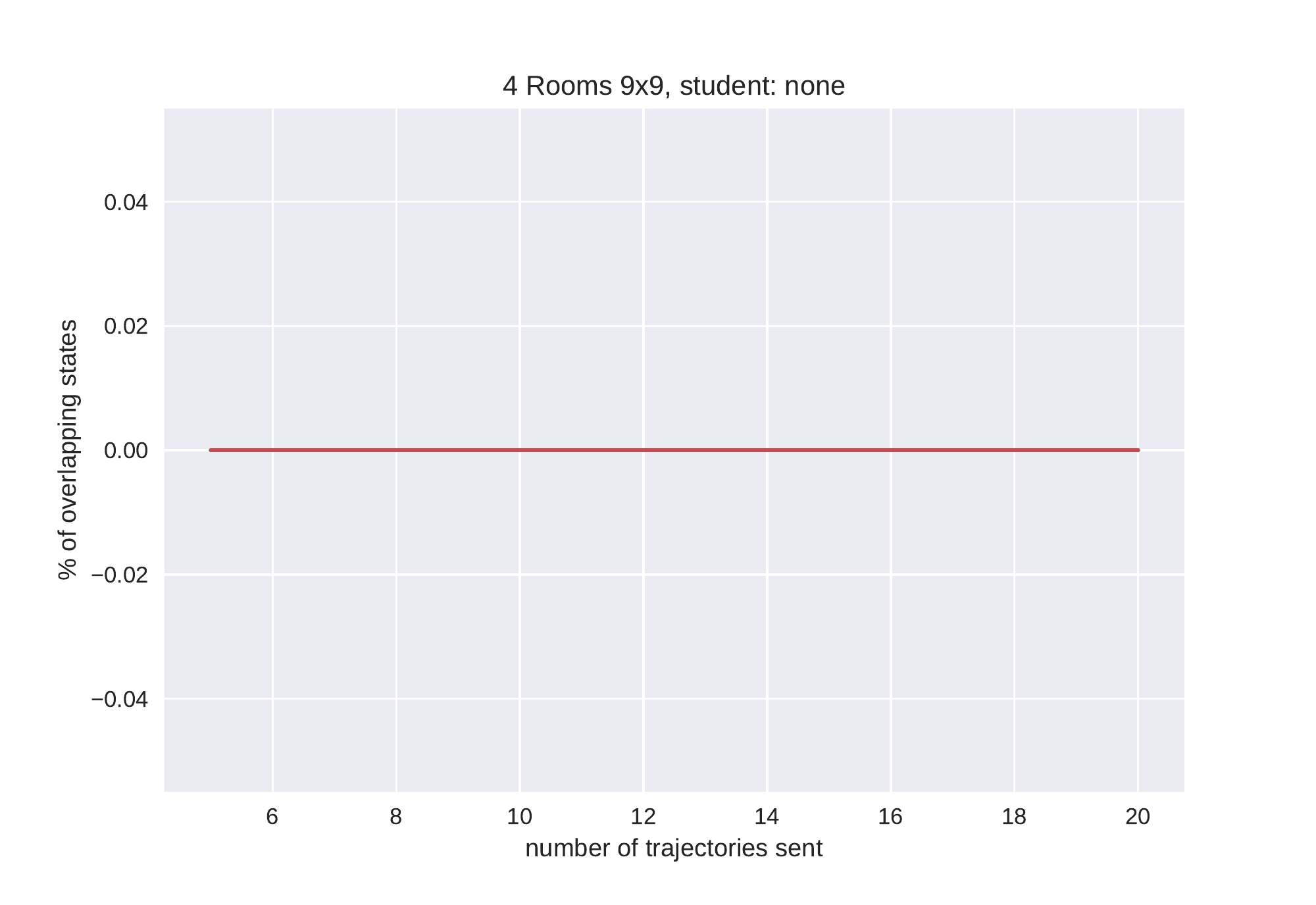}\\
    \midrule
    \includegraphics[width=\factor\textwidth]{images/gridworld/7x7/7x7_one.pdf} & \includegraphics[width=\graphfactor\textwidth]{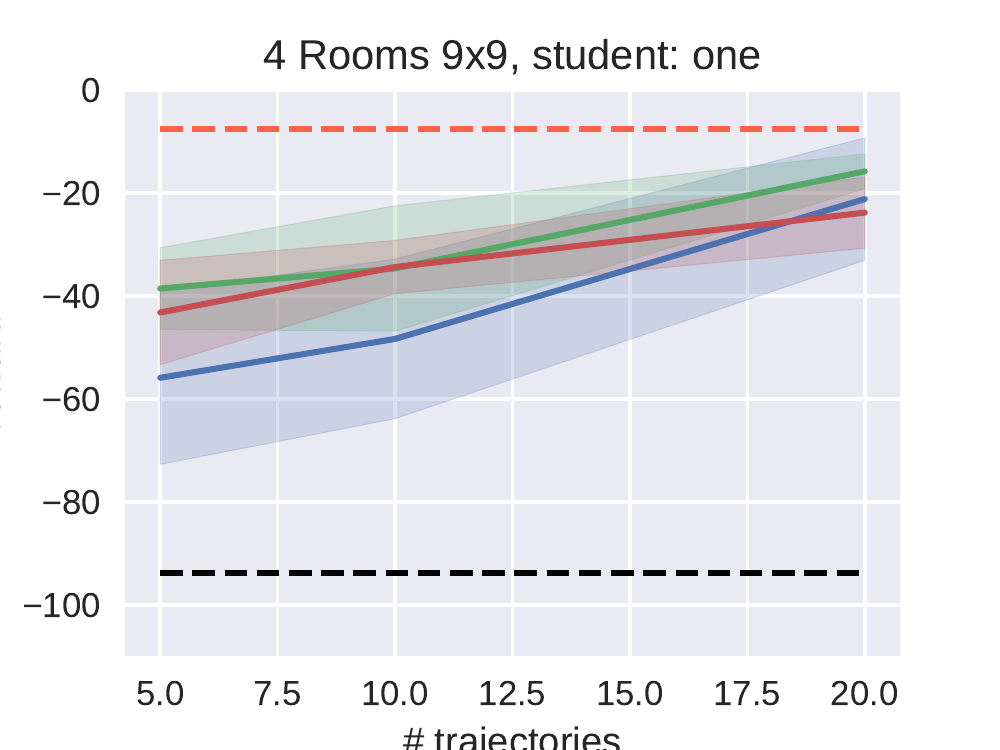} & \includegraphics[width=\graphfactor\textwidth]{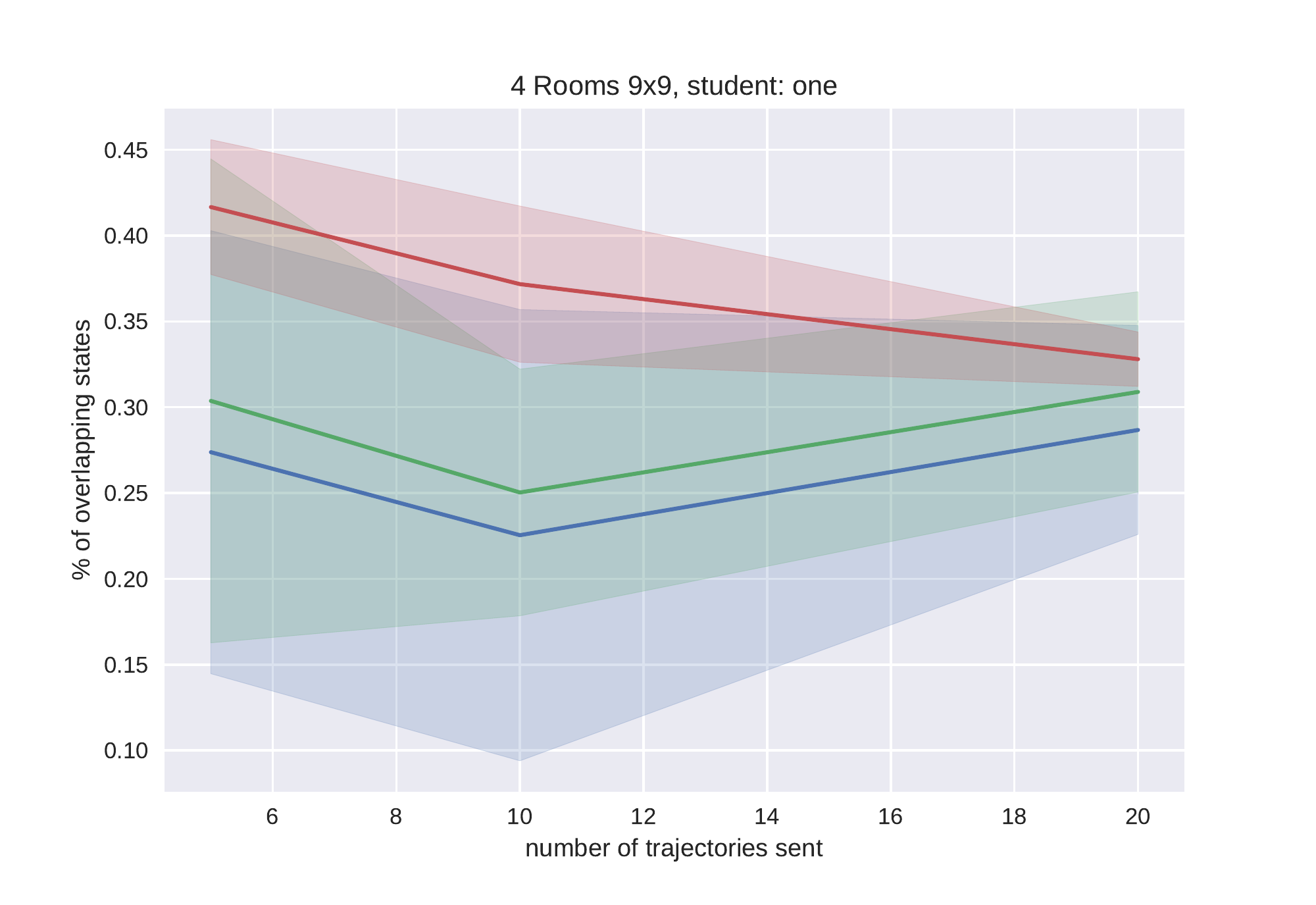} \\ 
    \midrule
    \includegraphics[width=\factor\textwidth]{images/gridworld/7x7/7x7_some.pdf} & \includegraphics[width=\graphfactor\textwidth]{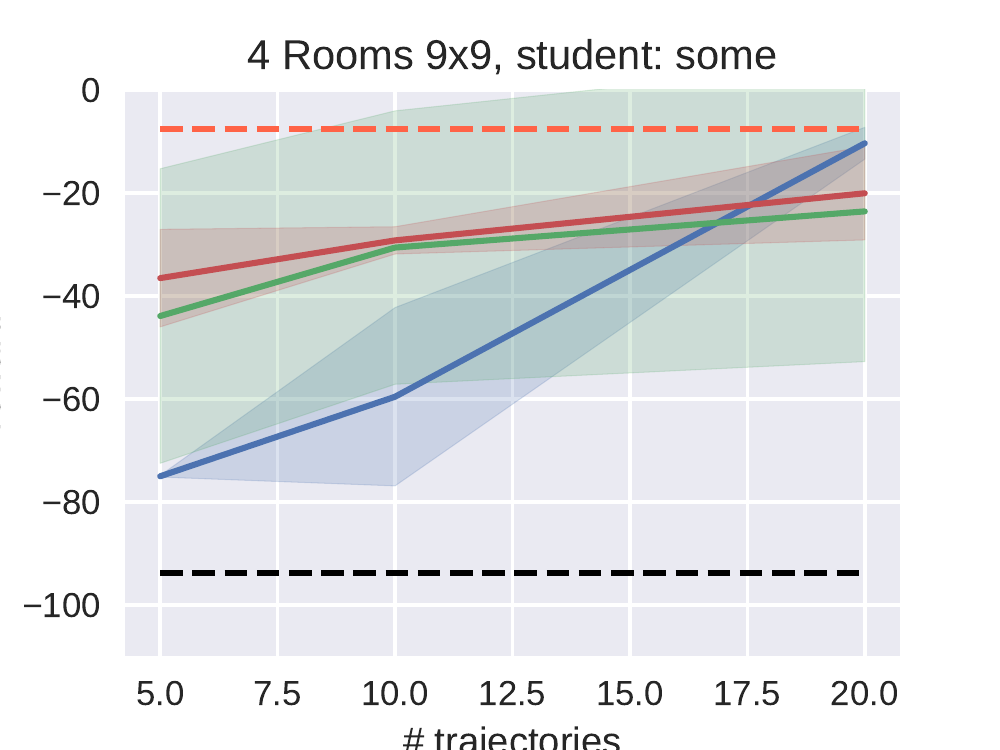} & \includegraphics[width=\graphfactor\textwidth]{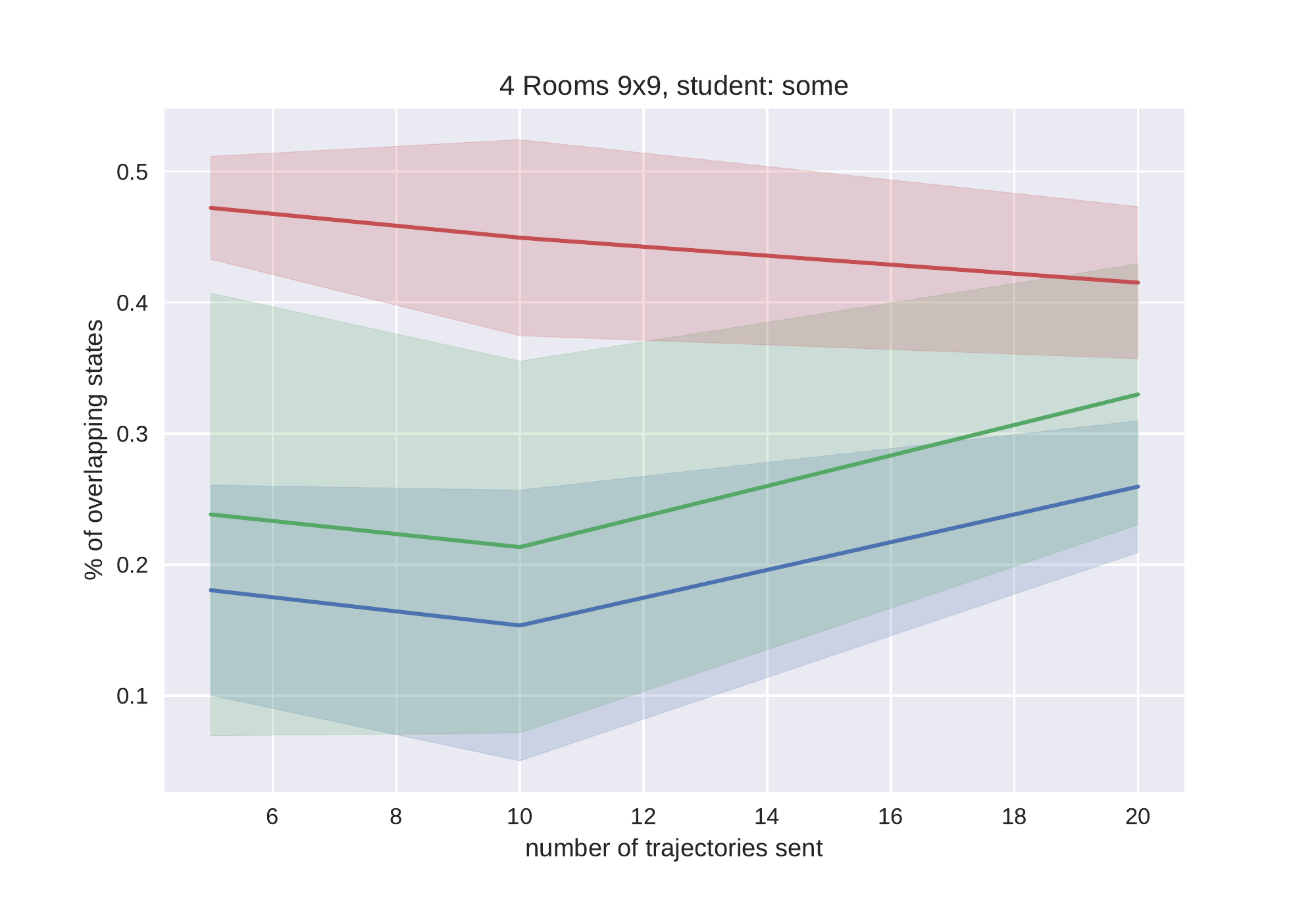} \\ 
    \midrule
    \includegraphics[width=\factor\textwidth]{images/gridworld/7x7/7x7_all.pdf} & \includegraphics[width=\graphfactor\textwidth]{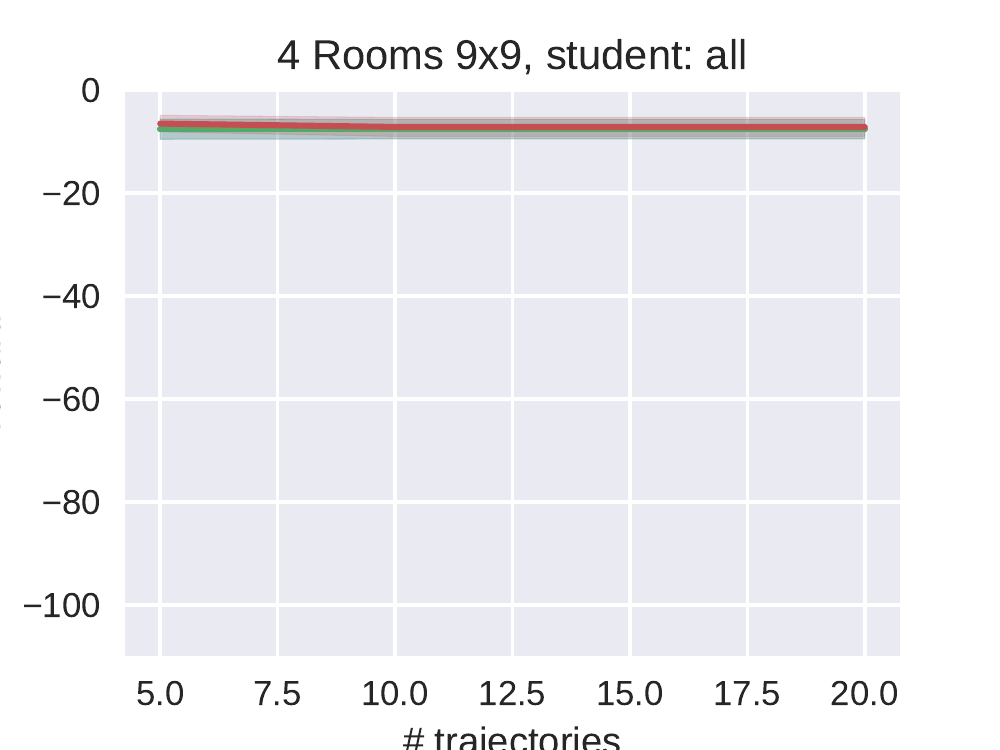} & \includegraphics[width=\graphfactor\textwidth]{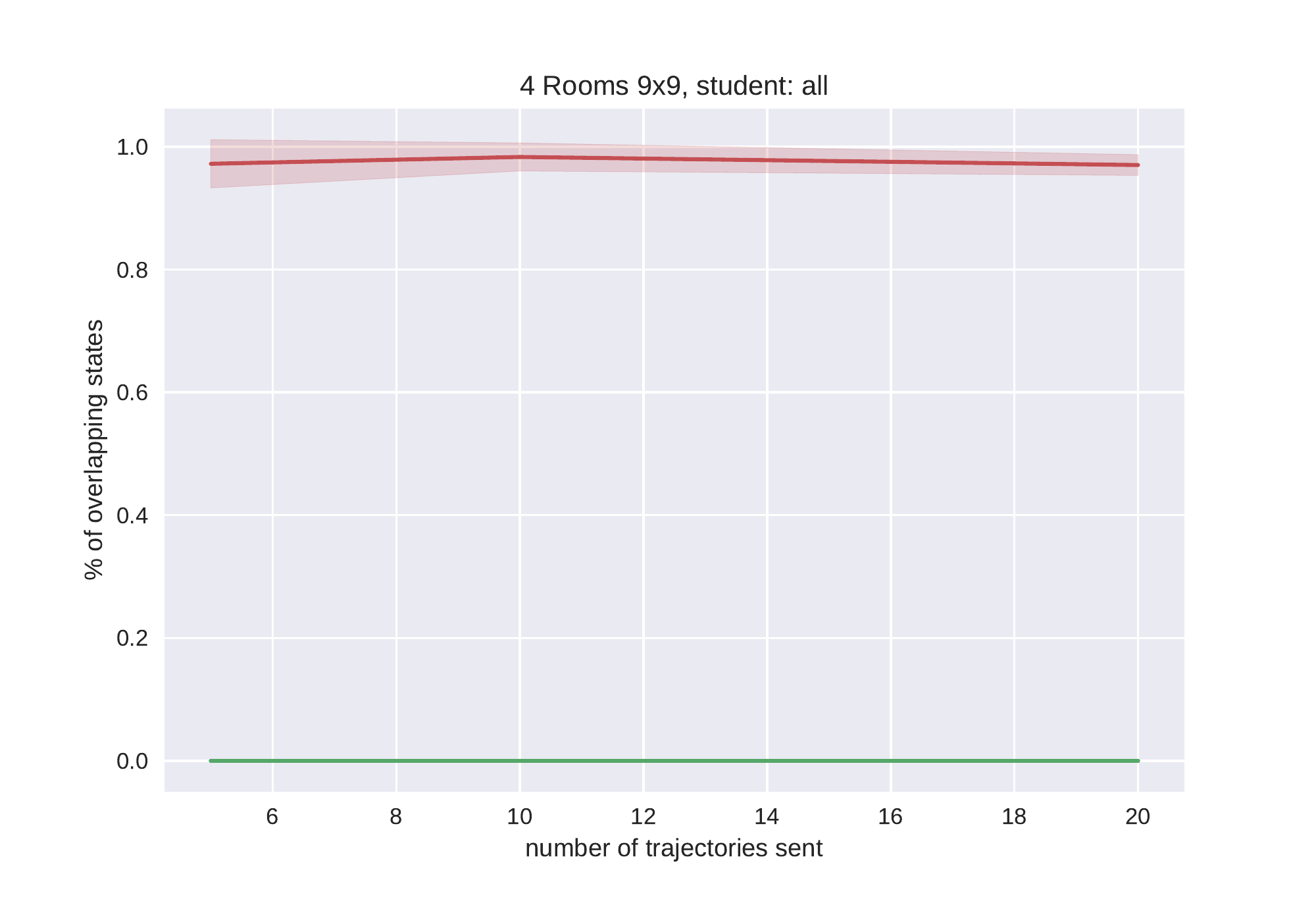} \\ 
    \bottomrule
    \end{tabular} 
    \caption{Student experience with how different teachers teach the student, 4 rooms 9. The dashed lines in the middle column denote the optimal policy's performance.} 
    \label{table:gridworld4rooms9} 
    \end{table} 
\end{center}

\begin{center} 
    \centering
    \newcommand{\factor}{0.20}
    \newcommand{\graphfactor}{0.33}
    \begin{table} 
    \begin{tabular}{c | c | c} 
    \toprule
    \bf Student experience & \bf Performance  & \bf \% overlap \\
    \midrule\midrule
    \includegraphics[width=0.25\textwidth]{images/gridworld/legend_reward_by_num_examples.pdf}  \\ \includegraphics[width=\factor\textwidth]{images/gridworld/7x7/7x7_none.pdf} & \includegraphics[width=\graphfactor\textwidth]{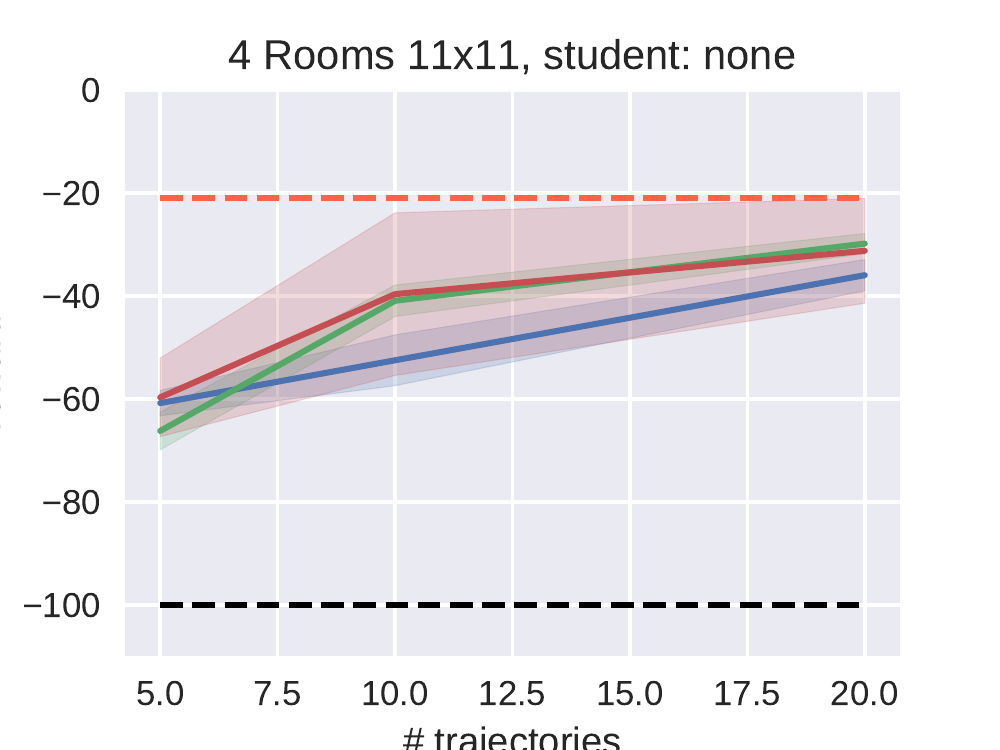} & \includegraphics[width=\graphfactor\textwidth]{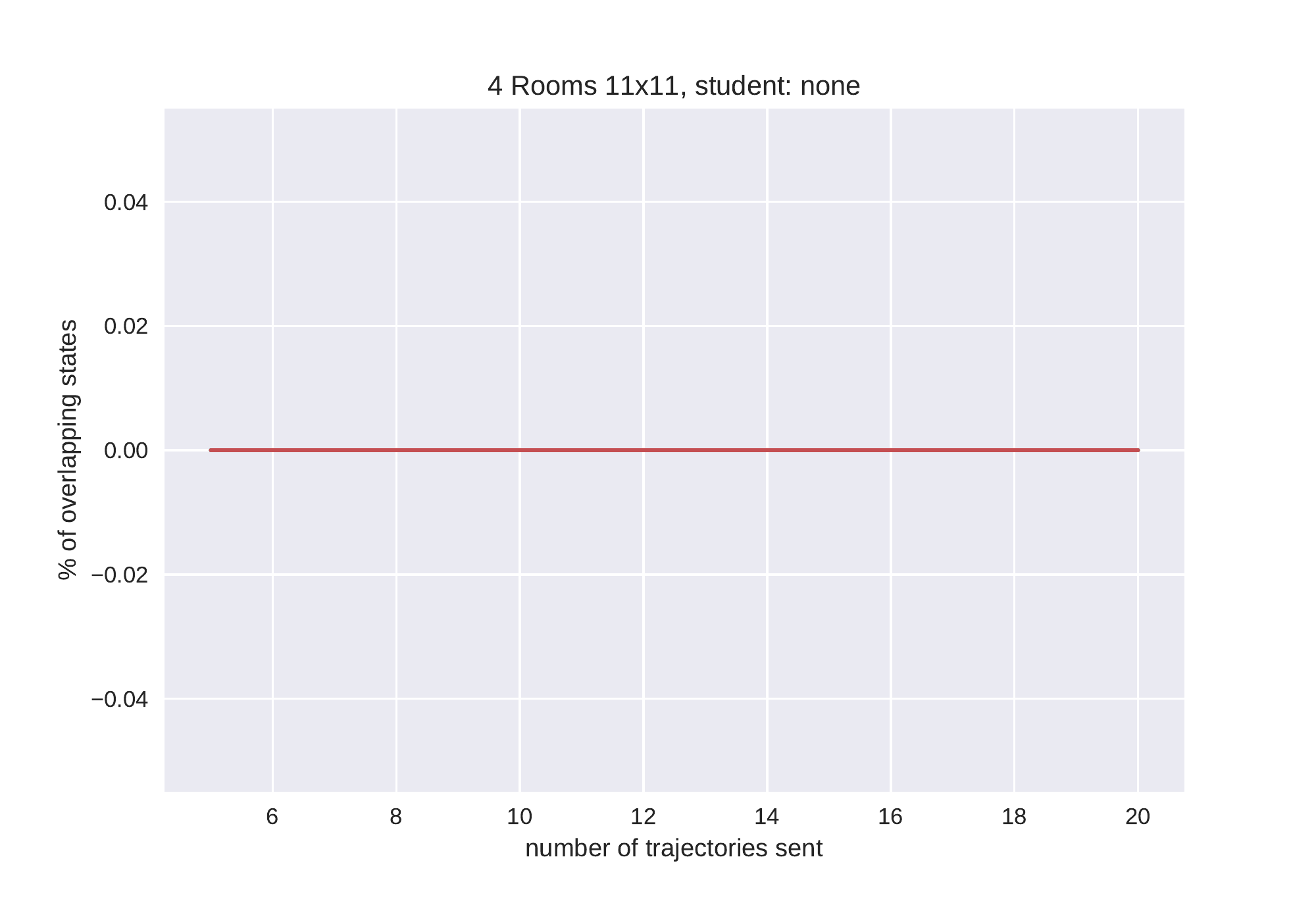}\\
    \midrule
    \includegraphics[width=\factor\textwidth]{images/gridworld/7x7/7x7_one.pdf} & \includegraphics[width=\graphfactor\textwidth]{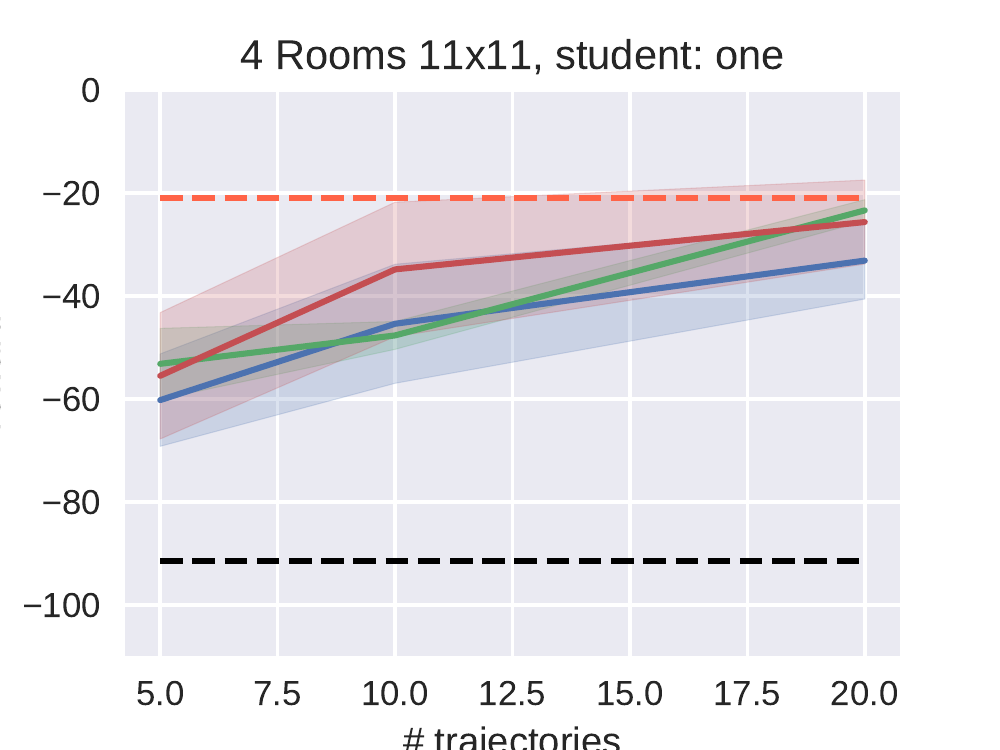} & \includegraphics[width=\graphfactor\textwidth]{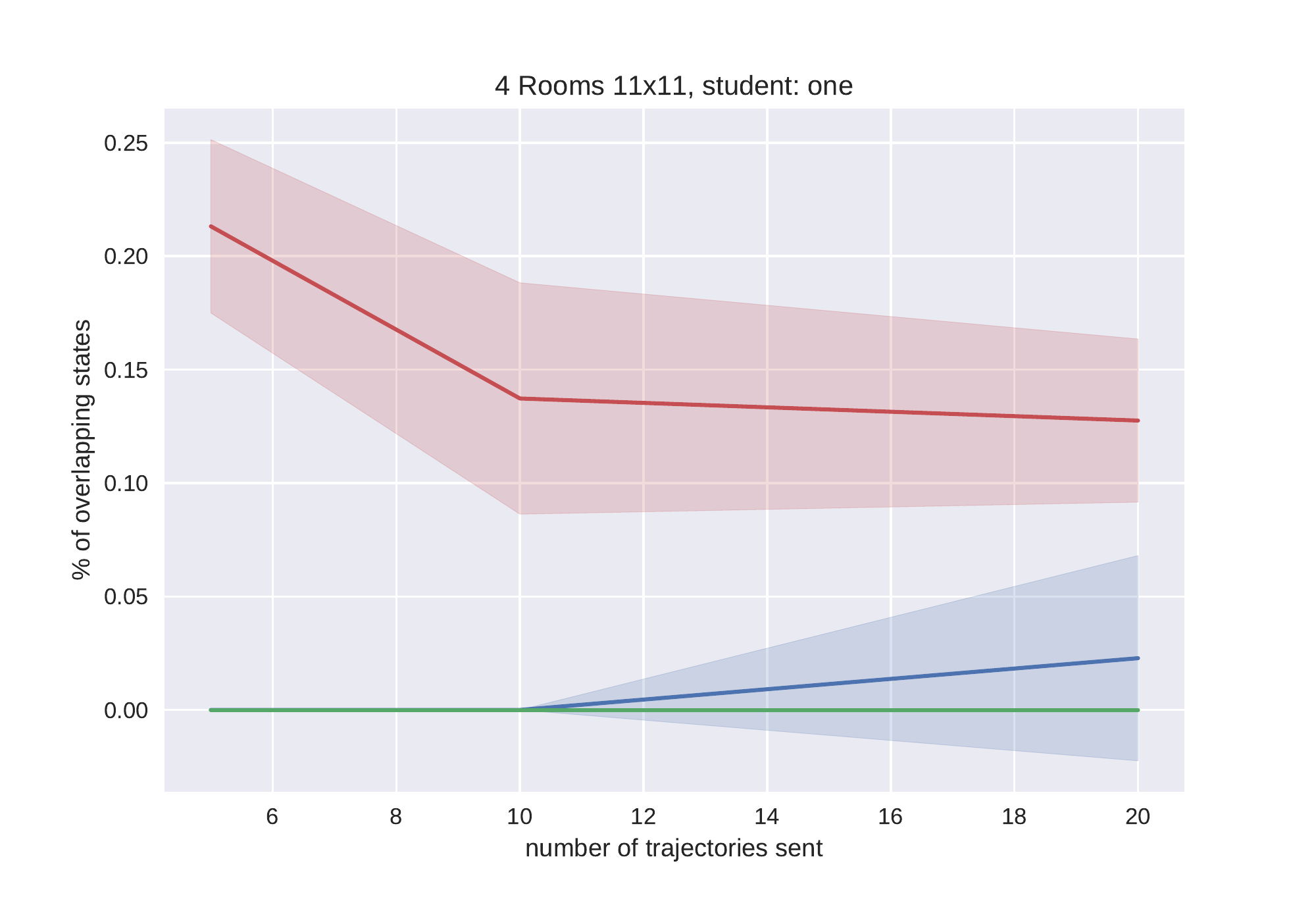} \\ 
    \midrule
    \includegraphics[width=\factor\textwidth]{images/gridworld/7x7/7x7_some.pdf} & \includegraphics[width=\graphfactor\textwidth]{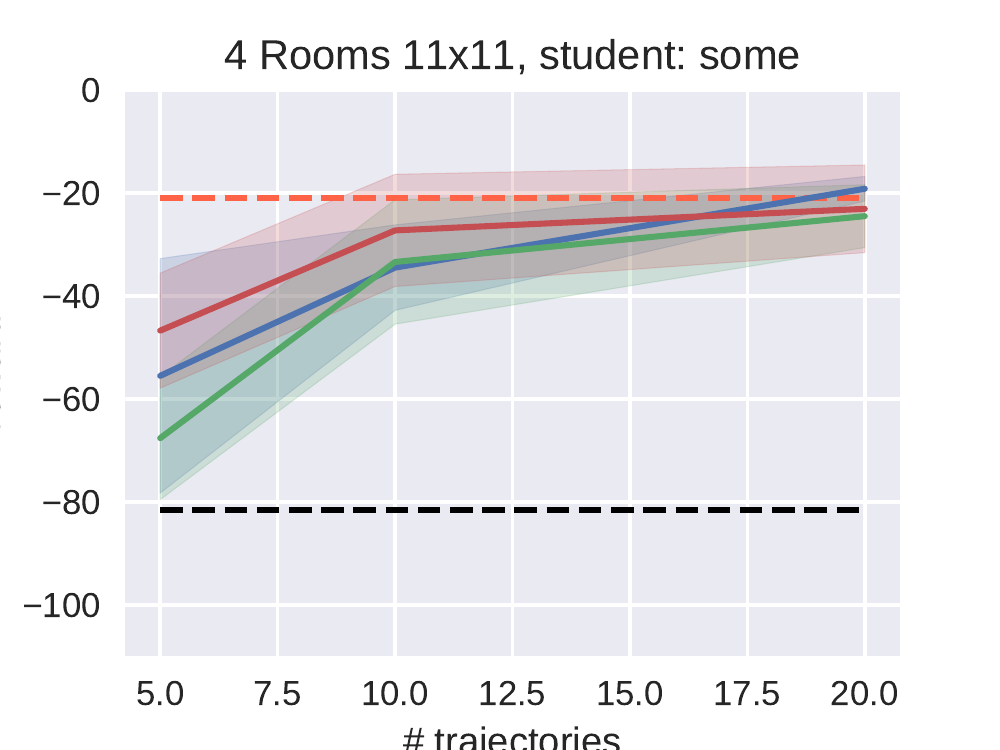} & \includegraphics[width=\graphfactor\textwidth]{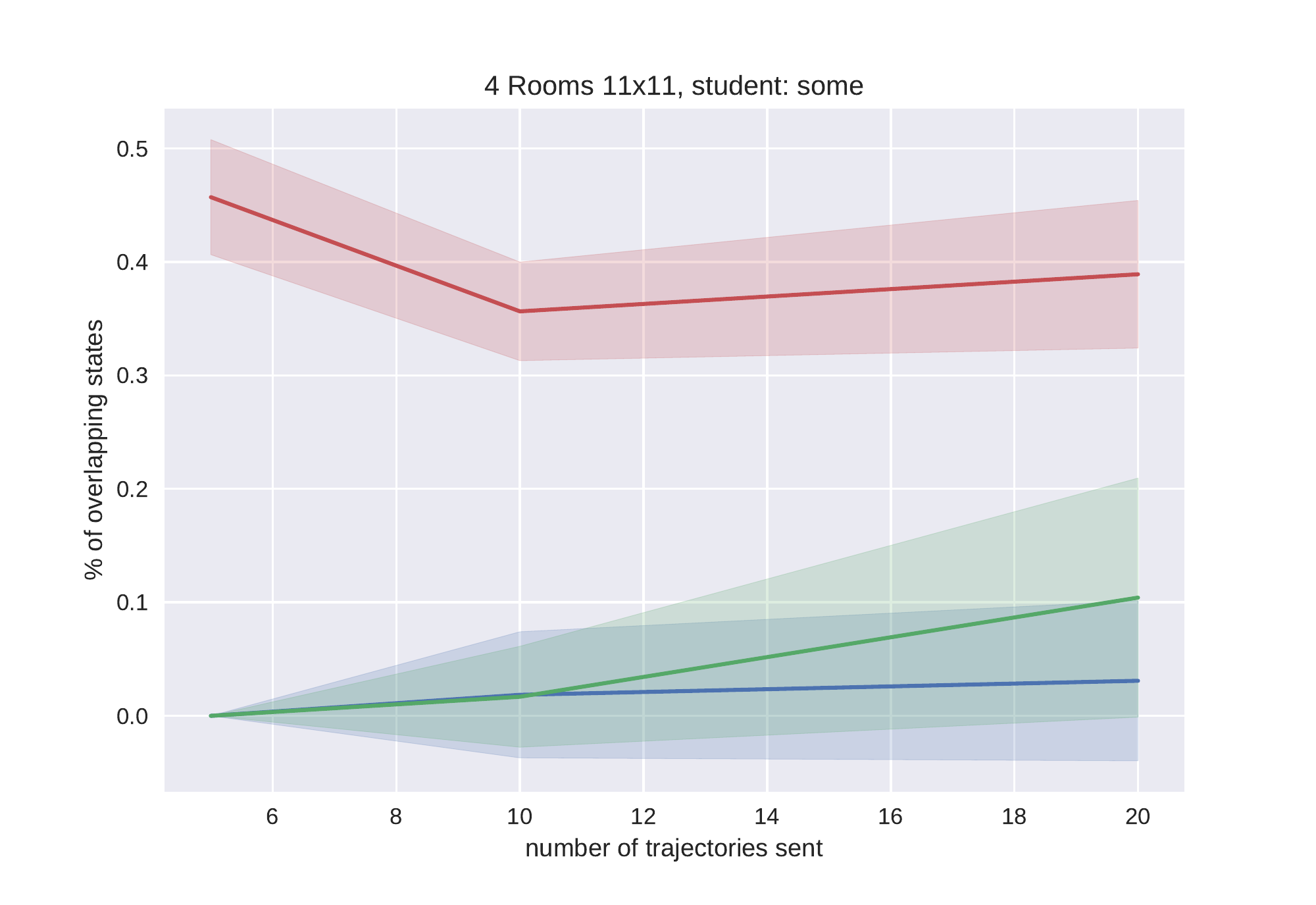} \\ 
    \midrule
    \includegraphics[width=\factor\textwidth]{images/gridworld/7x7/7x7_all.pdf} & \includegraphics[width=\graphfactor\textwidth]{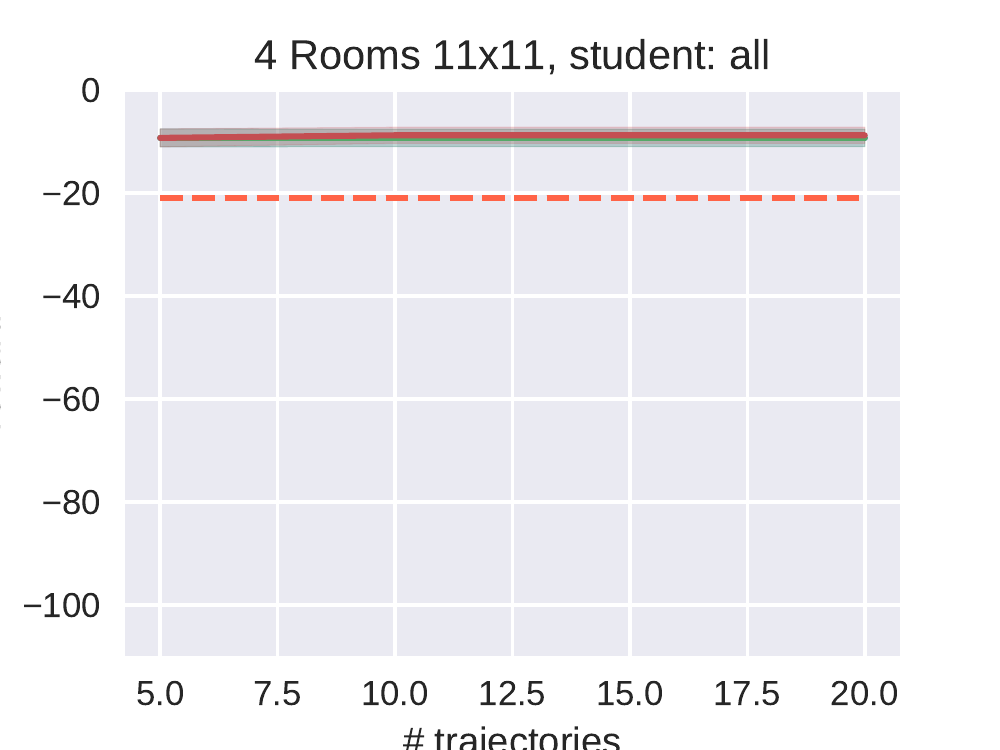} & \includegraphics[width=\graphfactor\textwidth]{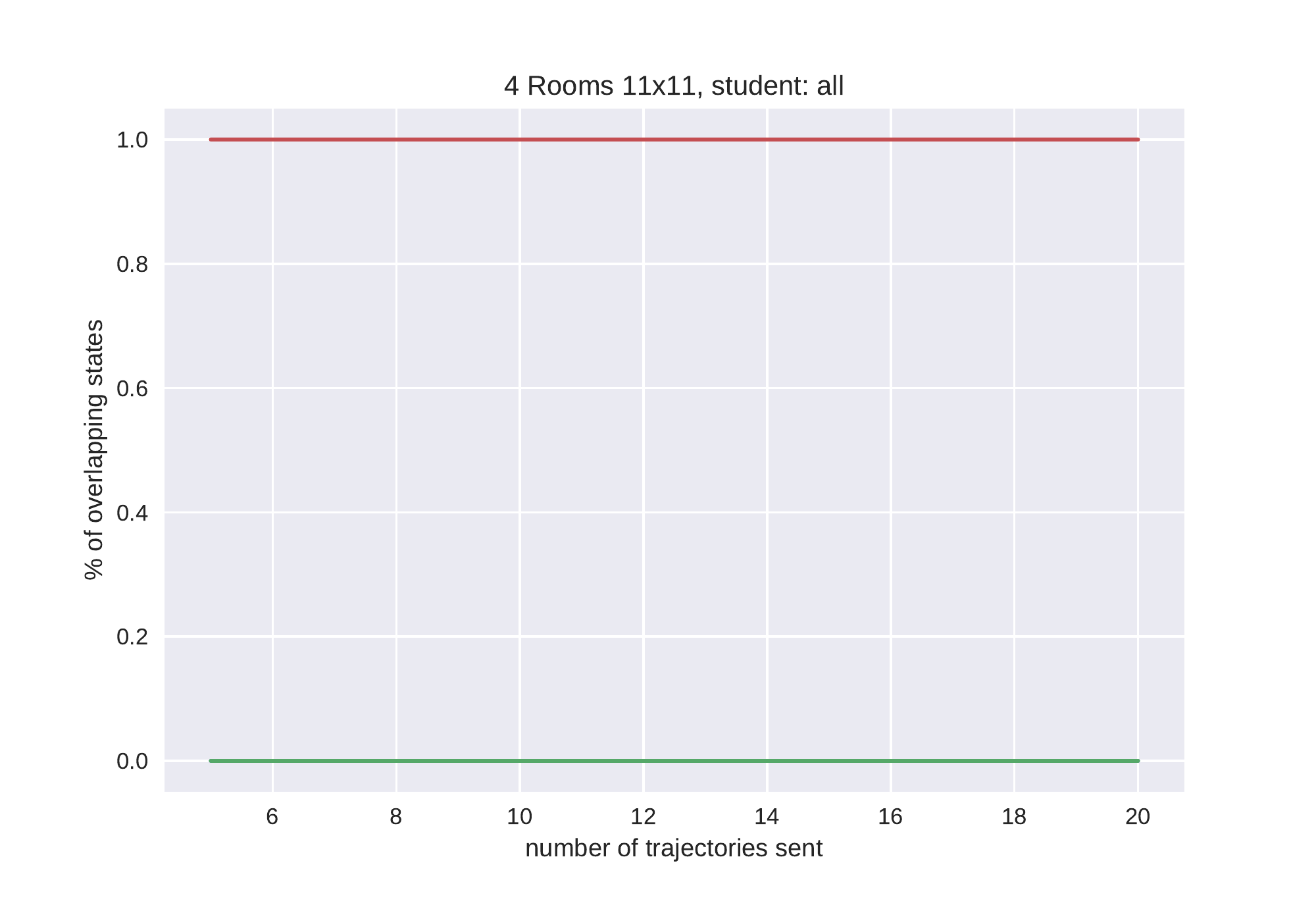} \\ 
    \bottomrule
    \end{tabular} 
    \caption{Student experience with how different teachers teach the student, 4 rooms 11. The dashed lines in the middle column denote the optimal policy's performance.} 
    \label{table:gridworld4rooms11} 
    \end{table} 
\end{center} 

\end{document}